\documentclass{article}

\PassOptionsToPackage{numbers, compress}{natbib}

\usepackage[final,eandd]{neurips_2026}

\usepackage[utf8]{inputenc} 
\usepackage[T1]{fontenc}    
\usepackage{hyperref}       
\usepackage{url}            
\usepackage{booktabs}       
\usepackage{amsfonts}       
\usepackage{nicefrac}       
\usepackage{microtype}      
\usepackage{xcolor}         
\usepackage{multirow}
\usepackage{graphicx}
\usepackage{wrapfig}
\usepackage{amsmath}
\usepackage{todonotes}
\usepackage{xcolor}
\usepackage{morefloats}
\usepackage[normalem]{ulem}
\usepackage[table]{xcolor}
\usepackage{placeins}
\usepackage{float}
\usepackage{amssymb}

\usepackage[T1]{fontenc}
\usepackage[utf8]{inputenc} %
\usepackage{inconsolata}
\usepackage{upquote}
\usepackage{fvextra}
\usepackage[most]{tcolorbox}
\usepackage{xcolor}
\definecolor{QwenBg}{HTML}{F4F8FF}
\definecolor{QwenBd}{HTML}{4A78C2}
\definecolor{ClaudeBg}{HTML}{FFF8F0}
\definecolor{ClaudeBd}{HTML}{B6782F}
\definecolor{InternBg}{HTML}{F3FBF6}
\definecolor{InternBd}{HTML}{2D8A57}

\makeatletter

\DeclareRobustCommand{\onedot}{%
  \futurelet\@let@token\@onedot%
}
\def\@onedot{%
  \ifx\@let@token.%
  \else.\null%
  \fi\xspace%
}
\makeatother
\usepackage{subcaption}

\newcommand{\abbrstyle}[1]{\textit{#1}}

\DeclareRobustCommand{\eg}{\abbrstyle{e.g}\onedot}

\DeclareRobustCommand{\ie}{\abbrstyle{i.e}\onedot}

\DeclareRobustCommand{\iec}{\ie,}

\DefineVerbatimEnvironment{TemplateVerb}{Verbatim}{
    breaklines=true,
    breakanywhere=true,
    breaksymbolleft={},
    breaksymbolright={},
    breakautoindent=false,
    fontsize=\footnotesize
  }

  \newtcolorbox{templatebox}[1]{
    enhanced,
    breakable,
    colback=gray!3,
    colframe=gray!55,
    boxrule=0.5pt,
    arc=2pt,
    left=6pt,right=6pt,top=6pt,bottom=6pt,
    title=\texttt{\detokenize{#1}},
    coltitle=black,
    fonttitle=\bfseries\footnotesize
  }
\usepackage{enumitem}
\usepackage{titletoc}
\usepackage{xcolor}
\usepackage[most]{tcolorbox}
  \usepackage{listings}

  \lstdefinestyle{taskjson}{
    basicstyle=\ttfamily\scriptsize,
    breaklines=true,
    columns=fullflexible,
    showstringspaces=false,
    keepspaces=true,
  }
\definecolor{AppPurple}{HTML}{7A35E8}

\newcommand{\AppendixTOCStyle}{%
\titlecontents{section}
  [2.2em]
  {\color{AppPurple}\bfseries}
  {\contentslabel{2.2em}}
  {}
  {\titlerule*[0.6pc]{.}\contentspage}
  [\vspace{0.35em}]
\titlecontents{subsection}
  [5.2em]
  {\color{AppPurple}\bfseries}
  {\contentslabel{2.6em}}
  {}
  {\titlerule*[0.6pc]{.}\contentspage}
  [\vspace{0.15em}]
}

\title{How to ``train'' a surgeon? Benchmarking generalist agents in surgical scene understanding}

\author{\textbf{Gengyuan Zhang}\textsuperscript{1,2}\thanks{Project lead. Email: \texttt{zhang@dbs.ifi.lmu.de}.} \quad \textbf{Xiao Han}\textsuperscript{1}\thanks{Xiao Han, Shijie Zhou, and Abdullah Koukash contributed equally.} \quad \textbf{Shijie Zhou}\textsuperscript{3}\footnotemark[2] \quad \textbf{Abdullah Koukash}\textsuperscript{1}\footnotemark[2] \\ \textbf{Jingpei Wu}\textsuperscript{1,6} \quad \textbf{Jindong Gu}\textsuperscript{4} \quad \textbf{Yuqicheng Zhu}\textsuperscript{5} \quad \textbf{Nassir Navab}\textsuperscript{2,3} \\ \textbf{Volker Tresp}\textsuperscript{1,2} \\[0.5em] {\scriptsize\normalfont \textsuperscript{1}LMU Munich\enspace \textsuperscript{2}MCML\enspace \textsuperscript{3}TU Munich\enspace \textsuperscript{4}University of Oxford\enspace \textsuperscript{5}University of Stuttgart\enspace 
\textsuperscript{6}relAI}\\[0.3em]
{\small\normalfont Project page: \url{https://vitagent.github.io/}}}
\usepackage{xspace}
\newcommand{\bench}{\textsc{SurgKnowBench}\xspace}

\newcommand{\gym}{\textsc{SurgAgentGym}\xspace}
\newcommand{\wiki}{\textsc{SurgWiki}\xspace}

\begin{document}

\maketitle

\begin{abstract}
Evaluating general-purpose multimodal large language models (MLLMs) for surgical scene understanding remains limited due to scarce surgical data and fragmented evaluation protocols. 
We address these challenges through a two-stage evaluation framework for surgical scene understanding. 
In Stage I, we introduce \bench, a diagnostic benchmark that harmonizes public surgical datasets across procedures and tasks and evaluates proprietary and open-source MLLMs to map the boundary of pretrained surgical competence.
\bench reveals that current models retain partial visual priors but remain weak on relational and procedural understanding and workflow assessment. 
Building on this failure frontier and motivated by the scaffolded nature of surgical training, we introduce \gym, an agentic environment where general-purpose MLLM agents can use bounded knowledge search over \wiki{} and general-purpose visual grounding for surgical tasks. 
\gym evaluates whether generalist agents can improve surgical understanding at inference time through tool-mediated evidence use, without additional surgical training.
Our results show that agentic inference narrows the gap mainly for perception-heavy tasks, while fine-grained localization and expert-style assessment remain persistent bottlenecks. This opens a path toward future studies of agentic surgical understanding.

\end{abstract}

\section{Introduction}

\begin{figure}[htbp]
    \centering
    \includegraphics[width=0.96\linewidth]{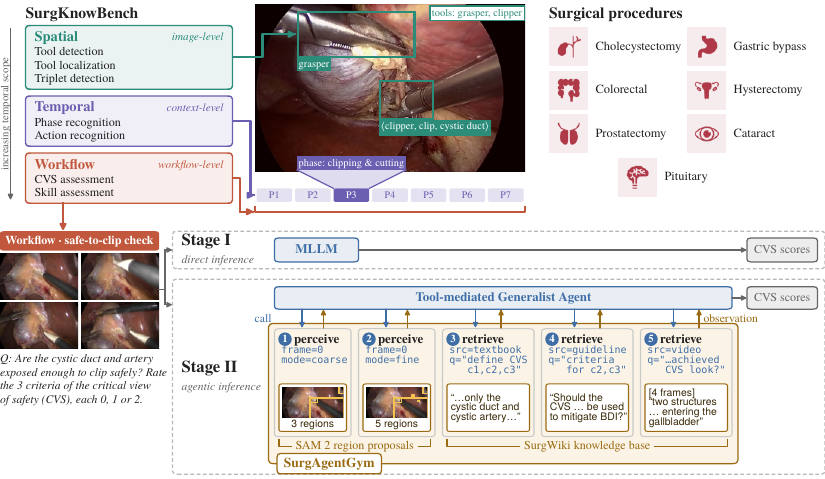}
\caption{How to "train" a surgeon? In our two-stage evaluation: Stage I uses \bench to diagnose zero-shot surgical competence of MLLMs, while Stage II uses \gym on \bench{}-Hard to test whether general-purpose MLLM agents can improve at inference time through bounded knowledge search and generalist visual grounding.}
    \label{fig:placeholder}
\end{figure}

Surgical superintelligence requires models to possess visual, procedural, and expertise-intensive scene understanding capabilities. A model must recognize procedure-specific anatomy, instruments, tissue states, and actions, situate them within an operation, and reason over the procedure-level workflow.
However, progress toward surgical superintelligence is always constrained by two obstacles. 
First, surgical data are scarce, fragmented, and difficult to share: intraoperative videos are restricted by privacy constraints, expert annotation is costly and procedure-specific, and surgical practice varies across procedures, institutions, and surgeons~\cite{nwoye2022rendezvous,wagner2023heichole}. These make it difficult to rely on the standard recipe of large-scale supervised pretraining and domain fine-tuning.
Second, the surgical competence of general-purpose MLLMs remains poorly characterized: it is unclear which surgical tasks are already supported by general-purpose multimodal pretraining, and which failures persist without domain-specific adaptation. 
\emph{These two obstacles motivate a staged question}: what do MLLMs understand about surgery with direct inference, and, when large-scale surgical fine-tuning is impractical, can MLLMs improve surgical understanding at inference time without additional training?

We begin with \textbf{\emph{Stage I: Direct Inference}}, inquiring what current general-purpose MLLMs already understand about surgery. 
Existing surgical benchmarks~\cite{valderrama2022towards, zeng2025surgvlm} and supervised models~\cite{nwoye2022rendezvous,twinanda2017endonet} are often narrow in procedure coverage, task type, or model selection, leaving a need for a unified evaluation protocol. To address this, we introduce \bench, a comprehensive diagnostic benchmark that aggregates and harmonizes a broad collection of surgical datasets spanning \emph{7 surgical procedures} and \emph{7 tasks}, yielding \emph{376{,}602 samples}.
We evaluate \emph{17} proprietary and open-source MLLMs in a zero-shot setting across multiple surgical procedures and three task families: spatial recognition, temporal reasoning, and workflow assessment, to systematically understand model performance bottlenecks.

The Stage I results (detailed in Sec.~\ref{sec:diag}) reveal a difficult frontier for current MLLMs in all task domains.
This raises a second question: when data-intensive surgical training is impractical, how can they achieve surgical understanding capability? 
Surgical training for human surgeons suggests a path forward. Human trainees do not learn from labeled examples alone; they study anatomy and surgical ontologies, consult procedural references, review demonstrations and prior cases, and connect abstract knowledge to concrete operative scenes. The transferable capability is not mere memorization, but \emph{epistemic} control: recognizing when knowledge is missing, seeking relevant evidence, grounding that evidence in the observed scene, and using it to support a decision. 
Inspired by this, we explore an analogous setting for MLLMs, \textbf{\emph{Stage II: Agentic Evaluation}}: if we harness general-purpose MLLMs as generalist agents and equip them with human-like abilities (to see, to search, and to think), can new surgical scene-understanding capabilities emerge?

As illustrated in Fig.~\ref{fig:placeholder}, we instantiate this evaluation with a new agentic environment, \gym, in which generalist MLLM agents can explore surgical tasks.
To equip agents with abilities equivalent to those of human trainees, \gym provides two complementary tools.
\emph{Visual grounding tools} help agents operate on the current surgical input, enabling them to localize semantic boundaries and bridge the visual grounding gap across models.
\emph{Knowledge acquisition tools} allow agents to retrieve and inspect evidence from a bounded multimodal surgical knowledge base containing videos, images, articles, tables, PDFs, and procedural references.
We curate an abundant set of \emph{open-access, modality-rich, broad-coverage, expert-reviewed} sources to build the knowledge base, and conduct thorough data cleaning and indexing to enable robust retrieval.

On a hard set of \bench, termed \bench-Hard, we evaluate agents with tool access. Final-answer scores measure task success, while trajectory analysis examines whether the agent recognizes an information need, selects appropriate tools, retrieves relevant evidence, grounds visual evidence in the surgical input, and integrates information into a contract-compliant answer. This makes \gym more than a question-answering benchmark: it exposes agent behavior as an observable and traceable object of evaluation, a requirement for medical-facing AI where the supporting evidence trail must be inspectable.

Our contributions are:
\begin{itemize}
    \item \textbf{\bench:} a diagnostic benchmark that harmonizes public surgical-understanding datasets to evaluate zero-shot general-purpose MLLMs' surgical competence.
    \item \textbf{\gym:} a controlled tool-mediated agentic environment for generalist MLLM agents, with bounded knowledge acquisition tools over \wiki and general-purpose visual grounding tools.
    \item \textbf{Two-stage evaluation on surgical scene understanding:} we provide a systematic evaluation comprising direct inference and agentic evaluation with trajectory-level analysis for evaluating generalist agents on surgical scene understanding.
\end{itemize}

\section{\bench: diagnosing pretrained surgical competence}
\label{sec:diag}
We first inquire what surgical competence is already present in pretrained MLLMs.

\subsection{Task taxonomy and datasets}
\label{sec:task_taxonomy}
Surgical scene understanding requires models to interpret an operative scene at multiple levels, from recognizing instruments and their interactions to identifying procedural state and assessing higher-level aspects of surgical workflow and performance~\cite{carstens2025artificial,maier2022surgical}. 

\noindent\textbf{Task taxonomy.}
\begin{table*}[t]
\caption{\bench inventory: we organize 11 datasets \(\times\) 7 task axes across 7 surgical procedures. \emph{Tool} = tool detection, \emph{Trip.} = \((\mathrm{instrument}, \mathrm{verb}, \mathrm{target})\) triplet detection, \emph{Loc.} = tool localization, \emph{CVS} = critical-view-of-safety grading, and \emph{Skill} subsumes the three GOALS-style assessments~\citep{wagner2023heichole} (holistic, Calot-phase, and dissection-phase). 
}
\label{tab:surgknowbench_inventory}
\centering\tiny
\setlength{\tabcolsep}{7pt}
\renewcommand{\arraystretch}{1.0}
\begin{tabular}{@{}ll*{7}{c}rr@{}}
\toprule
\multicolumn{2}{c}{Data Source} &
\multicolumn{3}{c}{Spatial} &
\multicolumn{2}{c}{Temporal} &
\multicolumn{2}{c}{Workflow} &
\multicolumn{2}{c}{\# samples} \\
\cmidrule(lr){1-2}
\cmidrule(lr){3-5}
\cmidrule(lr){6-7}
\cmidrule(lr){8-9}
\cmidrule(lr){10-11}
Dataset & Surgery &
Tool & Trip. & Loc. &
Phase & Action &
CVS & Skill &
Full & Hard \\
\midrule
Cholec80~\citep{twinanda2017endonet}            & Lap.\ cholecystectomy           &            &            &            & \checkmark &            &            &            &  42{,}098 &  50 \\
CholecT50~\citep{nwoye2022rendezvous}           & Lap.\ cholecystectomy           & \checkmark & \checkmark & \checkmark &            &            &            &            &  16{,}722 & 220 \\
Cholec80-CVS~\citep{rios2023cholec80cvs}        & Lap.\ cholecystectomy           &            &            &            &            &            & \checkmark &            &        96 &  96 \\
HeiChole~\citep{wagner2023heichole}             & Lap.\ cholecystectomy           &            &            &            &            & \checkmark &            & \checkmark &  36{,}137 &  86 \\
MultiBypass140~\citep{lavanchy2024multibypass140} & Lap.\ Roux-en-Y bypass        &            &            &            & \checkmark &            &            &            & 171{,}633 &  50 \\
AutoLaparo~\citep{wang2022autolaparo}           & Lap.\ hysterectomy              & \checkmark &            &            & \checkmark &            &            &            &  28{,}491 & 100 \\
HeiCo~\citep{maierhein2021heico}                & Lap.\ colorectal resection      &            &            &            & \checkmark &            &            &            &   4{,}057 &  50 \\
GraSP~\citep{ayobi2025tapis}                    & Robotic radical prostatectomy   & \checkmark &            &            & \checkmark &            &            &            &  44{,}022 & 100 \\
PitVis~\citep{das2025pitvis}                    & Endoscopic pituitary            & \checkmark &            &            & \checkmark &            &            &            &   2{,}310 & 100 \\
Cataract-101~\citep{schoeffmann2018cataract101} & Cataract surgery                &            &            &            & \checkmark &            &            &            &  12{,}503 &  50 \\
CATARACTS~\citep{alhajj2019cataracts}           & Cataract surgery                & \checkmark &            &            &            &            &            &            &  18{,}533 &  50 \\
\midrule
\textbf{Total} &  &
  &   &   &   &   &   &   &
\textbf{376{,}602} & \textbf{952} \\
\bottomrule
\end{tabular}

\end{table*}
As shown in Tab.~\ref{tab:surgknowbench_inventory}, \bench organizes surgical scene understanding tasks into three task domains: (1) \emph{Spatial recognition} tests whether a model can identify visible surgical entities and localized visual relations in an operative scene; (2) \emph{Temporal reasoning} evaluates whether a model can map visual evidence to procedural state; (3) \emph{Workflow assessment} evaluates whether a model can make expert-style judgments about operative safety and technical skill.

\noindent\textbf{Datasets.}
\label{sec:datasets}
We build on 11 public surgical datasets spanning seven procedures: laparoscopic cholecystectomy (Cholec80~\cite{twinanda2017endonet}, CholecT50~\cite{nwoye2022rendezvous}, Cholec80-CVS~\cite{rios2023cholec80cvs}, and HeiChole~\cite{wagner2023heichole}), Roux-en-Y gastric bypass (MultiBypass140~\cite{lavanchy2024multibypass140}), laparoscopic hysterectomy (AutoLaparo~\cite{wang2022autolaparo}), laparoscopic proctocolectomy (HeiCo~\cite{maierhein2021heico}), robot-assisted radical prostatectomy (GraSP~\cite{ayobi2025tapis}), endoscopic pituitary surgery (PitVis~\cite{das2025pitvis}), and cataract surgery (Cataract-101~\cite{schoeffmann2018cataract101} and CATARACTS~\cite{alhajj2019cataracts}).

\subsection{Evaluation protocol}
\label{sec:bench_eval_protocol}

\noindent\textbf{Dataset harmonization.}
For each source dataset, we convert raw annotations into a common QA-style evaluation format while preserving the label sets. Dataset-specific labels are mapped to canonical strings. Frame-level tasks
are represented by annotated frames, while interval-level tasks such as CVS and skill assessment retain temporal metadata and
representative visual inputs. Further details on frame sampling and answer schemas are provided in
Appendix~\ref{app:frame_sampling} and Appendix~\ref{app:task_schema}.

\noindent\textbf{Evaluation metrics.}
For recognition tasks with a fixed label set (\ie tool detection, action recognition, and triplet detection), we report macro-averaged \(F_1\), which weights each class equally and better exposes rare-label failures. Triplet detection is scored over complete \(\langle\mathrm{instrument},\mathrm{verb},\mathrm{target}\rangle\) triplets. For phase recognition, we report strict label accuracy. For tool localization, we report label-constrained \(F_1@0.5\) after Hungarian matching, requiring both instrument-label agreement and box overlap. For CVS assessment, we report mean per-key exact-score accuracy together with mean quadratic-weighted Cohen's \(\kappa\) over the three CVS criteria. For skill assessment, we report mean absolute error (MAE) together with mean quadratic-weighted \(\kappa\) over the GOALS-style rubric keys. Auxiliary metrics are defined in Appendix~\ref{app:metrics} and reported in Appendix~\ref{app_stage1}.

\subsection{Diagnosing pretrained MLLMs’ surgical competence} 
\begin{figure}[htbp]
    \centering
    \includegraphics[width=0.96\linewidth]{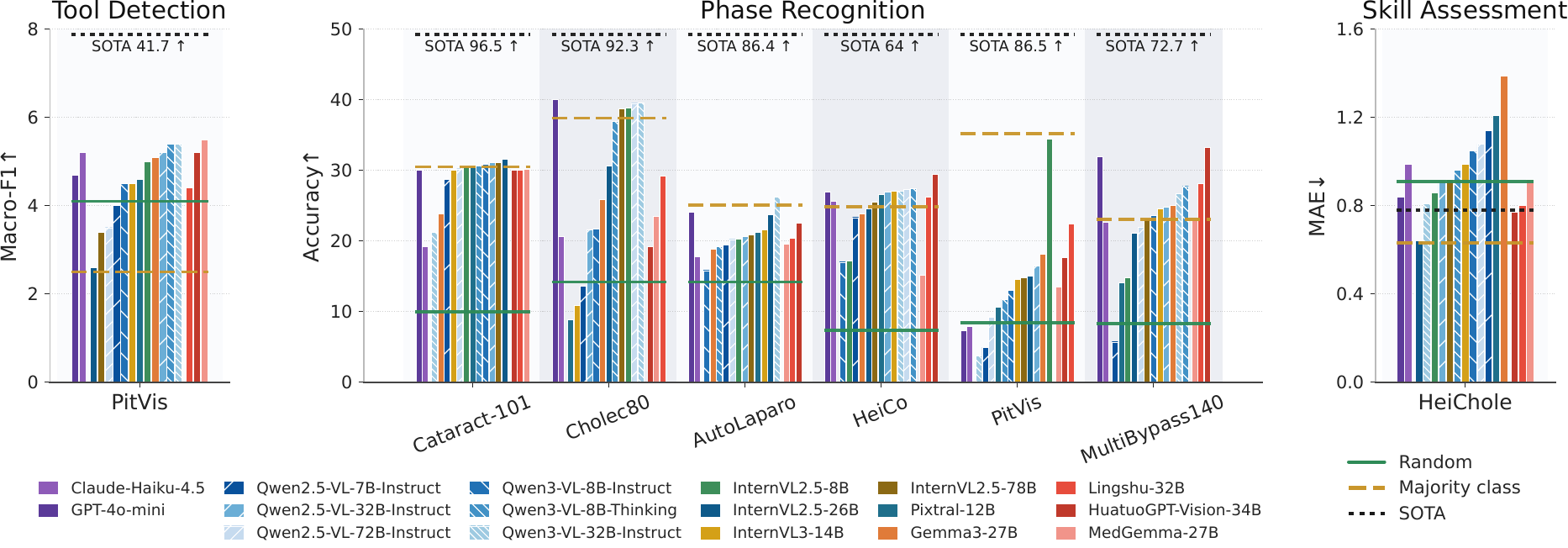}
\caption{Stage I performance on dataset--task breakdown.}
    \label{fig:sota}
\end{figure}

Stage I diagnoses the extent and limits of zero-shot surgical competence
in general-purpose MLLMs. Fig.~\ref{fig:sota} places selected dataset--task scores alongside published specialist reference results.

\noindent\textbf{Limited surgical competence.}
Tab.~\ref{tab:c0_full_suite_summary} shows that the evaluated MLLMs
have some ability to identify surgical instruments, but recognition
and localization remain weak. On tool detection, 16 of the 17 models
score above the majority baseline, yet the best score reaches only
26.1\% macro-F1, compared with 13.5\% for the majority baseline.
Accurate localization also remains difficult: the best model achieves
only 22.7\% (F1@0.5), which requires both the correct instrument
label and sufficient overlap with the reference bounding box.

\noindent\textbf{Relational and procedural failure frontiers.}
This limited recognition does not extend to reliable relational or
procedural judgments. Triplet detection is especially brittle:
every model scores below 5\% macro-F1, highlighting the difficulty
of identifying complete instrument--verb--target interactions.
Phase recognition exposes a related procedural gap: even the best
dataset-averaged accuracy remains below the majority baseline,
indicating weak procedural understanding from visual evidence.

\noindent\textbf{Models' weak agreement on surgical assessment.}
Workflow assessment exposes a different failure mode. Several models
achieve high CVS accuracy by collapsing to frequent rubric values,
yet their chance-corrected agreement remains near zero or negative.
Our analysis on prediction distribution distinguishes two response patterns:
majority collapse and optimistic
over-crediting, in which models assign achieved-criterion scores
when the reference indicates otherwise. The former results from
label imbalance. For example, InternVL2.5-8B, repeatedly predicts the most frequent score
for each criterion. This yields 71.2\% accuracy with $\kappa=0$:
the predictions match common scores in the dataset but fail to
distinguish cases with different reference scores.
By contrast, over-crediting reveals a directional bias
toward overstating the surgical skills. For instance,  GPT-4o-mini and Lingshu-32B frequently mark CVS criteria as achieved even though the human annotations indicate low scores. Their accuracies (15.6\% and 9.4\%) are even below the random baseline (32.8\%).
We also notice that for GOALS skill assessment, such dataset bias also prevails: ratings are concentrated around the middle of the 5-level scale. As a result, predicting the most frequent score for each criterion produces small errors on many cases without using the surgical images. 
This again yields a shortcut that allows models to achieve low MAE without really distinguishing surgical cases.

\noindent\textbf{Leaderboard.}
Qwen3-VL-32B achieves the best mean rank (3.6) across the seven tasks among the 17 evaluated models. However,
its lead over the runner-up is statistically significant only for tool detection. Performance does not increase monotonically with model size, even within the same model family.
Surprisingly, the evaluated proprietary and medical-domain models do not show
a consistent advantage over general-purpose open-source models.
These groups have task-specific strengths: GPT-4o-mini leads action recognition, while HuatuoGPT-Vision
achieves the highest phase accuracy. Their localization performance
is particularly weak: all five models score at most \(1.6\%\)
\(F_1@0.5\), compared with \(22.7\%\) for InternVL2.5-78B.
Medical-domain models also trail Qwen3-VL-32B on both
workflow-assessment tasks under \(\kappa\), indicating that
their specialization does not translate into stronger
rubric-based surgical judgments.

\begin{table}[t]
\centering
\tiny
\setlength{\tabcolsep}{4pt}
\caption{Stage~I diagnostic evaluation on the full \bench test set.
F1 and accuracy are percentages; $\kappa$ is multiplied by 100. Skill MAE is reported on the original ordinal scale (lower is better). \emph{Rank} is the mean rank across the seven task axes over all 17 evaluated models, using $\kappa$ for
workflow-assessment axes, shown with 95\% CIs from a video-clustered bootstrap
(1000 resamples). \textbf{Bold}/\underline{underline} indicate the best/second-best values among the evaluated models for each metric (including ties). $^*$ marks a significant lead over the runner-up (paired video-clustered bootstrap, BH-FDR $q\leq0.05$). $^\dagger$ marks task axes without significance tests due to limited source videos ($\leq5$). Green/red shading indicates better/worse performance than the random baseline; workflow columns are shaded by $\kappa$.}
\label{tab:c0_full_suite_summary}
\begin{tabular}{@{}l ccc cc cc c@{}}
\toprule
\multirow{3}{*}{Model}
  & \multicolumn{3}{c}{Spatial Recognition}
  & \multicolumn{2}{c}{Temporal Reasoning}
  & \multicolumn{2}{c}{Workflow Assessment}
  & \multirow{3}{*}{Rank$\downarrow$} \\
\cmidrule(lr){2-4} \cmidrule(lr){5-6} \cmidrule(lr){7-8}
 & Tool Det. & Triplet Det.$^\dagger$ & Tool Loc.$^\dagger$ & Phase Rec. & Action Rec. & CVS Ass. & Skill Ass. \\
 & {\scriptsize macro-F1$\uparrow$} & {\scriptsize macro-F1$\uparrow$} & {\scriptsize $F_1@0.5\uparrow$}
 & {\scriptsize Acc.$\uparrow$} & {\scriptsize macro-F1$\uparrow$}
 & {\scriptsize Acc./$\kappa\uparrow$} & {\scriptsize MAE$\downarrow$/$\kappa\uparrow$} & \\
\midrule
\multicolumn{9}{@{}l}{\textit{Proprietary}}\\
\cmidrule{1-9}
GPT-4o-mini~\citep{openai2024gpt4omini} & \cellcolor[HTML]{C9EAC9} \underline{22.8} & \cellcolor[HTML]{FDFEFD} 1.6 & 0.0 & \cellcolor[HTML]{BEE6BE} \underline{24.7} & \cellcolor[HTML]{BEE6BE} \textbf{27.1}\textsuperscript{*} & \cellcolor[HTML]{FCF1F1} 15.6/0.9 & \cellcolor[HTML]{FDF2F2} 0.84/0.2 & 7.7\,{\fontsize{4}{5}\selectfont[6.9,9.4]} \\
Claude-Haiku-4.5~\citep{anthropic2025claudehaiku45} & \cellcolor[HTML]{DAF1DA} 18.2 & \cellcolor[HTML]{DCF2DC} 2.9 & \cellcolor[HTML]{F3FBF3} 1.6 & \cellcolor[HTML]{D2EED2} 18.4 & \cellcolor[HTML]{D0EDD0} 21.8 & \cellcolor[HTML]{F5C8C8} 17.0/-4.8 & \cellcolor[HTML]{CAEBCA} 0.99/\underline{10.9} & 8.9\,{\fontsize{4}{5}\selectfont[8.4,9.9]} \\
\midrule
\multicolumn{9}{@{}l}{\textit{Open-source ($<$10B)}}\\
\cmidrule{1-9}
InternVL2.5-8B~\citep{chen2024internvl25} & \cellcolor[HTML]{FCEFEF} 10.4 & \cellcolor[HTML]{F1FAF1} 1.9 & \cellcolor[HTML]{D6EFD6} 11.3 & \cellcolor[HTML]{C1E7C1} 23.8 & \cellcolor[HTML]{FCF0F0} 11.7 & \cellcolor[HTML]{FBE8E8} \textbf{71.2}/0.0 & \cellcolor[HTML]{F7D4D4} 0.86/-7.2 & 11.7\,{\fontsize{4}{5}\selectfont[10.7,12.6]} \\
Qwen2.5-VL-7B-Instruct~\citep{bai2025qwen25vl} & \cellcolor[HTML]{E9F7E9} 15.0 & \cellcolor[HTML]{FEF9F9} 1.5 & \cellcolor[HTML]{D5EFD5} 11.4 & \cellcolor[HTML]{E4F4E4} 14.1 & \cellcolor[HTML]{FAE3E3} 8.7 & \cellcolor[HTML]{FBE8E8} \textbf{71.2}/0.0 & \cellcolor[HTML]{D5EFD5} 1.14/8.0 & 11.6\,{\fontsize{4}{5}\selectfont[11.1,13.4]} \\
Qwen3-VL-8B-Instruct~\citep{qwen2025qwen3vl} & \cellcolor[HTML]{D6EFD6} 19.3 & \cellcolor[HTML]{ECF8EC} 2.1 & \cellcolor[HTML]{D0EDD0} 13.8 & \cellcolor[HTML]{D1EDD1} 18.9 & \cellcolor[HTML]{FFFEFE} 13.5 & \cellcolor[HTML]{D2EED2} 31.6/\underline{5.4} & \cellcolor[HTML]{EAF7EA} 1.05/3.9 & 8.1\,{\fontsize{4}{5}\selectfont[7.1,9.7]} \\
Qwen3-VL-8B-Thinking~\citep{qwen2025qwen3vl} & \cellcolor[HTML]{CBEBCB} 22.1 & \cellcolor[HTML]{DEF2DE} 2.8 & \cellcolor[HTML]{CAEBCA} \underline{16.3} & \cellcolor[HTML]{C3E8C3} 23.2 & \cellcolor[HTML]{F7D3D3} 3.9 & \cellcolor[HTML]{ECF8EC} \underline{58.3}/2.7 & \cellcolor[HTML]{E5F5E5} 0.96/4.7 & \underline{6.3}\,{\fontsize{4}{5}\selectfont[5.1,8.6]} \\
\midrule
\multicolumn{9}{@{}l}{\textit{Open-source (10--30B)}}\\
\cmidrule{1-9}
Pixtral-12B-2409~\citep{agrawal2024pixtral} & \cellcolor[HTML]{E1F3E1} 16.7 & \cellcolor[HTML]{E8F6E8} 2.3 & \cellcolor[HTML]{EAF7EA} 4.0 & \cellcolor[HTML]{D5EFD5} 17.7 & \cellcolor[HTML]{C9EAC9} \underline{23.8} & \cellcolor[HTML]{FBE8E8} \textbf{71.2}/0.0 & \cellcolor[HTML]{FCEEEE} 1.21/-0.5 & 10.7\,{\fontsize{4}{5}\selectfont[9.4,11.3]} \\
InternVL3-14B~\citep{zhu2025internvl3} & \cellcolor[HTML]{EBF7EB} 14.7 & \cellcolor[HTML]{E0F3E0} 2.7 & \cellcolor[HTML]{DCF1DC} 8.9 & \cellcolor[HTML]{CEECCE} 19.5 & \cellcolor[HTML]{F3FBF3} 14.4 & \cellcolor[HTML]{FAE5E5} 24.0/-0.4 & \cellcolor[HTML]{E9F7E9} 0.99/4.0 & 9.7\,{\fontsize{4}{5}\selectfont[8.6,11.7]} \\
InternVL2.5-26B~\citep{chen2024internvl25} & \cellcolor[HTML]{ECF8EC} 14.6 & \cellcolor[HTML]{F3FAF3} 1.9 & \cellcolor[HTML]{CEECCE} 14.6 & \cellcolor[HTML]{C9EAC9} 21.3 & \cellcolor[HTML]{FBE7E7} 9.8 & \cellcolor[HTML]{FAE2E2} 34.4/-0.7 & \cellcolor[HTML]{FCF1F1} \textbf{0.64}/0.0 & 12.6\,{\fontsize{4}{5}\selectfont[9.9,12.9]} \\
Gemma-3-27B-it~\citep{gemmateam2025gemma3} & \cellcolor[HTML]{CDECCD} 21.5 & \cellcolor[HTML]{D1EDD1} \underline{3.6} & 0.0 & \cellcolor[HTML]{C8EAC8} 21.4 & \cellcolor[HTML]{FDF5F5} 12.6 & \cellcolor[HTML]{F6CDCD} 37.5/-3.9 & \cellcolor[HTML]{EEF8EE} 1.39/3.3 & 10.0\,{\fontsize{4}{5}\selectfont[8.0,10.7]} \\
\midrule
\multicolumn{9}{@{}l}{\textit{Open-source ($>$30B)}}\\
\cmidrule{1-9}
Qwen2.5-VL-32B-Instruct~\citep{bai2025qwen25vl} & \cellcolor[HTML]{DBF1DB} 18.0 & \cellcolor[HTML]{FCF0F0} 1.2 & \cellcolor[HTML]{E2F4E2} 6.5 & \cellcolor[HTML]{C6E9C6} 22.0 & \cellcolor[HTML]{F5FBF5} 14.3 & \cellcolor[HTML]{FAE4E4} 24.3/-0.4 & \cellcolor[HTML]{EAF7EA} 0.91/3.9 & 10.9\,{\fontsize{4}{5}\selectfont[9.1,12.0]} \\
Qwen3-VL-32B-Instruct~\citep{qwen2025qwen3vl} & \cellcolor[HTML]{BEE6BE} \textbf{26.1}\textsuperscript{*} & \cellcolor[HTML]{E4F5E4} 2.5 & \cellcolor[HTML]{CEECCE} 14.6 & \cellcolor[HTML]{C1E7C1} 23.7 & \cellcolor[HTML]{D2EED2} 21.3 & \cellcolor[HTML]{BEE6BE} 31.6/\textbf{8.2} & \cellcolor[HTML]{BEE6BE} 0.81/\textbf{14.3} & \textbf{3.6}\,{\fontsize{4}{5}\selectfont[2.7,5.4]} \\
Qwen2.5-VL-72B-Instruct~\citep{bai2025qwen25vl} & \cellcolor[HTML]{DAF1DA} 18.3 & \cellcolor[HTML]{EFF9EF} 2.0 & \cellcolor[HTML]{D5EFD5} 11.4 & \cellcolor[HTML]{C1E7C1} 23.6 & \cellcolor[HTML]{DDF2DD} 18.5 & \cellcolor[HTML]{F7D4D4} 47.9/-2.7 & \cellcolor[HTML]{D9F0D9} 1.08/7.2 & 8.3\,{\fontsize{4}{5}\selectfont[6.4,9.4]} \\
InternVL2.5-78B~\citep{chen2024internvl25} & \cellcolor[HTML]{C9EAC9} 22.6 & \cellcolor[HTML]{BEE6BE} \textbf{4.9} & \cellcolor[HTML]{BEE6BE} \textbf{22.7} & \cellcolor[HTML]{C0E7C0} 24.1 & \cellcolor[HTML]{FCF1F1} 11.8 & \cellcolor[HTML]{FAE4E4} 25.7/-0.4 & \cellcolor[HTML]{FCF1F1} 0.92/0.1 & 7.0\,{\fontsize{4}{5}\selectfont[4.7,7.9]} \\
\midrule
\multicolumn{9}{@{}l}{\textit{Medical-domain}}\\
\cmidrule{1-9}
HuatuoGPT-Vision-34B-hf~\citep{chen2024towards} & \cellcolor[HTML]{CFECCF} 21.1 & \cellcolor[HTML]{EFF9EF} 2.0 & \cellcolor[HTML]{F7FCF7} 0.9 & \cellcolor[HTML]{BEE6BE} \textbf{24.8} & \cellcolor[HTML]{D1EDD1} 21.4 & \cellcolor[HTML]{FBE9E9} 14.6/0.1 & \cellcolor[HTML]{FDFEFD} \underline{0.77}/1.7 & 7.6\,{\fontsize{4}{5}\selectfont[7.0,9.1]} \\
Lingshu-32B~\citep{lingshu2025} & \cellcolor[HTML]{DBF1DB} 17.9 & \cellcolor[HTML]{E5F5E5} 2.4 & \cellcolor[HTML]{F3FBF3} 1.6 & \cellcolor[HTML]{BFE6BF} 24.6 & \cellcolor[HTML]{E7F6E7} 16.5 & \cellcolor[HTML]{FBE7E7} 9.4/-0.2 & \cellcolor[HTML]{FDF3F3} 0.80/0.4 & 9.2\,{\fontsize{4}{5}\selectfont[7.4,10.8]} \\
MedGemma-27B-it~\citep{medgemma2025} & \cellcolor[HTML]{D9F0D9} 18.4 & \cellcolor[HTML]{E1F4E1} 2.6 & 0.0 & \cellcolor[HTML]{CEECCE} 19.7 & \cellcolor[HTML]{E8F6E8} 16.2 & \cellcolor[HTML]{DEF2DE} 19.4/4.0 & \cellcolor[HTML]{FFFFFF} 0.92/1.6 & 9.1\,{\fontsize{4}{5}\selectfont[8.0,10.7]} \\
\midrule
\multicolumn{9}{@{}l}{\textit{Statistical Baseline}}\\
\cmidrule{1-9}
Random   & 12.5 & 1.6 & 0.0 & 10.2 & 13.5 & 32.8/1.7 & 0.91/1.6 & -- \\
Majority & 13.5 & 1.2 & 0.0 & 29.9 & 17.6 & 71.2/0.0 & 0.63/0.0 & -- \\
\bottomrule
\end{tabular}
\end{table}

\subsection{Constructing the hard set: tasks that resist pretrained competence}
\label{sec:hardset}
Stage I reveals the limited surgical competence of general-purpose MLLMs on \bench. 
We build \bench{}-Hard, a subset where all models fail as a failure frontier. By focusing on episodes that MLLMs fail without external tooling, \bench{}-Hard reduces the pretrained-competence confound in later agentic evaluation and provides a controlled stress test for whether generalist agents can make measurable progress at inference time.

\noindent\textbf{Panel-based difficulty scoring.}
We score the difficulty of samples in Stage I  from a panel of open-source MLLMs. For an episode \(x\), let \(\mathcal{M}_x\) denote the panel models. We define the panel error as \(\mathrm{PanelError}(x)=\frac{1}{|\mathcal{M}_x|}\sum_{m\in\mathcal{M}_x}\left(1-\mathrm{Perf}_{\tau(x)}(m,x)\right)\), where \(\tau(x)\) is the task type of \(x\), and \(\mathrm{Perf}_{\tau}(m,x)\in[0,1]\) is a task-normalized per-episode score.
We construct \bench-Hard by ranking candidate episodes within each dataset by \(\mathrm{PanelError}\).

\section{\gym: harnessing generalist agents for surgical understanding}\label{sec:gym}
Surgical training for human surgeons is a scaffolded process: trainees learn anatomical and surgical procedural knowledge from oracle references and canonical surgical cases and generalize this to real-world tasks.
\gym explores whether generalist agents can achieve surgical competence by solving tasks through an epistemic process with accessible tools, to \emph{see} (\iec visual grounding), to \emph{search} (\iec knowledge acquisition), and to \emph{think}, and (\iec internalized reasoning abilities).

To measure the agents’ behavior in an attributable and explainable way, \gym is designed according to the following key principles:

\noindent\textbf{Generalist perception access.}
Pretrained MLLM backbones have different levels of pixel-level visual grounding ability. For the vision-intensive surgical domain, this can constitute a "congenital" grounding ability gap across different models and can impact surgical understanding ability.

\noindent\textbf{Regulated knowledge search.}
Surgical reasoning requires bounded and auditable evidence access. Unlike general-domain agent benchmarks, surgical agents need to control source identity and provenance to guarantee high-stakes judgments. 
We therefore introduce \wiki, a closed-domain surgical knowledge base whose sources are carefully selected, curated, and provenance-tracked.

\subsection{\wiki: a regulated, multimodal surgical knowledge base}\label{sec:kb}
\begin{figure}[htbp]
    \centering
    \begingroup
    \setlength{\fboxsep}{0.6pt}%
    \makebox[\textwidth][l]{%
      \includegraphics[width=\textwidth]{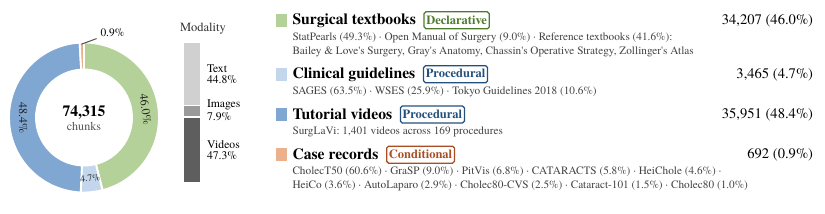}%
      \hspace*{-\textwidth}%
      \raisebox{18.48pt}[0pt][0pt]{\hspace*{0.3530\textwidth}%
        \colorbox{white}{\fontsize{5.42}{6.6}\selectfont\makebox[209pt][l]{\rule[-1.2pt]{0pt}{6.2pt}%
          CholecT50 (60.6\%) \textperiodcentered\ GraSP (9.0\%) \textperiodcentered\ PitVis (6.8\%) \textperiodcentered\ CATARACTS (5.8\%) \textperiodcentered\ HeiChole (4.6\%) \textperiodcentered}}}}%
    \endgroup
    \caption{\wiki knowledge-base composition. The corpus contains 74,315 chunks spanning surgical textbooks, clinical guidelines, tutorial videos, and case records. }
    \label{fig:kb_distribution}
\end{figure}

We present \wiki, a “learning platform” for generalist agents to acquire surgical knowledge.
To ensure broad and sufficient coverage of surgical knowledge, we (1) \emph{regulate} our knowledge selection based on common practices in surgical training~\cite{krathwohl2002revision, drake2013general} and prior surgical knowledge bases~\cite{xiong2024benchmarking, xiong2025mirage, huo2024clinical}; (2) cover \emph{multimodal}, heterogeneous knowledge spanning diverse sources to meet different task requirements; (3) include fundamental surgical knowledge and cover the procedures and task families evaluated in \bench.
\wiki contains 74{,}315 indexable knowledge chunks: 37{,}672 indexed text chunks, 35{,}951 tutorial-video chunks, and 692 case-record chunks, as shown in Fig.~\ref{fig:kb_distribution}.

\noindent\textbf{Reproducibility and governance.} 
For accessibility, we prioritize public, open-access, citable, or otherwise reproducible sources whenever possible. 
For clinical plausibility, we hired 3 practicing surgeons for sanity-checking the assembled knowledge sources for procedural relevance, source appropriateness, and clinical plausibility.
We also evaluate \wiki-Open, a corpus configuration excluding commercial sources (Tab.~\ref{tab:cr_rb_open}). The curation pipeline reconstructs this configuration from sources obtained directly from their providers; we do not distribute the source data.

\noindent\textbf{Knowledge sources.}
We construct \wiki from four provenance-tracked sources: surgical textbooks, clinical practice guidelines, narrated tutorial videos, and annotated case records. These sources provide complementary evidence for the \emph{declarative–procedural–conditional}~\cite{krathwohl2002revision} knowledge ladder. 
Fig.~\ref{fig:kb_distribution} summarizes the distribution and more details are provided in Appendix~\ref{app:kb_sources}.

\noindent\textbf{Data ingestion.}
For multimodal and heterogeneous knowledge, we apply a staged and semantics-aware ingestion pipeline that preserves document structure, source provenance, and visual context.
Ingestion details can be found in the Appendix~\ref{app:kb_ingestion}.

\noindent\textbf{Knowledge search backend.}
We implement knowledge search using a \emph{deterministic} and \emph{hybrid} search backend.
The agent defines an epistemic decision to search: \textbf{what to search} (\iec query) and \textbf{where to search} (\iec knowledge source). The search backend is isolated from agent behavior. Full retrieval pipelines are described in Appendix~\ref{app:kb_retrieval}.

\subsection{Tool definition}
\label{sec:tool_definition}
In \gym, agents are provided with two tool families. The visual grounding tool helps the agent parse the current visual input with general-purpose perception ability; the knowledge tool gives the agent access to surgical knowledge in \wiki. 

\noindent\textbf{Visual grounding tool.}
The \texttt{perceive()} tool helps the agent parse the visual inputs by returning query-free, class-agnostic visual proposals. 
We implement \texttt{perceive()} with SAM~2 automatic mask generation~\cite{ravi2024sam2} with the \texttt{facebook/sam2-hiera-base-plus} model to generate visual proposals that indicate semantic objects and boundaries. 
The returned observation contains a visual Set-of-Mark overlay~\citep{yang2023setofmark} with numeric labels. Details are provided in Appendix~\ref{app:perceive_details}.

\noindent\textbf{Knowledge acquisition tool.}
The \texttt{retrieve()} tool is the agent-facing action for accessing \wiki. Each tool call specifies three arguments: \texttt{query}, the agent-generated natural-language query; \texttt{knowledge\_type}, which records the agent's declared knowledge type (declarative, procedural, conditional, or metacognitive) for trajectory analysis; and \texttt{source\_type}, which selects guidelines, textbooks, tutorial videos, or case records through a hard source-level filter.

\noindent\textbf{Retrieval quality estimate.}
We therefore evaluate the \texttt{retrieve()} tool on real trajectories with a TREC-style graded relevance protocol. Two experts annotate 50 \texttt{retrieve()} calls sampled from pooled real trajectories as a calibration set; GPT-5~\cite{openai2025gpt5} as an LLM judge, scores the full pooled sample with the same rubric. \texttt{retrieve()} returns a relevance score with every item, passed to the agent as part of its observation. This score separates relevant from non-relevant evidence at AUROC = 0.894 against the LLM judge and 0.817 against human expert annotations, and holds within every knowledge source (0.68--0.87). The rubric and agreement statistics are in Appendix~\ref{app:cr_retrieval_quality}.

\section{Evaluating generalist agents for surgical scene understanding}
\label{sec:agent_eval}
We then evaluate how well generalist agents with tool access perform on \bench-Hard.

\subsection{Agent, scaffold, and harness}
\label{sec:agent_harness}
\noindent\textbf{Backbone selection.}
We evaluate 16 MLLM backbones spanning proprietary models, open-source
models across three scale tiers, and medical-domain models. The
proprietary group includes GPT-4o-mini~\citep{openai2024gpt4omini} and
Claude-Haiku-4.5~\citep{anthropic2025claudehaiku45}. The open-source
group is stratified by parameter scale: below 10B parameters, we
evaluate InternVL2.5-8B~\citep{chen2024internvl25},
Qwen2.5-VL-7B-Instruct~\citep{bai2025qwen25vl},
Qwen3-VL-8B-Instruct, and
Qwen3-VL-8B-Thinking~\citep{qwen2025qwen3vl}; between 10B and 30B
parameters, we evaluate Pixtral-12B-2409~\citep{agrawal2024pixtral},
InternVL3-14B~\citep{zhu2025internvl3},
InternVL2.5-26B~\citep{chen2024internvl25}, and
Gemma-3-27B-it~\citep{gemmateam2025gemma3}; above 30B parameters, we
evaluate Qwen2.5-VL-32B-Instruct~\citep{bai2025qwen25vl},
Qwen3-VL-32B-Instruct~\citep{qwen2025qwen3vl},
Qwen2.5-VL-72B-Instruct~\citep{bai2025qwen25vl}, and
InternVL2.5-78B~\citep{chen2024internvl25}. The medical-domain group
includes Lingshu-32B~\citep{lingshu2025} and
MedGemma-27B-it~\citep{medgemma2025}.

\noindent\textbf{Agent harness.}
\gym implements tool use as a universal environment-level protocol shared across backbones. Each episode maintains an explicit state consisting of the task prompt, visual input, observation history, available tools, remaining tool budgets, and the final-answer contract. At each step, the backbone either emits a tool action or submits an answer. The environment parses tool actions and validates them against the tool schema. To prevent unbounded looping over a deterministic backend, \texttt{retrieve()} enforces a source-level call budget. Each \texttt{source\_type} can be queried at most twice per episode, yielding an implicit maximum of eight retrieval calls across the four evidence families. \texttt{perceive()} is limited to two invocations for single-image tasks and three for multi-frame tasks, with duplicate calls removed.
Full runtime constraint specifications are provided in Appendix~\ref{app:runtime_constraints}.

\begin{table}[htbp]
\centering
\caption{Stage~II performance on \textbf{\bench-Hard}.
\textbf{A} compares Stage~I direct answers with Stage~II agents. \textbf{B} compares tools off and on within a matched agent setup. Each cell reports the difference. $^*$ marks a significant change (paired video-cluster bootstrap, BH-FDR $q\leq0.05$); $^\dagger$ marks axes not significance-tested. Green/red indicates positive/negative changes. \textbf{Bold}/\underline{underlining} marks the largest/second-largest change.}
\label{tab:stage_delta_summary}
\begingroup
\tiny
\setlength{\tabcolsep}{6pt}
\newcommand{\sgShade}[1]{\ifdefined\cellcolor\cellcolor[HTML]{#1}\fi}
\begin{tabular}{@{}l ccc cc cc@{}}
\toprule
\multirow{3}{*}{Model}
  & \multicolumn{3}{c}{Spatial Recognition}
  & \multicolumn{2}{c}{Temporal Reasoning}
  & \multicolumn{2}{c}{Workflow Assessment} \\
\cmidrule(lr){2-4} \cmidrule(lr){5-6} \cmidrule(lr){7-8}
 & Tool Det. & Triplet Det.$^{\dagger}$ & Tool Loc.$^{\dagger}$ & Phase Rec. & Action Rec. & CVS Ass. & Skill Ass. \\
 & {\scriptsize $\Delta$F1$\uparrow$} & {\scriptsize $\Delta$F1$\uparrow$} & {\scriptsize $\Delta F_1@0.5\uparrow$}
 & {\scriptsize $\Delta$Acc.$\uparrow$} & {\scriptsize $\Delta$F1$\uparrow$}
 & {\scriptsize $\Delta\kappa\uparrow$} & {\scriptsize $\Delta\kappa\uparrow$} \\
\midrule
\multicolumn{8}{@{}l}{\textbf{A. Across-stage comparison}
  \quad I = direct answer, II = agent with tools} \\
\multicolumn{8}{@{}l}{\textit{Proprietary}} \\
\cmidrule{1-8}
GPT-4o-mini~\citep{openai2024gpt4omini} & \sgShade{D6EFD6}$+3.7$ & \textemdash & \sgShade{D6EFD6}$\mathbf{+5.7}$ & $0.0$ & \sgShade{F7D4D4}$-17.7$ & \sgShade{FBE7E7}$-1.9$ & \sgShade{E9F7E9}$+1.2$ \\
Claude-Haiku-4.5~\citep{anthropic2025claudehaiku45} & \sgShade{E9F7E9}$+1.7$ & \textemdash & $0.0$ & \sgShade{E9F7E9}$+0.8$ & \sgShade{FBE7E7}$-7.6$ & \sgShade{BEE6BE}$\underline{+7.0}^{\ast}$ & \sgShade{FBE7E7}$-4.7$ \\
\multicolumn{8}{@{}l}{\textit{Open-source ($<10$B)}} \\
\cmidrule{1-8}
InternVL2.5-8B~\citep{chen2024internvl25} & \sgShade{D6EFD6}$+6.9^{\ast}$ & \sgShade{E9F7E9}$+0.3$ & \sgShade{F7D4D4}$-10.6$ & \sgShade{E9F7E9}$+0.9^{\ast}$ & \sgShade{C9EAC9}$+18.0^{\ast}$ & \sgShade{F7D4D4}$-4.8$ & \sgShade{BEE6BE}$\mathbf{+13.0}$ \\
Qwen2.5-VL-7B-Instruct~\citep{bai2025qwen25vl} & \sgShade{D6EFD6}$+3.9^{\ast}$ & \textemdash & \sgShade{FCF0F0}$-0.9$ & \sgShade{D6EFD6}$+3.1^{\ast}$ & \sgShade{E9F7E9}$+3.0^{\ast}$ & \sgShade{FBE7E7}$-4.3$ & \sgShade{F5C8C8}$-12.7$ \\
Qwen3-VL-8B-Instruct~\citep{qwen2025qwen3vl} & \sgShade{D6EFD6}$+6.2^{\ast}$ & \textemdash & \sgShade{E9F7E9}$+1.6$ & \sgShade{C9EAC9}$+3.7^{\ast}$ & \sgShade{BEE6BE}$\underline{+27.3}^{\ast}$ & \sgShade{FCF0F0}$-1.8$ & \sgShade{FCF0F0}$-2.1$ \\
Qwen3-VL-8B-Thinking~\citep{qwen2025qwen3vl} & \sgShade{F7D4D4}$-5.1^{\ast}$ & \textemdash & \sgShade{FBE7E7}$-5.8$ & \sgShade{E9F7E9}$+1.7^{\ast}$ & \sgShade{D6EFD6}$+13.5^{\ast}$ & \sgShade{FBE7E7}$-3.5$ & \sgShade{C9EAC9}$\underline{+7.5}$ \\
\multicolumn{8}{@{}l}{\textit{Open-source (10--30B)}} \\
\cmidrule{1-8}
Pixtral-12B-2409~\citep{agrawal2024pixtral} & \sgShade{BEE6BE}$\underline{+13.5}^{\ast}$ & \textemdash & \sgShade{D6EFD6}$\underline{+4.9}$ & \sgShade{D6EFD6}$+2.6^{\ast}$ & \sgShade{BEE6BE}$\mathbf{+27.6}^{\ast}$ & \sgShade{FBE7E7}$-1.6$ & \sgShade{D6EFD6}$+6.5$ \\
InternVL3-14B~\citep{zhu2025internvl3} & \sgShade{C9EAC9}$+8.3^{\ast}$ & \sgShade{E9F7E9}$+2.1$ & \sgShade{E9F7E9}$+2.4$ & \sgShade{E9F7E9}$+2.3^{\ast}$ & \sgShade{E9F7E9}$+4.2^{\ast}$ & \sgShade{E9F7E9}$+1.0$ & \sgShade{E9F7E9}$+0.8$ \\
InternVL2.5-26B~\citep{chen2024internvl25} & \sgShade{D6EFD6}$+2.4$ & \sgShade{E9F7E9}$+1.5$ & \sgShade{F7D4D4}$-9.7$ & \sgShade{E9F7E9}$+1.4^{\ast}$ & \sgShade{E9F7E9}$+4.4^{\ast}$ & \sgShade{E9F7E9}$+1.4$ & \sgShade{FCF0F0}$-0.9$ \\
Gemma-3-27B-it~\citep{gemmateam2025gemma3} & \sgShade{E9F7E9}$+1.6$ & \sgShade{E9F7E9}$+2.1$ & \sgShade{D6EFD6}$\mathbf{+5.7}$ & \sgShade{D6EFD6}$+3.1^{\ast}$ & \sgShade{D6EFD6}$+12.8^{\ast}$ & \sgShade{FBE7E7}$-1.5$ & \sgShade{FBE7E7}$-4.4$ \\
\multicolumn{8}{@{}l}{\textit{Open-source ($>30$B)}} \\
\cmidrule{1-8}
Qwen2.5-VL-32B-Instruct~\citep{bai2025qwen25vl} & \sgShade{C9EAC9}$+9.2^{\ast}$ & \sgShade{BEE6BE}$\mathbf{+7.7}$ & $0.0$ & \sgShade{BEE6BE}$\mathbf{+7.4}^{\ast}$ & \sgShade{BEE6BE}$+25.1^{\ast}$ & \sgShade{D6EFD6}$+3.7$ & \sgShade{FBE7E7}$-4.6$ \\
Qwen3-VL-32B-Instruct~\citep{qwen2025qwen3vl} & \sgShade{D6EFD6}$+4.9^{\ast}$ & \textemdash & \sgShade{D6EFD6}$+3.2$ & \sgShade{C9EAC9}$\underline{+5.1}^{\ast}$ & \sgShade{BEE6BE}$+26.5^{\ast}$ & \sgShade{FCF0F0}$-0.5$ & \sgShade{F5C8C8}$-15.1$ \\
Qwen2.5-VL-72B-Instruct~\citep{bai2025qwen25vl} & \sgShade{D6EFD6}$+5.3^{\ast}$ & \textemdash & $0.0$ & \sgShade{C9EAC9}$+4.9^{\ast}$ & \sgShade{BEE6BE}$+25.6^{\ast}$ & \sgShade{BEE6BE}$\mathbf{+7.5}$ & \sgShade{FBE7E7}$-2.8$ \\
InternVL2.5-78B~\citep{chen2024internvl25} & \sgShade{E9F7E9}$+1.1$ & \sgShade{D6EFD6}$\underline{+3.8}$ & \sgShade{F5C8C8}$-19.5$ & \sgShade{E9F7E9}$+1.4^{\ast}$ & \sgShade{D6EFD6}$+9.2^{\ast}$ & \sgShade{FBE7E7}$-2.3$ & \sgShade{FBE7E7}$-6.4$ \\
\multicolumn{8}{@{}l}{\textit{Medical-domain}} \\
\cmidrule{1-8}
Lingshu-32B~\citep{lingshu2025} & \sgShade{BEE6BE}$\mathbf{+14.8}^{\ast}$ & \textemdash & \sgShade{D6EFD6}$\underline{+4.9}$ & \sgShade{C9EAC9}$+4.9^{\ast}$ & \sgShade{BEE6BE}$+26.7^{\ast}$ & \sgShade{E9F7E9}$+1.3$ & \sgShade{FBE7E7}$-2.5$ \\
MedGemma-27B-it~\citep{medgemma2025} & \sgShade{C9EAC9}$+9.2^{\ast}$ & \sgShade{E9F7E9}$+1.0$ & \sgShade{D6EFD6}$+4.0$ & \sgShade{D6EFD6}$+3.1^{\ast}$ & \sgShade{BEE6BE}$+24.0^{\ast}$ & \sgShade{FBE7E7}$-4.2$ & \sgShade{F7D4D4}$-10.5$ \\
\midrule
\multicolumn{8}{@{}l}{\textbf{B. Matched tool comparison}
  \quad Off = tools disabled, On = tools enabled} \\
\cmidrule{1-8}
InternVL2.5-8B & $+1.9$ & $+0.6$ & $+8.1$ & $+0.3$ & $+14.0$ & $+6.3$ & $-1.7$ \\
Qwen2.5-VL-7B-Instruct & $+1.4$ & $0.0$ & $+10.9$ & $+2.6$ & $+4.6$ & $-5.0$ & $-2.1$ \\
Qwen3-VL-8B-Instruct & $+4.6$ & $0.0$ & $+1.7$ & $+3.7$ & $+26.2$ & $-13.5$ & $+5.6$ \\
Gemma-3-27B-it & $+4.1$ & $+3.0$ & $+4.9$ & $+1.7$ & $+6.6$ & $+2.5$ & $+4.5$ \\
Lingshu-32B & $+9.7$ & $-2.2$ & $-2.4$ & $+2.3$ & $+16.5$ & $+0.9$ & $-5.7$ \\
\bottomrule
\end{tabular}
\endgroup
\end{table}

\subsection{Main results: do tool-mediated agents close the gap?}
We first examine how generalist agents perform on the difficult episodes in \bench-Hard. 
Tool-mediated inference produces task-specific gains rather than uniform improvements in surgical understanding.
Tab.~\ref{tab:stage_delta_summary} reports two comparisons. Panel~A measures the overall change from Stage~I direct answers to the Stage~II agent setting. Panel~B compares tools off and on under a matched agent harness, including prompts, to assess the contribution of tool access.
Statistical analysis is provided in Appendix~\ref{app:cr_statistical_analysis}. Appendix~\ref{app:tool_configuration} extends ablation studies.

\noindent\textbf{General performance.}
The full agent setting improves tool and action recognition for most evaluated models, but the gains vary across tasks
and backbones. Tool detection improves for 15 of the 16 models and action recognition improves for all 14 open-source models,
whereas GPT-4o-mini and Claude-Haiku-4.5 decline by 17.7 and 7.6 macro-F1 points, respectively.
Phase recognition shows smaller, predominantly positive changes. These gains do not extend uniformly to localization or workflow assessment: localization declines across the InternVL2.5 family, while CVS and skill assessment show mixed changes across models. This reveals limitations similar to those observed in Stage I.
\emph{The gains also need to be considered alongside the attained performance.} In the Full setting evaluated in Panel~A, the
highest action-recognition score is \(27.6\%\) macro-F1, still below the majority baseline of \(30.0\%\).
Thus, even a task with substantial gains remains difficult in absolute terms. 
This emphasizes a prevailing frontier for generalist agents to solve domain-expertise surgical tasks.
Detailed absolute scores and per-dataset results are provided in Appendix~\ref{app_stage2}.

\noindent\textbf{Benefits of tool access.}
In Panel~B, we match the prompt and compare the net performance change caused by tool access.
The matched comparison confirms positive tool-detection and action-recognition gains across all five evaluated backbones,
ranging from \(+1.4\) to \(+9.7\) and from \(+4.6\) to \(+26.2\) macro-F1 points, respectively.
Phase recognition also improves across all five backbones, although by smaller margins.
Localization improves for four of the five backbones, showing that the declines observed across stages do not
necessarily imply a negative contribution from tool access.
For example, InternVL2.5-8B loses 10.6 localization points in Panel~A but gains 8.1 points when tools are enabled in the matched setup.
CVS and skill assessment, by contrast, still present a mixed signal with improvements for some backbones and deterioration for others: tool access does not provide a consistent benefit for rubric-based judgments. Mere tool access cannot solve subjective tasks.

\subsection{Trajectory analysis}
\label{sec:trajectory_analysis}
We analyze trajectories from our experiments to examine how agents use tools to solve surgical tasks. We focus on the following questions.

\begin{figure}[htbp]
      \centering
      \includegraphics[width=1\linewidth]{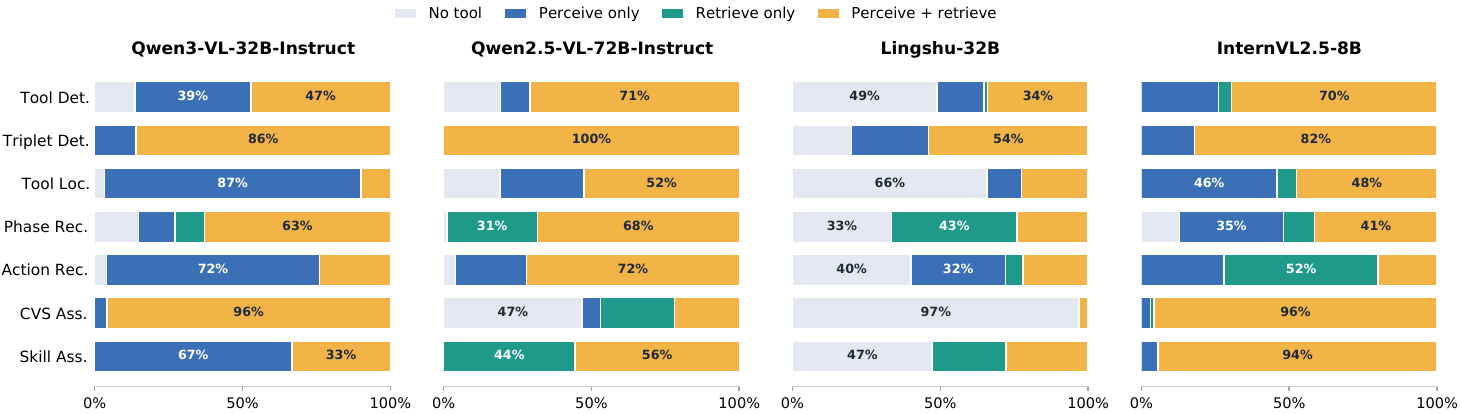}
      \caption{Full-setting tool-use patterns across four representative agents on \bench-Hard.}
      \label{fig:traj_tool_use_pattern}
  \end{figure}

\noindent\textbf{Q1: Which tools do agents use across tasks?}
Fig.~\ref{fig:traj_tool_use_pattern} shows task-dependent tool routing, with frequent visual grounding
in spatial recognition and retrieval in phase recognition. This is consistent with our intuition that agents should use external tools to enhance their visual grounding abilities.
Agents invoke \texttt{perceive()} in 78.2\% of tool-detection episodes, 95.0\% of triplet-detection episodes, and 76.2\%
of tool-localization episodes. For phase recognition, agents invoke \texttt{retrieve()} in 72.6\% of episodes.
Fig.~\ref{fig:traj_retrieve_source} shows that evidence-source selection in \texttt{retrieve()} also varies by task. Tutorial videos dominate retrieval for action recognition, accounting for 80.3\% of retrieval calls. They also constitute the majority of retrieval calls for triplet detection (62.2\%), tool localization (53.6\%),
and phase recognition (51.9\%).
Workflow assessment shows a different source preference: textbooks and guidelines together account for 80.8\% of CVS retrieval calls and 66.9\% of skill-assessment retrieval calls.
This shows the agents' tendency to seek temporal context for temporal tasks and procedural knowledge for workflow-level tasks.

\noindent\textbf{Q2: Do tool results actually help the agents?}
Tool results do not necessarily provide usable evidence for the task because of imperfect tool outputs. We investigate whether tool results actually help the agents answer the questions.
For \texttt{retrieve()}, we evaluate results from real agent trajectories using a graded relevance rubric covering topical match, constraint satisfaction, and information coverage. Two surgeon experts annotate a calibration sample of 50 retrieval calls, and an LLM judge evaluates the pooled sample using the same rubric. The reported precision@1 estimates range from 57\% to 82\%, indicating that retrieved items frequently address the agents' generated queries, but relevance cannot be assumed for every call. The retrieval relevance score achieves an AUROC of 0.817 against expert annotations. These results indicate that retrieved items are adequately relevant for the queries. The evaluation protocol and agreement statistics are provided in Appendix~\ref{app:cr_retrieval_quality}. Thus, a failure case is more attributed to that the generated query itself rather than the retrieval pipeline.
\emph{Visual grounding might also return less usable evidence for agents.} In the action-recognition trajectories of GPT-4o-mini and Claude-Haiku-4.5, \texttt{perceive()} returns no usable region proposals in 90\%--100\% of trajectories. These calls therefore provide no additional region-level evidence for recognizing the action, despite the agents having proactively invoked a visual grounding tool. This exposes the necessity of domain-specific tools beyond domain-specific agents.

\noindent\textbf{Q3: Tool use does not guarantee success.}
Agents actively use tools even in failed episodes.
Tab.~\ref{tab:traj_success_conditioned} shows that successful
and failed episodes average 2.32 and 2.36 tool calls,
respectively. Their tool-use profiles differ:
\texttt{perceive()} is more frequent in failed episodes
(69.8\% versus 56.4\%), whereas \texttt{retrieve()} is
more frequent in successful episodes (71.8\% versus 65.0\%).
The negative association between using both tools and
success largely disappears after task-centering,
while retrieval use becomes positively associated with
success. These patterns motivate targeted post-training
of tool-use policies that select tools according to
the task and the agent's information needs.

\subsection{Why do agents fail?}
To sum up, we identify agent bottlenecks at three stages of the epistemic process:
formulating tool queries, integrating imperfect evidence, and grounding retrieved evidence in the final answer.

\noindent\textbf{Does the agent proactively seek evidence relevant to the question? (Query quality).} We show that evidence needs are modality-specific: for perception-heavy tasks, visual exemplars are often more direct than abstract textbook passages (much as for human trainees). Yet agents show a preference for authoritative text, querying textbooks for a "golden answer" where a visual case would serve better. In the trajectory analysis, tutorial videos account for 80.3\% of retrieval calls in action recognition, while textbooks and guidelines account for 80.8\% in CVS assessment. Task framing and query generation are metacognitive skills acquired through agentic learning like post-training or self-evolution.

\noindent\textbf{Can the agent withstand imperfect evidence? (Agent robustness).}
Retrieved evidence can often be partially irrelevant or noisy, and in most failure cases agents do not withstand it. In CVS episodes, retrieved guideline text can even override correct visual understanding. In the two proprietary models’ action-recognition trajectories, \texttt{perceive()} returns no usable region proposals in 90–100\% of trajectories. GPT-4o-mini’s recall for the dominant action, “hold,” falls from 0.76 under direct inference to 0.07 in the agent setting. Robustness to imperfect or noisy evidence is a core capability that current models lack. Explicit uncertainty and credibility quantification and calibration can help agents remain robust to misleading evidence.

\noindent\textbf{Can the agent leverage useful evidence? (Evidence grounding).} Even perfectly relevant evidence must be integrated into the model’s reasoning. Under heavy tool use the accumulated context itself becomes a burden, and useful evidence is retrieved but ignored. In our trajectory analysis, failed episodes invoke \texttt{perceive()} more often than successful episodes (69.8\% vs. 56.4\%) and use both tools slightly more often (53.0\% vs. 48.2\%). Evidence grounding is the binding constraint.

\section{Related work}
\label{sec:related}
\noindent\textbf{Surgical superintelligence.}
Surgical AI has advanced task-specific scene understanding on curated specific datasets~\cite{nwoye2022rendezvous,allan2019endovis,wagner2023heichole}, while recent surgical MLLMs extend supervision through video-text pretraining, instruction tuning, VQA, scene graphs, and multi-task benchmarks~\cite{yuan2024hecvl,li2024llavasurg,yuan2024ssgvqa,wei2025surgbench,zeng2025surgvlm,perez2026sureon,li2026surgpubvideo,zeng2026surgsigma}. Surgical foundation models further scale representation learning across surgery procedures~\cite{pezold2025vjepa,wu2026unisurg,park2026zen,tu2026ovfm,schmidgall2024gpvls}. 

\noindent\textbf{Medical agent benchmarks.}
Medical agent benchmarks such as AgentClinic, MedAgentBench, and MMedAgent evaluate general clinical agents with action or diagnostic tools, but do not target surgical visual scene understanding or attribution of knowledge acquisition and visual grounding~\cite{schmidgall2024agentclinic,jiang2025medagentbench,mmedagent}. 
Closer to surgery, SurgRAW and CARES study agentic reasoning for robotic surgical video analysis and surgical error detection~\cite{low2026surgraw,low2025cares}.

\section{Conclusion}
We presented a two-stage evaluation of surgical competence for general MLLMs. First, we examined what these models can infer directly from surgical scenes without domain-specific adaptation. Despite strong general-purpose capabilities, current MLLMs show only limited surgical understanding, particularly beyond object-level recognition toward relational, procedural, and assessment tasks.
We then evaluated whether tool-augmented agents can compensate for these knowledge gaps by retrieving external information and grounding visual evidence. Agentic inference improves some tasks, showing that generalist MLLMs possess epistemic capabilities that allow them to seek missing domain knowledge proactively. However, our analysis also reveals that these capabilities remain imperfect: agents may formulate the wrong queries, fail to reconcile imperfect evidence, or fail to ground evidence in reasoning. Our findings therefore suggest that MLLM agents could potentially bridge their domain knowledge gap through epistemic reasoning and active tool-use, but these metacognitive capabilities themselves are not taken for granted and still remain an important challenge for generalist agents in the surgical domain.

\noindent\textbf{Limitations.}
This work has several limitations. (1) \bench harmonizes public surgical datasets, so its coverage is bounded by the procedures, institutions, camera views, annotation schemes, and task definitions available in these sources.  (2) \wiki is a closed and provenance-tracked evidence corpus, but its contents are still limited by source availability and licensing.

\newpage
\begin{ack}
\begingroup
\setlength{\intextsep}{0pt}
\setlength{\columnsep}{1em}
\begin{wrapfigure}{r}{0.26\textwidth}
\includegraphics[width=\linewidth]{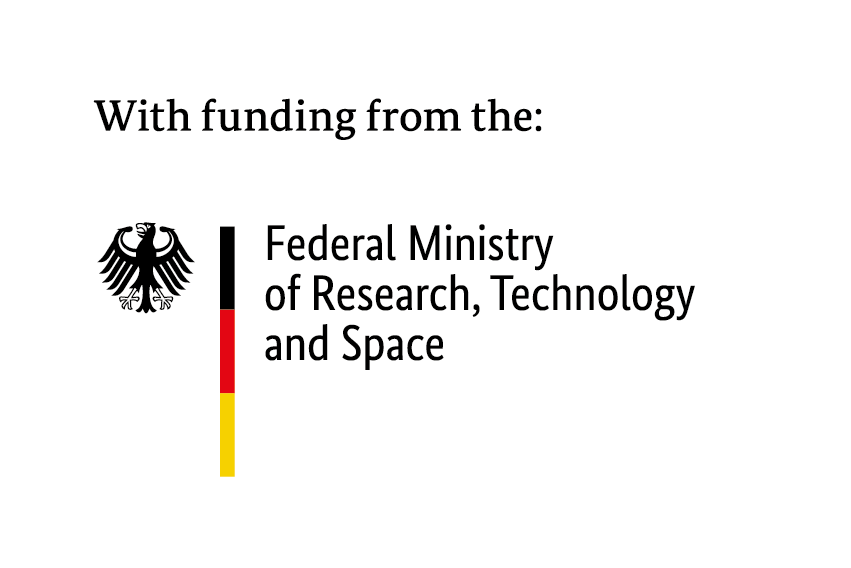}
\end{wrapfigure}
The project on which this report is based was sponsored by the Federal Ministry of Research, Technology and Space (BMFTR) under the funding code 16IS26022. Responsibility for the content of this publication lies with the authors.
We thank Feng Jin for providing expert advice and assistance with data curation. We also thank Xinling Xie for assistance.
The authors gratefully acknowledge the scientific support and HPC resources provided by the Erlangen National High Performance Computing Center (NHR@FAU) of the Friedrich-Alexander-Universität Erlangen-Nürnberg (FAU) under the NHR project b306dd. NHR funding is provided by federal and Bavarian state authorities.
We acknowledge the EuroHPC Joint Undertaking for awarding this project access to the EuroHPC supercomputer LEONARDO, hosted by CINECA (Italy) and the LEONARDO consortium, through a EuroHPC Development Access call (project EHPC-DEV-2026D02-157).
\par
\endgroup
\end{ack}

\bibliographystyle{unsrtnat}
\bibliography{references}

\medskip

\clearpage
\appendix
\startcontents[app] %

\section*{Appendix}
\vspace{1.2em}
{\bfseries Table of contents\par}
\vspace{0.4em}
\hrule
\vspace{1.0em}

\AppendixTOCStyle
\setcounter{tocdepth}{2} 
\printcontents[app]{l}{1}{} 

\vspace{0.6em}
\hrule
\AddToHook{cmd/section/before}[appendix-float-order]{\FloatBarrier}
\AddToHook{cmd/subsection/before}[appendix-float-order]{\FloatBarrier}
\AddToHook{cmd/subsubsection/before}[appendix-float-order]{\FloatBarrier}
\AddToHook{env/figure/before}[appendix-float-order]{\FloatBarrier}
\AddToHook{env/figure*/before}[appendix-float-order]{\FloatBarrier}
\AddToHook{env/figure/after}[appendix-float-order]{\FloatBarrier}
\AddToHook{env/figure*/after}[appendix-float-order]{\FloatBarrier}

\newpage
\section{Societal impact}
\label{app:societal_impact}

This work supports safer surgical AI research by providing reproducible benchmarks and traceable evidence-use trajectories. 
\bench may also aid medical education, but the artifacts are intended for research evaluation, not autonomous clinical decision-making.

\section{Source datasets}
\label{app:benchmark}

Our benchmark is built from eleven public surgical datasets spanning seven surgical
procedures, from which we instantiate seven tasks grouped into three task types.
Section~\ref{app:benchmark:datasets} describes each source dataset and the tasks we
draw from it; Section~\ref{app:benchmark:tasks} describes the prediction tasks and
answer formats. The dataset--task inventory is given in
Tab.~\ref{tab:surgknowbench_inventory}.

Throughout, we evaluate \emph{frozen} general-purpose multimodal language models
in a zero-shot manner, with \textbf{no task-specific training or fine-tuning}.
Each episode pairs its visual input with a task prompt and a structured answer
contract. Visual-input handling, output schemas, and canonical task prompts are
described in Appendices~\ref{app:frame_sampling}, \ref{app:task_schema},
and~\ref{app:task-templates}, respectively.

\subsection{Datasets}
\label{app:benchmark:datasets}

\paragraph{Cholec80.}
\begin{itemize}[leftmargin=1.4em,itemsep=2pt,topsep=2pt]
  \item \textbf{Description.} Cholec80~\cite{twinanda2017endonet} is a public CAMMA
        dataset (University of Strasbourg) of \textbf{80} laparoscopic cholecystectomy
        videos recorded at 25~fps, with every frame annotated for the surgical phase and
        tool presence.
  \item \textbf{Surgery type.} Laparoscopic cholecystectomy
  \item \textbf{Tasks.} Phase recognition.
\end{itemize}

\paragraph{Cholec80-CVS.}
\begin{itemize}[leftmargin=1.4em,itemsep=2pt,topsep=2pt]
  \item \textbf{Description.} Cholec80-CVS~\cite{rios2023cholec80cvs} augments the same
        Cholec80 videos with expert annotations of Strasberg's Critical View of Safety
        (CVS), scored on three criteria: two structures, cystic plate, and hepatocystic
        triangle.
  \item \textbf{Surgery type.} Laparoscopic cholecystectomy 
  \item \textbf{Tasks.} CVS assessment.
\end{itemize}

\paragraph{CholecT50.}
\begin{itemize}[leftmargin=1.4em,itemsep=2pt,topsep=2pt]
  \item \textbf{Description.} CholecT50~\cite{nwoye2022rendezvous} is a CAMMA dataset of
        \textbf{50} laparoscopic cholecystectomy videos annotated for fine-grained
        surgical activity as action triplets $\langle$instrument, verb, target$\rangle$
        (6 instruments, 10 verbs, 15 targets, 100 triplet classes).
  \item \textbf{Surgery type.} Laparoscopic cholecystectomy 
  \item \textbf{Tasks.} Tool detection; tool localization; triplet detection.
\end{itemize}

\paragraph{Cataract-101.}
\begin{itemize}[leftmargin=1.4em,itemsep=2pt,topsep=2pt]
  \item \textbf{Description.} Cataract-101~\cite{schoeffmann2018cataract101} is a public
        dataset from Klagenfurt University of \textbf{101} cataract-surgery videos (four
        surgeons, two expertise levels) recorded at 25~fps and annotated with 10 surgical
        phases.
  \item \textbf{Surgery type.} Cataract surgery / phacoemulsification (ophthalmology).
  \item \textbf{Tasks.} Phase recognition.
\end{itemize}

\paragraph{HeiChole.}
\begin{itemize}[leftmargin=1.4em,itemsep=2pt,topsep=2pt]
  \item \textbf{Description.} HeiChole~\cite{wagner2023heichole} is a multicentric
        benchmark of laparoscopic cholecystectomy from the EndoVis~2019 sub-challenge:
        \textbf{33} videos from three centres, annotated framewise for surgical phases,
        four surgical actions, instrument presence, and five skill dimensions.
  \item \textbf{Surgery type.} Laparoscopic cholecystectomy (hepatobiliary / general
        surgery).
  \item \textbf{Tasks.} Action recognition; skill assessment.
\end{itemize}

\paragraph{MultiBypass140.}
\begin{itemize}[leftmargin=1.4em,itemsep=2pt,topsep=2pt]
  \item \textbf{Description.} MultiBypass140~\cite{lavanchy2024multibypass140} is a
        multicentric benchmark of laparoscopic Roux-en-Y gastric bypass: \textbf{140}
        videos at 25~fps from two centres (StrasBypass70, Strasbourg; BernBypass70,
        Bern), annotated with 12 phases, 46 steps, and 5 intra-operative adverse events.
  \item \textbf{Surgery type.} Laparoscopic Roux-en-Y gastric bypass (bariatric / general
        surgery).
  \item \textbf{Tasks.} Phase recognition.
\end{itemize}

\paragraph{PitVis.}
\begin{itemize}[leftmargin=1.4em,itemsep=2pt,topsep=2pt]
  \item \textbf{Description.} PitVis~\cite{das2025pitvis} is the dataset of the PitVis-2023
        Challenge (an Endoscopic Vision sub-challenge at MICCAI~2023): \textbf{25} videos
        of endoscopic pituitary surgery at 24~fps, the first public dataset with both
        surgical-step and instrument annotations.
  \item \textbf{Surgery type.} Endoscopic pituitary surgery, transsphenoidal approach
        (neurosurgery).
  \item \textbf{Tasks.} Phase recognition; tool detection.
\end{itemize}

\paragraph{GraSP.}
\begin{itemize}[leftmargin=1.4em,itemsep=2pt,topsep=2pt]
  \item \textbf{Description.} GraSP~\cite{ayobi2025tapis} is a multi-granular benchmark of
        robot-assisted radical prostatectomy: \textbf{13} da~Vinci videos (Fundaci\'on
        Santa F\'e de Bogot\'a) annotated for surgical phases, steps, instruments, and
        atomic actions.
  \item \textbf{Surgery type.} Robot-assisted radical prostatectomy (urology).
  \item \textbf{Tasks.} Phase recognition; tool detection.
\end{itemize}

\paragraph{AutoLaparo.}
\begin{itemize}[leftmargin=1.4em,itemsep=2pt,topsep=2pt]
  \item \textbf{Description.} AutoLaparo~\cite{wang2022autolaparo} is a public multi-task
        dataset from CUHK for image-guided automation in laparoscopic hysterectomy:
        \textbf{21} videos at 25~fps with annotations for surgical phase, motion
        prediction, and instrument/keypoint tracking.
  \item \textbf{Surgery type.} Laparoscopic hysterectomy (gynaecology).
  \item \textbf{Tasks.} Phase recognition; tool detection.
\end{itemize}

\paragraph{HeiCo.}
\begin{itemize}[leftmargin=1.4em,itemsep=2pt,topsep=2pt]
  \item \textbf{Description.} HeiCo~\cite{maierhein2021heico} provides laparoscopic
        colorectal surgery videos with workflow annotations.
  \item \textbf{Surgery type.} Laparoscopic proctocolectomy.
  \item \textbf{Tasks.} Phase recognition.
\end{itemize}

\paragraph{CATARACTS.}
\begin{itemize}[leftmargin=1.4em,itemsep=2pt,topsep=2pt]
  \item \textbf{Description.} CATARACTS~\cite{alhajj2019cataracts} provides cataract
        surgery videos with surgical-instrument annotations.
  \item \textbf{Surgery type.} Cataract surgery.
  \item \textbf{Tasks.} Tool detection.
\end{itemize}

\subsection{Tasks}
\label{app:benchmark:tasks}

We group the seven tasks into three task types: \textbf{Spatial Recognition}
(tool detection, tool localization, triplet detection), \textbf{Temporal Reasoning}
(phase recognition, action recognition), and \textbf{Workflow Assessment}
(CVS assessment, skill assessment). Frame-level labels and interval-level rubric
scores retain their respective annotation units; an interval's timestamps do not
imply that every frame in that interval is provided to the model.

\paragraph{Tool detection.}
\begin{itemize}[leftmargin=1.4em,itemsep=2pt,topsep=2pt]
  \item \textbf{Description.} Identify every distinct surgical-instrument type
        visible in the supplied image. Several instrument types may be present.
  \item \textbf{Task type.} Spatial Recognition.
  \item \textbf{Visual input.} An annotated frame and its dataset-specific
        instrument vocabulary.
  \item \textbf{Answer.} An unordered set of instrument-label strings, represented
        as a JSON array. Each type is listed once; an empty array is valid when
        no instrument is visible.
\end{itemize}

\paragraph{Tool localization.}
\begin{itemize}[leftmargin=1.4em,itemsep=2pt,topsep=2pt]
  \item \textbf{Description.} Identify and localize the instrument in a
        box-annotated image.
  \item \textbf{Task type.} Spatial Recognition.
  \item \textbf{Visual input.} A box-annotated frame.
  \item \textbf{Answer.} An instrument label and bounding box under the
        model-family-specific grounding contract. The evaluator normalizes these
        outputs into labeled image coordinates; the contracts are given in
        Appendix~\ref{app:task-templates}.
\end{itemize}

\paragraph{Triplet detection.}
\begin{itemize}[leftmargin=1.4em,itemsep=2pt,topsep=2pt]
  \item \textbf{Description.} Identify the surgical interactions visible in an
        annotated CholecT50 frame, expressed as
        $\langle$instrument, verb, target$\rangle$ triplets. Multiple triplets
        may co-occur.
  \item \textbf{Task type.} Spatial Recognition.
  \item \textbf{Visual input.} An annotated frame and the instrument, verb,
        and anatomical-target vocabularies.
  \item \textbf{Answer.} An unordered set of complete triplet strings, represented
        as a JSON array. An empty array is valid when no interaction is visible.
\end{itemize}

\paragraph{Phase recognition.}
\begin{itemize}[leftmargin=1.4em,itemsep=2pt,topsep=2pt]
  \item \textbf{Description.} Identify the surgical phase or dataset-specific
        workflow step shown in the supplied annotated frame.
  \item \textbf{Task type.} Temporal Reasoning.
  \item \textbf{Visual input.} An annotated frame with the procedure description
        and the dataset-specific phase or step vocabulary.
  \item \textbf{Answer.} One canonical label string copied from the supplied
        vocabulary. The task is scored against the corresponding frame-level label.
\end{itemize}

\paragraph{Action recognition.}
\begin{itemize}[leftmargin=1.4em,itemsep=2pt,topsep=2pt]
  \item \textbf{Description.} Identify every distinct surgical action visible
        in the supplied HeiChole frame. Multiple action labels may apply.
  \item \textbf{Task type.} Temporal Reasoning.
  \item \textbf{Visual input.} An annotated frame with the procedure description
        and the action vocabulary supplied by the task prompt.
  \item \textbf{Answer.} An unordered set of action-label strings, represented
        as a JSON array and scored against the frame-level annotation. An empty
        array is valid when no action is visible.
\end{itemize}

\paragraph{CVS assessment.}
\begin{itemize}[leftmargin=1.4em,itemsep=2pt,topsep=2pt]
  \item \textbf{Description.} Assess Strasberg's three Critical View of Safety
        criteria: two structures, cystic plate, and hepatocystic triangle.
  \item \textbf{Task type.} Workflow Assessment.
  \item \textbf{Visual input.} One or more representative images associated with
        the annotated interval, in temporal order. The prompt asks the model to
        assess the interval jointly from the provided images.
  \item \textbf{Answer.} A JSON dictionary with keys \texttt{c1}, \texttt{c2},
        and \texttt{c3}, each taking an ordinal value in $\{0,1,2\}$. The
        criterion-specific rubric is given in Appendix~\ref{app:task-templates}.
\end{itemize}

\paragraph{Skill assessment.}
\begin{itemize}[leftmargin=1.4em,itemsep=2pt,topsep=2pt]
  \item \textbf{Description.} Rate surgical execution over the annotated interval
        using the GOALS-style rubric used in HeiChole. We retain the holistic,
        Calot-phase, and dissection-phase assessment sub-axes.
  \item \textbf{Task type.} Workflow Assessment.
  \item \textbf{Visual input.} The representative visual inputs associated with
        the whole-procedure or phase-segment episode, subject to its frame budget.
  \item \textbf{Answer.} Five ordinal scores from 1 to 5, returned as a JSON
        dictionary with keys \texttt{depth\_perception},
        \texttt{bimanual\_dexterity}, \texttt{efficiency},
        \texttt{tissue\_handling}, and \texttt{case\_difficulty}.
\end{itemize}

\section{\bench construction}

\subsection{Dataset harmonization}
\label{app:dataset_harmonization}

The source datasets in \bench differ in annotation units, label vocabularies, and metadata conventions. We harmonize them by
converting each source annotation into a common episode record while preserving the original supervision target. Each record
stores the dataset identity, procedure type, split, video or case identifier, temporal location, task type, visual-input
references, and structured ground truth.

Harmonization is performed at the dataset--task level. Frame-level annotations, such as tool presence, phase labels, and
instrument boxes, are converted into single-frame episodes. Interval-level annotations, such as CVS and GOALS-style skill
assessment, are converted into interval episodes with start/end timestamps and endpoint visual references. Dataset-specific
labels are mapped to the canonical strings used by the corresponding task template and evaluator, while source metadata is
retained for per-dataset reporting, hard-set construction, and leakage checks.

This step standardizes the evaluation substrate without redefining the scoring rules. The answer schemas are specified separately
in Appendix~\ref{app:task_schema}, frame sampling is specified in Appendix~\ref{app:frame_sampling}, and metric definitions are
specified in Appendix~\ref{app:metrics}.

\noindent\textbf{Annotation agreement.}
To document the reliability evidence available for the inherited annotations, Tab.~\ref{tab:cr_rebuttal_source_iaa} summarizes inter-annotator agreement reported by the source datasets. These entries reflect each source's annotation protocol.

\begin{table}[htbp]
\centering
\scriptsize
\setlength{\tabcolsep}{3pt}
\renewcommand{\arraystretch}{1.15}
\caption{Inter-annotator agreement reports in the source datasets.}
\label{tab:cr_rebuttal_source_iaa}
\begin{tabular}{@{}>{\raggedright\arraybackslash}p{0.18\linewidth} >{\raggedright\arraybackslash}p{0.24\linewidth} >{\raggedright\arraybackslash}p{0.11\linewidth} >{\raggedright\arraybackslash}p{\dimexpr0.47\linewidth-6\tabcolsep\relax}@{}}
\toprule
Dataset & Annotators & IAA reported? & Metric / value \\
\midrule
Cholec80~\citep{twinanda2017endonet} & 1 senior surgeon & No & Single annotator; none \\
CholecT50~\citep{nwoye2022rendezvous} & 2 surgeons + 3rd-clinician mediation & No & Consensus only, no metric \\
Cholec80-CVS~\citep{rios2023cholec80cvs} & 4 (2 surgeons + 2 residents) & No & Discussion consensus, no metric \\
HeiChole~\citep{wagner2023heichole} & Phase: 3 students / Skill: 2 students & Partial & Phase: Fleiss' $\kappa$ $0.92\pm0.06$ (train) / $0.95\pm0.04$ (test); Skill (GOALS): none \\
MultiBypass140~\citep{lavanchy2024multibypass140} & 2 board-certified surgeons & Yes & Cohen's $\kappa$ 96\% (phase) / 81\% (step) --- cited from group's prior ontology work \\
AutoLaparo~\citep{wang2022autolaparo} & 2 gynaecologists & No & Proofread / cross-checked, no metric \\
HeiCo~\citep{maierhein2021heico} & 1 resident + 5 expert reviewers & Yes & Phase: Cohen's $\kappa$ intra 0.834, inter 0.68--0.79, Fleiss' $\kappa$ 0.712; segmentation inter-annotator DSC 0.96 \\
GraSP~\citep{ayobi2025tapis} & 2 residents + 17 annotators & No & Independent double-annotation + consensus, no metric \\
PitVis~\citep{das2025pitvis} & 2 trainees + 1 consultant & No & Expert-confirmed consensus, no metric \\
Cataract-101~\citep{schoeffmann2018cataract101} & 1 ophthalmologist & No & Single annotator; none \\
CATARACTS~\citep{alhajj2019cataracts} & 2 non-MD experts & Yes* & Per-tool percent agreement 0.630--0.998 (*not chance-corrected; no kappa) \\
\bottomrule
\end{tabular}
\end{table}

  \subsection{Frame sampling and visual inputs}
  \label{app:frame_sampling}

  \bench contains both frame-level and interval-level visual tasks. For frame-level tasks, such as tool detection, phase
  recognition, action recognition, triplet detection, and tool localization, the episode is anchored to the annotated frame and the
  primary visual input is that frame. For interval-level tasks, such as CVS assessment and GOALS-style skill assessment, the
  episode stores the annotated temporal span together with representative visual references and interval metadata.

  An annotation interval specifies the scope of a reference score; it does not
  imply that the model sees every frame of that interval. The frame budget is an
  upper bound on the images provided, not a fixed clip length: a single-frame
  episode is not expanded into a multi-frame clip by setting a larger cap.

  To control visual context length, we cap the number of frames exposed to a model by \texttt{max\_frames}. Given \(N\) candidate
  frames and a frame budget \(K\), if \(N>K\), we sample uniformly with endpoint preservation:
  \[
  \mathcal{I}
  =
  \left\{
  \mathrm{round}\!\left(
  \frac{i(N-1)}{K-1}
  \right)
  \right\}_{i=0}^{K-1}.
  \]
  The sampled indices are then sorted and deduplicated. This preserves the first and last frames of an interval while distributing
  the remaining visual evidence across the temporal span. For Stage~I diagnostic evaluation, we set \(K=16\).

  This sampling rule applies only to the visual inputs supplied as part of the benchmark episode. In \gym, agents may additionally
  request tool-generated visual evidence through \texttt{perceive()}; the runtime limits for such tool calls are specified
  separately in Appendix~\ref{app:runtime_constraints}, and the proposal-generation details are specified in
  Appendix~\ref{app:perceive_details}.

  \subsection{Task output schemas}
  \label{app:task_schema}

  Each benchmark episode specifies a constrained final-answer schema. The same task-level answer schema is used for direct
  prediction in Stage~I and for the \texttt{submit()} action in \gym. This separates the benchmark answer contract from the agent-control schema described in Appendix~\ref{app:agent_schema}.

  \paragraph{Categorical labels.}
  Phase recognition uses a single canonical label string:
  \[
      \hat{y} \in \mathcal{C}_{\mathrm{phase}}.
  \]

  \paragraph{Label sets.}
  Tool detection, action recognition, and triplet detection use unordered label sets:
  \[
      \hat{Y} \subseteq \mathcal{C}.
  \]
  For triplet detection, each label is a complete \(\langle\mathrm{instrument},\mathrm{verb},\mathrm{target}\rangle\) string. Empty
  sets are valid outputs when no label is present.

  \paragraph{Bounding boxes.}
  Tool localization uses a list of labeled axis-aligned boxes:
  \[
      \{(\ell_j, x_j, y_j, w_j, h_j)\}_{j=1}^{M},
  \]
  where coordinates are normalized to \([0,1]\) relative to image width and height.

  \paragraph{Ordinal rubric dictionaries.}
  CVS and skill assessment use fixed-key ordinal dictionaries. CVS uses three keys \(\{c1,c2,c3\}\) with scores in \(\{0,1,2\}\).
  Skill assessment uses GOALS-style keys, including depth perception, bimanual dexterity, efficiency, tissue handling, and case
  difficulty, on the original ordinal scale.

  All model responses are parsed into these structured objects before scoring. Malformed but recoverable responses undergo one
  deterministic repair pass; responses that remain non-conformant are mapped to the task-specific null prediction and retained in
  the evaluation.
\subsection{Hard-set difficulty scoring}
\label{app:hardset_scoring}

For phase recognition, \(\mathrm{Perf}_{\tau}\) is exact-match accuracy. For label-set tasks, including tool detection, action recognition, and triplet detection, \(\mathrm{Perf}_{\tau}\) is Jaccard similarity between predicted and reference label sets. For tool localization, \(\mathrm{Perf}_{\tau}\) is mean label-matched IoU. For ordinal workflow assessment, including CVS grading on \(\{0,1,2\}\) and GOALS skill ratings on \(\{1,\ldots,5\}\), \(\mathrm{Perf}_{\tau}\) is range-normalized agreement:
\begin{equation}
    \mathrm{Perf}_{\tau}(m,x)
    =
    1 -
    \frac{1}{|\mathcal{K}_{\tau}|}
    \sum_{k \in \mathcal{K}_{\tau}}
    \frac{|\hat{y}_{m,x,k} - y_{x,k}|}{b_{\tau}-a_{\tau}},
    \label{eq:ordinal_perf}
\end{equation}
where \(\mathcal{K}_{\tau}\) is the set of rubric keys, and \([a_{\tau}, b_{\tau}]\) is the valid ordinal range. All per-episode scores lie in \([0,1]\), so \(1-\mathrm{Perf}_{\tau}\) places errors from heterogeneous tasks on a common difficulty scale. Episodes with high \(\mathrm{PanelError}\) therefore represent failures shared by multiple pretrained models.

\noindent\textbf{Prediction stability.}
To assess the stability of predictions used in hard-set selection, Tab.~\ref{tab:cr_rebuttal_episode_stability} reports outcome flip rates and per-episode score variance across three reruns for two representative backbones. This measures repeated-inference variation at the episode level. Across three reruns, Qwen2.5-VL-7B and Qwen3-VL-8B each have an episode flip rate of 0.32\%, with mean per-episode score variance of 0.0005 and 0.0003, respectively (Tab.~\ref{tab:cr_rebuttal_episode_stability}).

\begin{table}[htbp]
\centering
\scriptsize
\setlength{\tabcolsep}{3pt}
\renewcommand{\arraystretch}{1.15}
\caption{Flip rate = fraction of episodes whose outcome changes across the three reruns; mean per-episode variance computed over episode-level scores.}
\label{tab:cr_rebuttal_episode_stability}
\begin{tabular*}{\linewidth}{@{\extracolsep{\fill}}lcc@{}}
\toprule
Model & Episode flip rate & Mean per-episode score variance \\
\midrule
Qwen2.5-VL-7B & 0.32\% & 0.0005 \\
Qwen3-VL-8B & 0.32\% & 0.0003 \\
\bottomrule
\end{tabular*}
\end{table}

\subsection{Evaluation metrics}
\label{app:metrics}

This section gives the full metric definitions used by our evaluator. Model outputs are first normalized into one of three structured formats: label sets, ordinal rubric dictionaries, or labeled bounding boxes. The main paper reports a compact subset of these metrics in Tab.~\ref{tab:c0_full_suite_summary}; the remaining metrics are used for diagnostic analyses and are reported in Appendix~\ref{app_stage1}.

\paragraph{Main-table metrics.}
Tab.~\ref{tab:c0_full_suite_summary} reports macro-\(F_1\) for tool detection, action recognition, and full triplet detection; strict accuracy for phase recognition; \(F_1@0.5\) for tool localization; mean per-key accuracy and mean quadratic-weighted Cohen's \(\kappa\) for CVS assessment; and mean absolute error (MAE) and mean quadratic-weighted \(\kappa\) for skill assessment. The rank column is computed across the seven task axes; for workflow-assessment columns, \(\kappa\) is used for ranking because it corrects for chance agreement and is less vulnerable to majority-value collapse.

\paragraph{Score aggregation.}
We first compute performance for each available \((\mathrm{dataset}, \mathrm{task})\) cell. Task-level scores are then obtained by a macro mean over datasets supporting that task: \(s_{\tau}(m) = \frac{1}{|\mathcal{D}_{\tau}|} \sum_{d \in \mathcal{D}_{\tau}} s_{d,\tau}(m)\), where \(s_{d,\tau}(m)\) is the score of model \(m\) on dataset \(d\) for task axis \(\tau\), and \(\mathcal{D}_{\tau}\) is the set of datasets with annotations for \(\tau\). This aggregation prevents large datasets from dominating task-level performance. The \emph{Rank} column in Tab.~\ref{tab:c0_full_suite_summary} reports each model's mean rank across the seven task axes; for paired workflow columns, we rank by the chance-corrected \(\kappa\) value. Per-dataset scores and auxiliary metrics are in Appendix~\ref{app_stage1}.

\paragraph{Label-set metrics.}
Tool detection, action recognition, and triplet detection are evaluated as label-set prediction tasks. Phase recognition is treated as a single-label special case and is reported with strict accuracy in the main table. For sample \(i\), let \(\hat{Y}_i\) denote the predicted label set and \(Y_i\) the ground-truth label set. Exact-set accuracy is
\[
    \mathrm{Acc}_{\mathrm{set}}
    =
    \frac{1}{N}
    \sum_{i=1}^{N}
    \mathbb{I}\!\left[\hat{Y}_i = Y_i\right].
\]
For class \(c\), let \(TP_c\), \(FP_c\), and \(FN_c\) denote class-level true positives, false positives, and false negatives. Class-wise \(F_1\) and macro-averaged \(F_1\) are
\[
    F_1^{(c)}
    =
    \frac{2TP_c}{2TP_c+FP_c+FN_c},
    \qquad
    F_1^{\mathrm{macro}}
    =
    \frac{1}{|\mathcal{C}|}
    \sum_{c\in\mathcal{C}} F_1^{(c)}.
\]
Macro-\(F_1\) gives equal weight to every class and is therefore less dominated by frequent surgical labels. The evaluator also reports micro-averaged precision, recall, and \(F_1\), computed by pooling \(TP\), \(FP\), and \(FN\) over all classes, together with Jaccard similarity:
\[
    J =
    \frac{TP}{TP+FP+FN}.
\]
For label-set tasks where the empty set is a valid answer, empty predictions and empty references are preserved. When appropriate, the evaluation vocabulary is augmented with a synthetic \(\langle\mathrm{no\_label}\rangle\) class so that empty-set agreement is counted as a valid prediction event.

\paragraph{Triplet detection metrics.}
For triplet detection, each label in the headline metric is a complete \(\langle\mathrm{instrument},\mathrm{verb},\mathrm{target}\rangle\) tuple. A prediction receives full-triplet credit only when all three components are jointly correct. For diagnostics, we additionally project predicted and ground-truth triplet sets onto the instrument, verb, and target axes:
\[
\begin{aligned}
    \hat{Y}^{\mathrm{inst}}_i &= \{\mathrm{instrument}(t): t\in\hat{Y}_i\}, \\
    \hat{Y}^{\mathrm{verb}}_i &= \{\mathrm{verb}(t): t\in\hat{Y}_i\}, \\
    \hat{Y}^{\mathrm{target}}_i &= \{\mathrm{target}(t): t\in\hat{Y}_i\}.
\end{aligned}
\]
The same label-set metrics are then computed on each projection. These component-wise scores help distinguish instrument, verb, and target recognition failures from compositional binding failures.

\paragraph{Workflow assessment metrics.}
CVS assessment and skill assessment are ordinal dictionary-valued tasks. CVS uses three criteria,
\[
    \mathcal{K}_{\mathrm{CVS}}=\{c1,c2,c3\},
\]
with scores in \(\{0,1,2\}\). Skill assessment uses GOALS-style rubric keys, including depth perception, bimanual dexterity, efficiency, tissue handling, and case difficulty, on the original ordinal scale. For a rubric key \(k\), per-key exact-score accuracy is
\[
    \mathrm{Acc}_{k}
    =
    \frac{1}{N}
    \sum_{i=1}^{N}
    \mathbb{I}\!\left[
    \hat{v}_{i,k}=v_{i,k}
    \right].
\]
The CVS accuracy reported in the main table is the mean of \(\mathrm{Acc}_{k}\) over the three CVS keys. We also compute strict dictionary accuracy,
\[
    \mathrm{Acc}_{\mathrm{dict}}
    =
    \frac{1}{N}
    \sum_{i=1}^{N}
    \mathbb{I}\!\left[
    \forall k\in\mathcal{K},\,
    \hat{v}_{i,k}=v_{i,k}
    \right],
\]
which requires all rubric keys to be correct for a sample to receive credit.

For ordinal error magnitude, we report mean absolute error per key and averaged over keys:
\[
    \mathrm{MAE}_{k}
    =
    \frac{1}{N}
    \sum_{i=1}^{N}
    |\hat{v}_{i,k}-v_{i,k}|,
    \qquad
    \mathrm{MAE}_{\mathrm{mean}}
    =
    \frac{1}{|\mathcal{K}|}
    \sum_{k\in\mathcal{K}}
    \mathrm{MAE}_{k}.
\]
For chance-corrected ordinal agreement, we compute quadratic-weighted Cohen's \(\kappa\) per key and average across keys. For one key, let \(O\) be the observed agreement matrix and \(E\) the expected agreement matrix under independent marginals. The quadratic weight between ordinal labels \(l_i\) and \(l_j\) is
\[
    W_{ij}
    =
    \frac{(l_i-l_j)^2}
         {(l_{\max}-l_{\min})^2},
\]
and the weighted kappa is
\[
    \kappa_w
    =
    1 -
    \frac{\sum_{i,j} W_{ij}O_{ij}}
         {\sum_{i,j} W_{ij}E_{ij}},
    \qquad
    \kappa^{\mathrm{mean}}_w
    =
    \frac{1}{|\mathcal{K}|}
    \sum_{k\in\mathcal{K}}
    \kappa_{w,k}.
\]
Skill assessment additionally reports Spearman rank correlation per key as a diagnostic for whether a model preserves the relative ordering of cases.

\paragraph{Localization metrics.}
Tool localization predictions are lists of labeled boxes. Each predicted box and ground-truth box is represented by an instrument label and normalized image coordinates. The intersection-over-union between prediction \(p\) and ground truth \(g\) is
\[
    \mathrm{IoU}(p,g)
    =
    \frac{|p\cap g|}{|p\cup g|}.
\]
For each frame, predictions and ground-truth boxes are matched using Hungarian assignment with cost \(1-\mathrm{IoU}\). Label mismatches are assigned zero IoU before matching, so a geometrically accurate box with the wrong instrument label receives no localization credit. At threshold \(0.5\), a matched prediction is a true positive only if it has the correct label and \(\mathrm{IoU}\geq0.5\):
\[
    \mathrm{Precision}@0.5
    =
    \frac{TP_{0.5}}{N_{\mathrm{pred}}},
    \qquad
    \mathrm{Recall}@0.5
    =
    \frac{TP_{0.5}}{N_{\mathrm{gt}}},
\]
\[
    F_1@0.5
    =
    \frac{
    2\,\mathrm{Precision}@0.5\,\mathrm{Recall}@0.5
    }{
    \mathrm{Precision}@0.5+\mathrm{Recall}@0.5
    }.
\]
The main table reports \(F_1@0.5\). The appendix additionally reports mIoU, precision, and recall.

\paragraph{Parsing and null predictions.}
All metrics are computed on parsed structured outputs. The evaluator first applies task-specific normalization and one deterministic repair pass. Non-conformant outputs after repair are converted to the task-specific null prediction: an empty set for label-set tasks, an invalid or missing rubric entry for dictionary tasks, and an empty box list for localization. These failures are kept in the score.

\subsubsection{Statistical analysis}
\label{app:cr_statistical_analysis}
\paragraph{Model ranking.}
We report 95\% confidence intervals for the mean-rank column of Tab.~\ref{tab:c0_full_suite_summary}, recomputing the full model ranking in each of 1,000 video-level bootstrap resamples.

\paragraph{Across-stage comparison.}
Paired per (model, axis) at the video level, testing $\Delta=\text{Stage~II}-\text{Stage~I}$ on the hard set (the deltas reported in Tab.~\ref{tab:stage_delta_summary}, Panel~A), with the same clustered bootstrap and BH-FDR correction. Of 80 testable cells (16 models $\times$ 5 axes), 39 are significant improvements and only 1 a significant regression (Qwen3-VL-8B-Thinking, Tool Det.).
The significant improvements comprise Phase 14/16 models, Action 14/16, Tool Det. 10/16, CVS 1/16, and Skill 0/16.

\section{\gym implementation}
\subsection{Agent schema}
  \label{app:agent_schema}

  \gym separates the logical agent interface from the concrete response format used by each model backend. The shared runtime abstraction is a \texttt{StructuredOutputContract}, which specifies the prompt suffix,
  backend-specific request arguments, parser, and repair prompt. A model response is parsed into either an \texttt{AgentStep} for tool-using episodes or a \texttt{ParsedText} object for no-tool direct-answer episodes.

  \paragraph{Tool-mode output.}
  In tool-using conditions, the environment expects the agent to take one tool action at a time. Internally, each action is represented as an \texttt{ActionCall} with an \texttt{action} string and a \texttt{params}
  object:
  \begin{templatebox}{internal_action_call}
  \begin{TemplateVerb}
  {
    "action": "<tool_name>",
    "params": {
      "<parameter_name>": "<parameter_value>"
    }
  }
  \end{TemplateVerb}
  \end{templatebox}

  The concrete model-side carrier depends on the active structured-output contract. Prompt-only text contracts ask the model to emit the \texttt{ActionCall} object directly, optionally preceded by a short
  \texttt{Reasoning:} line. Backend-native structured contracts wrap the same action call inside an \texttt{agent\_step} object:
  \begin{templatebox}{agent_step_wrapper}
  \begin{TemplateVerb}
  {
    "reasoning": "<brief rationale>",
    "action_call": {
      "action": "<tool_name>",
      "params": {
        "<parameter_name>": "<parameter_value>"
      }
    }
  }
  \end{TemplateVerb}
  \end{templatebox}

  The \texttt{reasoning} field is included only when reasoning mode is enabled. OpenAI models use \texttt{response\_format} JSON schema, vLLM-served models use \texttt{guided\_json} constrained decoding, and Claude
  models use Anthropic native tool use with an internal \texttt{emit\_agent\_step} tool. These carriers differ at the API level but are parsed into the same \texttt{AgentStep} representation.

  \paragraph{No-tool direct-answer output.}
  In \texttt{C0}, the agent does not call environment tools. The structured-output contract instead asks the model to emit a task-aware direct-answer object:
  \begin{templatebox}{direct_answer_schema}
  \begin{TemplateVerb}
  {
    "answer": <task_specific_answer>
  }
  \end{TemplateVerb}
  \end{templatebox}

  When reasoning mode is enabled, the same object includes a required \texttt{reasoning} field:
  \begin{templatebox}{direct_answer_with_reasoning}
  \begin{TemplateVerb}
  {
    "reasoning": "<brief rationale>",
    "answer": <task_specific_answer>
  }
  \end{TemplateVerb}
  \end{templatebox}

  The type of \texttt{answer} is selected from the active task metadata. Label-set tasks use a list of strings; triplet detection uses a list of strings in the form \texttt{<instrument, verb, target>}; localization uses
  a list of box objects; rubric tasks use an ordinal score dictionary; scalar tasks use a string.

  For example, tool detection is parsed as a list of label strings:
  \begin{templatebox}{tool_detection_answer_schema}
  \begin{TemplateVerb}
  {
    "answer": ["grasper", "hook"]
  }
  \end{TemplateVerb}
  \end{templatebox}

  Triplet detection is parsed as a list of complete instrument--verb--target strings:
  \begin{templatebox}{triplet_detection_answer_schema}
  \begin{TemplateVerb}
  {
    "answer": [
      "<grasper, retract, gallbladder>",
      "<hook, dissect, cystic_plate>"
    ]
  }
  \end{TemplateVerb}
  \end{templatebox}

  The default localization schema uses normalized \([0,1]\) \((x,y,w,h)\) boxes:
  \begin{templatebox}{tool_localization_default_schema}
  \begin{TemplateVerb}
  {
    "answer": [
      {
        "instrument": "grasper",
        "x1": 0.18,
        "y1": 0.24,
        "w": 0.31,
        "h": 0.19
      }
    ]
  }
  \end{TemplateVerb}
  \end{templatebox}

  Grounding-specific contracts may instead use the native box convention of the model, represented with \texttt{bbox\_2d} and \texttt{label}. The evaluator later converts these coordinates back to the benchmark
  coordinate system using the stored \texttt{bbox\_convention} metadata.
  \begin{templatebox}{grounding_native_schema}
  \begin{TemplateVerb}
  {
    "answer": [
      {
        "bbox_2d": [120, 240, 510, 430],
        "label": "grasper"
      }
    ]
  }
  \end{TemplateVerb}
  \end{templatebox}

  Rubric-style tasks use a fixed dictionary schema. CVS assessment uses three ordinal criteria:
  \begin{templatebox}{cvs_answer_schema}
  \begin{TemplateVerb}
  {
    "answer": {
      "c1": 2,
      "c2": 1,
      "c3": 2
    }
  }
  \end{TemplateVerb}
  \end{templatebox}

  Skill assessment uses the GOALS-style rubric keys:
  \begin{templatebox}{skill_answer_schema}
  \begin{TemplateVerb}
  {
    "answer": {
      "depth_perception": 4,
      "bimanual_dexterity": 4,
      "efficiency": 3,
      "tissue_handling": 4,
      "case_difficulty": 2
    }
  }
  \end{TemplateVerb}
  \end{templatebox}

  \paragraph{Tool schemas.}
  Tool declarations define the action space exposed to tool-using agents. Each callable tool is represented by a \texttt{ToolSchema} containing the tool name, description, parameter list, required fields, default values,
  and optional enumerated choices. Each parameter is represented by a \texttt{ParameterSchema} with type drawn from \{\texttt{string}, \texttt{integer}, \texttt{float}, \texttt{boolean}, \texttt{array}, \texttt{object},
  \texttt{any}\}. The environment serializes the currently available tool schemas into the observation field \texttt{available\_tools}; the step engine validates the parsed \texttt{ActionCall} against this schema before
  dispatch.

  A simplified schema for \texttt{perceive()} is:
  \begin{templatebox}{perceive_tool_schema}
  \begin{TemplateVerb}
  {
    "name": "perceive",
    "parameters": {
      "frame": {
        "type": "any",
        "default": "current"
      },
      "granularity": {
        "type": "string",
        "enum": ["coarse", "fine"],
        "default": "coarse"
      }
    }
  }
  \end{TemplateVerb}
  \end{templatebox}

  For multi-frame tasks, the \texttt{frame} parameter is dynamically specialized to an integer index aligned with the visual window shown to the model:
  \begin{templatebox}{perceive_multiframe_schema}
  \begin{TemplateVerb}
  {
    "name": "perceive",
    "parameters": {
      "frame": {
        "type": "integer",
        "minimum": 0,
        "maximum": "N-1",
        "default": "N//2"
      },
      "granularity": {
        "type": "string",
        "enum": ["coarse", "fine"],
        "default": "coarse"
      }
    }
  }
  \end{TemplateVerb}
  \end{templatebox}

  The \texttt{retrieve()} tool requires three arguments: \texttt{query} contains the natural-language query; \texttt{knowledge\_type} records the agent's declared knowledge type for trajectory analysis; and \texttt{source\_type} applies the hard source filter. Only \texttt{source\_type} restricts the retrieval pool:
  \begin{templatebox}{retrieve_tool_schema}
  \begin{TemplateVerb}
  {
    "name": "retrieve",
    "parameters": {
      "query": {
        "type": "string",
        "required": true
      },
      "knowledge_type": {
        "type": "string",
        "enum": [
          "declarative",
          "procedural",
          "conditional",
          "metacognitive"
        ],
        "required": true
      },
      "source_type": {
        "type": "string",
        "enum": [
          "guideline",
          "textbook",
          "tutorial_video",
          "case_record"
        ],
        "required": true
      }
    }
  }
  \end{TemplateVerb}
  \end{templatebox}

  The \texttt{submit()} tool terminates the episode. Its \texttt{answer} parameter is task-aware and mirrors the direct-answer schema used in \texttt{C0}:
  \begin{templatebox}{submit_tool_schema}
  \begin{TemplateVerb}
  {
    "name": "submit",
    "parameters": {
      "answer": <task_specific_answer_schema>
    }
  }
  \end{TemplateVerb}
  \end{templatebox}

  \paragraph{Observation schema.}
  Every observation contains the base fields \texttt{goal}, \texttt{available\_tools}, \texttt{chat\_messages}, \texttt{step}, and \texttt{elapsed\_time}. The surgical environment extends this with multimodal and
  configuration fields:
  \begin{templatebox}{observation_schema}
  \begin{TemplateVerb}
  {
    "goal": "<task instruction>",
    "available_tools": [<serialized ToolSchema objects>],
    "chat_messages": [<conversation history>],
    "step": 0,
    "elapsed_time": 0.0,
    "media": [
      {
        "kind": "image",
        "role": "primary",
        "path": "<frame_path>"
      }
    ],
    "context": {
      "<task_metadata_key>": "<task_metadata_value>"
    },
    "system_config": {
      "condition": "C1",
      "visual_config": "V1",
      "tools": ["perceive", "submit"]
    },
    "artifacts": [
      {
        "artifact_id": "<artifact_id>",
        "kind": "perceive_overlay",
        "step": 0,
        "annotated_image": "<overlay_path>"
      }
    ]
  }
  \end{TemplateVerb}
  \end{templatebox}

  The \texttt{media} field contains the visual inputs exposed to the model. The \texttt{context} field stores task metadata after removing internal-only keys. The \texttt{system\_config} field records the active
  condition and visual configuration. The \texttt{artifacts} field accumulates tool-generated outputs, such as perception overlays, so that they remain visible in later turns.

\paragraph{Execution constraints and logging.}
  \label{app:runtime_constraints}
  Tool availability depends on the experimental condition. In \texttt{C0}, no tools are exposed and the model emits a direct answer. In \texttt{C1}, the available tools are \texttt{perceive()} and \texttt{submit()}. In
  \texttt{C2} and higher knowledge conditions, \texttt{retrieve()} is also available.

  Invalid calls do not terminate the episode; instead, the environment returns a \texttt{success=False} observation with a guidance message explaining the violated constraint and listing the remaining valid options. This design keeps tool-use drift visible in the trajectory.
  The harness enforces three runtime constraints. First, \texttt{perceive()} is capped at two calls for single-image tasks and three calls for multi-frame tasks. Duplicate calls with the same \((\mathrm{frame}, \mathrm{granularity})\) pair are rejected because \texttt{perceive()} is deterministic under fixed inputs. Second, \texttt{retrieve()} is capped at two calls per \texttt{source\_type} per episode, yielding an implicit maximum of eight retrieval calls across the four evidence families. A repeated query to the same source family is rejected when its dense-embedding cosine similarity to an earlier query is at least \(0.95\), encouraging the agent to rephrase substantively or query a complementary source. Third, when fewer than two steps remain before episode truncation, the environment restricts \texttt{available\_tools} to \texttt{submit()}, forcing the agent to commit an answer. 
  These constraints are applied identically to proprietary API models, open-source MLLMs, and medical-domain models. They prevent unbounded looping over deterministic tools, encourage source diversification, and ensure that every episode ends with a scoreable answer when possible. The full observation history---including valid and invalid tool calls, call arguments, returned evidence, visual-grounding outputs, intermediate model messages, and final answers---is stored as the agent trajectory. Thus, evaluation reflects both final-answer correctness and the agent's behavior under a stable, enforced tool contract.

  Each timestep is stored as a \texttt{StepRecord}:
  \begin{templatebox}{step_record_schema}
  \begin{TemplateVerb}
  {
    "step": 0,
    "reasoning": "<optional reasoning>",
    "action": {
      "action": "<tool_name>",
      "params": {}
    },
    "direct_answer": null,
    "raw_request": [<model request messages>],
    "raw_response": "<raw model output>",
    "reward": 0.0,
    "terminated": false,
    "truncated": false,
    "tool_success": true,
    "validation_errors": [],
    "validation_warnings": [],
    "elapsed": 0.0,
    "error": ""
  }
  \end{TemplateVerb}
  \end{templatebox}

  A complete rollout is stored as an \texttt{EpisodeRecord}:
  \begin{templatebox}{episode_record_schema}
  \begin{TemplateVerb}
  {
    "task_id": "<task_id>",
    "idx": 0,
    "reward": 1.0,
    "terminated": true,
    "truncated": false,
    "steps": 2,
    "elapsed": 3.14,
    "trajectory": [<StepRecord objects>],
    "chat": [<conversation messages>],
    "metadata": {
      "agent_config": {},
      "env_config": {},
      "artifact_registry": {},
      "ground_truth": "<ground_truth>",
      "prediction": "<prediction>"
    }
  }
  \end{TemplateVerb}
  \end{templatebox}

  This contract-based schema keeps the benchmark interface consistent while allowing each backend to use its strongest available structured-output mechanism: prompt-only JSON, OpenAI JSON schema, vLLM guided decoding,
  Anthropic tool use, or model-native grounding formats.

\noindent\textbf{Parse validity.}
Tab.~\ref{tab:cr_rebuttal_parse_rates} compares initial parse-error rates with those after re-run rescue for each model. This audit tracks compliance with the structured-output contract separately from correctness on the surgical task.

\begin{table}[htbp]
\centering
\scriptsize
\setlength{\tabcolsep}{3pt}
\renewcommand{\arraystretch}{1.15}
\caption{Per-model truncation and parameter-mismatch rates (\%; lower is better). Arrows indicate initial rates $\to$ rates after re-run rescue. Single values are unchanged.}
\label{tab:cr_rebuttal_parse_rates}
\begin{tabular*}{\linewidth}{@{\extracolsep{\fill}}lcc@{}}
\toprule
Model & Truncation (\%) $\downarrow$ & Parameter mismatch (\%) $\downarrow$ \\
\midrule
\multicolumn{3}{@{}l}{\textit{Proprietary (API, unconstrained decoding)}} \\
\cmidrule{1-3}
GPT-4o-mini & $9.7\to0.0$ & 0.0 \\
Claude-Haiku-4.5 & $0.1\to0.0$ & $4.3\to0.1$ \\
\midrule
\multicolumn{3}{@{}l}{\textit{Open-source (constrained decoding)}} \\
\cmidrule{1-3}
InternVL2.5-8B & 0.0 & 0.0 \\
Qwen2.5-VL-7B & 0.0 & 0.0 \\
Qwen3-VL-8B & 0.0 & 0.0 \\
Qwen3-VL-8B-Thinking & 0.0 & 0.0 \\
Pixtral-12B & 0.0 & 0.0 \\
InternVL3-14B & 0.0 & 0.0 \\
InternVL2.5-26B & 0.0 & 0.0 \\
Gemma-3-27B & 0.0 & 0.0 \\
Qwen2.5-VL-32B & 0.0 & 0.0 \\
Qwen3-VL-32B & 0.0 & 0.0 \\
Qwen2.5-VL-72B & 0.0 & 0.0 \\
InternVL2.5-78B & 0.0 & 0.0 \\
\midrule
\multicolumn{3}{@{}l}{\textit{Medical-domain (constrained decoding)}} \\
\cmidrule{1-3}
Lingshu-32B & 0.0 & 0.0 \\
MedGemma-27B & 0.0 & 0.0 \\
\bottomrule
\end{tabular*}
\end{table}

\subsection{Prompt engineering}
We provide full task templates in Appendix~\ref{app:task-templates}.

\subsubsection{Task prompt templates}
\label{app:task-templates}

We report one canonical template per task type (deduplicated across datasets). Placeholders such as
\texttt{\{labels\}}, \texttt{\{instruments\}}, \texttt{\{verbs\}}, and \texttt{\{targets\}} are
populated from dataset-specific label files.

\begin{templatebox}{phase_recognition}
\begin{TemplateVerb}
This frame was captured during {surgery_type}. What is the current surgical phase shown in this frame? Reply with exactly one label copied verbatim from the following list:
{labels}
Use the label text exactly as written. Do not wrap it in JSON, quotes, or any extra punctuation.
Example answer: preparation
\end{TemplateVerb}
\end{templatebox}

\begin{templatebox}{phase_recognition}
\begin{TemplateVerb}
This frame was captured during {surgery_type}. What is the current surgical step shown in this frame? Reply with exactly one label copied verbatim from the following list:
{labels}
Use the label text exactly as written. Do not wrap it in JSON, quotes, or any extra punctuation.

Example answer: nasal corridor creation
\end{TemplateVerb}
\end{templatebox}

\begin{templatebox}{tool_detection}
\begin{TemplateVerb}
This frame was captured during {surgery_type}. Identify every distinct surgical instrument type visible in this frame. Multiple visible instances may belong to the same type, but list each visible type only once. 
Choose labels only from the following list:
{labels}
Use the label text exactly as written.

Respond with a JSON array of label strings, for example:
["grasper", "hook"]

If no instrument is visible in the frame, return an empty array `[]`. Do not write the word `"none"` or any other placeholder — use `[]`.
\end{TemplateVerb}
\end{templatebox}

\begin{templatebox}{triplet_detection}
\begin{TemplateVerb}
This frame was captured during {surgery_type}. Identify every distinct surgical action triplet visible in this frame. Each action triplet describes one surgical interaction: which instrument is acting, what action it performs, and what target it acts on. Each triplet must use one instrument, one verb, and one target from the label lists below. A frame may contain zero, one, or multiple distinct triplets. If no surgical interaction is visible in the frame, return an empty list `[]` — do not write "none" or any placeholder.

**Label space:**
Instruments:{instruments}
Verbs:{verbs}
Targets:{targets}

Use the label text exactly as written for instrument, verb, and target. List each distinct triplet only once, even if it appears multiple times in the frame.

Respond with a JSON array of triplet strings. Each string uses the format "<instrument, verb, target>" with angle brackets, comma-separated, in this exact order. Example (two triplets):
["<grasper, retract, gallbladder>", "<grasper, grasp, peritoneum>"]
Example (no triplet visible):[]
\end{TemplateVerb}
\end{templatebox}

\begin{templatebox}{action_recognition}
\begin{TemplateVerb}
This frame was captured during {surgery_type}. Identify every distinct surgical action visible in this frame. One or multiple actions may apply. 
Choose labels only from the following list:
{labels}
Use the label text exactly as written.

Respond with a JSON array of label strings, for example:
["grasp", "hold"]

If no action is visible in the frame, return an empty array `[]`. Do not write the word `"none"` or any other placeholder — use `[]`.
\end{TemplateVerb}
\end{templatebox}

\begin{templatebox}{cvs_assessment}
\begin{TemplateVerb}
You are given one or more images in time order from the same interval during {surgery_type}. Assess Strasberg's Critical View of Safety (CVS) for this interval using all provided images jointly.

Key mapping:
- c1 = two structures
- c2 = cystic plate
- c3 = hepatocystic triangle

Score each criterion independently using this rubric:

c1 (two structures)
2: Exactly two structures can immediately and clearly be seen connecting to the gallbladder.
1: Two structures are visible, but overlap or technical issues reduce clarity.
0: Two separate structures cannot be confirmed, or more or fewer than two structures are visible.

c2 (cystic plate)
2: The cystic plate is clearly visible to approximately its lower one third.
1: The cystic plate is visible but incompletely exposed or not optimally seen.
0: The cystic plate is not visible, or more dissection is needed.

c3 (hepatocystic triangle)
2: The hepatocystic triangle is cleared of tissue so the cystic structures are clearly visible and no other structures remain in the triangle.
1: Much of the triangle is visible, but incomplete dissection or technical issues reduce certainty.
0: Tissue or technical issues obscure the triangle so clearance cannot be confirmed.

Allowed integer values: 0, 1, 2

Rules:
- Use all provided images together to judge the interval.
- Base the scores only on visible evidence in the provided interval.
- If visibility is insufficient, use the lower score rather than infer unseen anatomy.
- Do not output a total score, a separate CVS yes/no flag, explanations, markdown, or extra keys.

Use exactly these integer score keys: c1, c2, c3. Each value must be one of the allowed integer scores above. Assign each score independently from the current interval; any example numbers shown elsewhere are illustrative only and should not be copied.

Respond with a JSON object:
{"c1": <int>, "c2": <int>, "c3": <int>}
\end{TemplateVerb}
\end{templatebox}

\begin{templatebox}{skill_assessment}
\begin{TemplateVerb}
This task is surgical skill assessment over the provided full-procedure temporal interval from {surgery_type}. Assess the entire interval on the following five criteria, using an integer from 1 to 5 for each:
- depth_perception: judgment of spatial relationships and distances
- bimanual_dexterity: coordination of both hands and instruments
- efficiency: economy and fluency of motion and workflow
- tissue_handling: care and appropriateness when manipulating tissue
- case_difficulty: overall technical difficulty of the case or interval

Use exactly these score keys: depth_perception, bimanual_dexterity, efficiency, tissue_handling, case_difficulty. Each value must be an integer from 1 to 5. Assign each score independently from the current interval; any example numbers shown elsewhere are illustrative only and should not be copied. Do not add a total score, explanations, markdown, or extra keys.

Respond with a JSON object:
{"depth_perception": <int>, "bimanual_dexterity": <int>, "efficiency": <int>, "tissue_handling": <int>, "case_difficulty": <int>}
\end{TemplateVerb}
\end{templatebox}

\definecolor{QwenBg}{HTML}{F4F8FF}
\definecolor{QwenBd}{HTML}{4A78C2}
\definecolor{ClaudeBg}{HTML}{FFF8F0}
\definecolor{ClaudeBd}{HTML}{B6782F}
\definecolor{InternBg}{HTML}{F3FBF6}
\definecolor{InternBd}{HTML}{2D8A57}
\definecolor{DefaultBg}{HTML}{F7F7F7}
\definecolor{DefaultBd}{HTML}{777777}

\begin{templatebox}{tool_localization (unified task, model-specific output)}
\textbf{Task intent (shared across models).}
Exactly one surgical instrument is visible in the frame; the model must output one instrument label and
one bounding box.

\vspace{4pt}
\textbf{Output contracts by model family:}

\begin{tcolorbox}
[colback=QwenBg,colframe=QwenBd,boxrule=0.6pt,arc=1.5pt,left=6pt,right=6pt,top=4pt,bottom=4pt]
\textbf{Qwen-family (Qwen2.5 / Qwen3)}\\
\texttt{\detokenize{Use this exact format: [{"bbox_2d":[x1,y1,x2,y2],"label":"..."}]}}
\end{tcolorbox}

\begin{tcolorbox}
[colback=ClaudeBg,colframe=ClaudeBd,boxrule=0.6pt,arc=1.5pt,left=6pt,right=6pt,top=4pt,bottom=4pt]
\textbf{Claude-family}\\
\texttt{\detokenize{Use this exact format: [{"bbox_2d":[x1,y1,x2,y2],"label":"..."}]}}
\end{tcolorbox}

\begin{tcolorbox}
[colback=InternBg,colframe=InternBd,boxrule=0.6pt,arc=1.5pt,left=6pt,right=6pt,top=4pt,bottom=4pt]
\textbf{InternVL-family}\\
Use this exact format: \\
\texttt{\detokenize{<ref>instrument_type</ref><box>[[x1,y1,x2,y2]]</box>}}
\end{tcolorbox}

\begin{tcolorbox}
[colback=DefaultBg,colframe=DefaultBd,boxrule=0.6pt,arc=1.5pt,left=6pt,right=6pt,top=4pt,bottom=4pt]
\textbf{Other models (no dedicated grounding contract)}\\
\texttt{\detokenize{Use this exact format:
[{"instrument":"...","x1":0.12,"y1":0.34,
"w":0.20,"h":0.15}]}}\\
\texttt{(normalized xywh in [0,1], default template path)}
\end{tcolorbox}

\end{templatebox}

Although \texttt{tool\_localization} is a single task, we use model-specific output contracts to match
each model family's native grounding behavior. Qwen and Claude variants use a JSON array with
\texttt{bbox\_2d} and \texttt{label}, while InternVL uses tag-based \texttt{<ref>/<box>} output. Models
without a dedicated grounding contract follow the default normalized \texttt{xywh} schema
(\texttt{instrument, x1, y1, w, h}). This design keeps prompting stable for each model family and lets
the evaluator normalize coordinates into a unified benchmark space.
  \clearpage

\subsubsection{Tool specifications}
\label{app:tool-specs}

\paragraph{Tool inventory.}
In tool-calling mode, the agent operates over three tools:
\texttt{perceive()}, \texttt{retrieve()}, and \texttt{submit()}.
At runtime, the environment may dynamically narrow the available tool set
(e.g., force-submit window).
We describe the \texttt{perceive()} tool for single-frame tasks as follows:
\begin{templatebox}{perceive() (single-frame)}
\begin{TemplateVerb}
Name: perceive

Purpose:
Generate query-free candidate regions for the current frame.

Core output:
- proposal list (bbox/mask/area_ratio/score/proposal_id)
- optional overlay image with numbered proposals
(attached in the same observation when proposal_count > 0)

Output boundary:
- perceive returns geometric proposals only.
- perceive does NOT return semantic labels
(no instrument/verb/target/anatomy/phase names).

Behavior constraints:
- deterministic for identical (frame, granularity)
- repeating the same (frame, granularity) is redundant
- use granularity="coarse" for compact evidence
- use granularity="fine" for denser coverage
- HARD BUDGET: call perceive AT MOST 2 times in single-frame tasks

Typical good trajectories:
- perceive(frame="current", granularity="coarse") -> submit(...)
- perceive(frame="current", granularity="coarse")
-> perceive(frame="current", granularity="fine")
-> submit(...)

Typical bad trajectories:
- perceive(frame="current", granularity="coarse")
-> perceive(frame="current", granularity="coarse")
- perceive(frame="current", granularity="fine")
-> perceive(frame="current", granularity="fine")
\end{TemplateVerb}
\end{templatebox}
We describe the \texttt{perceive()} tool for multi-frame tasks as follows:
\begin{templatebox}{perceive() (multi-frame)}
\begin{TemplateVerb}
Name: perceive

Purpose:
Generate query-free candidate regions for one selected frame per call.

Frame argument:
- frame is an integer index aligned to the visible window: 0..N-1
- frame 0 is the first visible frame, frame N-1 is the last

Core output:
- proposal list (bbox/mask/area_ratio/score/proposal_id)
- optional overlay image with numbered proposals
(attached in the same observation when proposal_count > 0)

Output boundary:
- perceive returns geometric proposals only.
- perceive does NOT return semantic labels
(no instrument/verb/target/anatomy/phase names).

Behavior constraints:
- deterministic for identical (frame, granularity)
- never repeat the same (frame, granularity) pair
- prefer sparse sampling (start/mid/end) over dense sequential scanning
- HARD BUDGET: call perceive AT MOST 3 times in multi-frame tasks

Typical good trajectories:
- perceive(frame=N//2, granularity="coarse") -> submit(...)
- perceive(frame=0, granularity="coarse")
-> perceive(frame=N//2, granularity="coarse")
-> perceive(frame=N-1, granularity="coarse")
-> submit(...)
- perceive(frame=N//2, granularity="coarse")
-> perceive(frame=N//2, granularity="fine")
-> submit(...)

Typical bad trajectories:
- perceive(frame=5, granularity="coarse")
-> perceive(frame=5, granularity="coarse")
- perceive(frame=5, granularity="fine")
-> perceive(frame=5, granularity="fine")
- perceive(frame=0, granularity="coarse")
-> perceive(frame=1, granularity="coarse")
-> perceive(frame=2, granularity="coarse")
-> perceive(frame=3, granularity="coarse")
\end{TemplateVerb}
\end{templatebox}
We describe the retrieve tool for single-frame tasks as follows:
\begin{templatebox}{retrieve}
\begin{TemplateVerb}
Name: retrieve

Purpose:
Retrieve one best-matching knowledge chunk from the surgical KB
to ground reasoning (definitions, criteria, workflow, examples).

Required arguments:
- query: natural-language knowledge need
- knowledge_type: one of
{declarative, procedural, conditional, metacognitive}
(logged for analysis; does NOT filter retrieval)
- source_type: one of
{guideline, textbook, tutorial_video, case_record}
(hard-restricts which source is searched)

Core output:
- one best-matching chunk with provenance + relevance score
- if source_type in {tutorial_video, case_record},
sampled frame images are attached in the same observation

Behavior constraints:
- budget: AT MOST 2 calls per source_type
- near-paraphrase repeated queries under the same source_type are rejected
(similarity threshold = 0.95)
- prefer complementary source_type choices across calls
(e.g., textbook + guideline + case_record)

Usage guidance:
- call retrieve when definitions/criteria/prototypical patterns are uncertain
- skip retrieve when visual evidence is already sufficient
\end{TemplateVerb}
\end{templatebox}

\clearpage
\subsubsection{System prompt assembly}
\label{app:system-prompt-assembly}

\definecolor{SysBlueBg}{HTML}{F4F8FF}
\definecolor{SysBlueBd}{HTML}{4A78C2}
\definecolor{SysGreenBg}{HTML}{F3FBF6}
\definecolor{SysGreenBd}{HTML}{2D8A57}
\definecolor{SysGrayBg}{HTML}{F7F7F7}
\definecolor{SysGrayBd}{HTML}{777777}

We include system prompt details as follows:
\begin{templatebox}{system prompt composition (setting-dependent)}

\begin{tcolorbox}
[colback=SysGrayBg,colframe=SysGrayBd,boxrule=0.6pt,arc=1.5pt,left=6pt,right=6pt,top=4pt,bottom=4pt]
\textbf{A) Base prompt (no-tool mode)}
\begin{TemplateVerb}
You are a helpful AI assistant. Answer the task directly.
\end{TemplateVerb}
\end{tcolorbox}

\begin{tcolorbox}
[colback=SysBlueBg,colframe=SysBlueBd,boxrule=0.6pt,arc=1.5pt,left=6pt,right=6pt,top=4pt,bottom=4pt]
\textbf{B) Base prompt (tool-calling mode)}
\begin{TemplateVerb}
You are a research-style reasoning agent operating in a tool-using environment. You do NOT have specialized surgical or anatomical knowledge — you have general reasoning ability plus tools that can supply the visual evidence and domain knowledge you lack. Your goal is to answer the current question as accurately and reliably as possible by planning deliberately around what you don't know rather than performing expertise you don't have. Use the available tools to gather relevant evidence when the answer is uncertain, incomplete, hard to localize, or requires verification of a domain criterion or definition. Prefer tool use over unsupported guessing when key evidence is missing. Avoid redundant tool calls when earlier tool results already provide sufficient evidence. At the beginning of an episode, if the answer is not directly obvious from the observation, prefer gathering evidence with the available tools before submitting. Do not invent tool names, parameters, observations, or evidence. Only choose actions from the provided tool list. Return exactly one action at a time. If more evidence is needed, call an appropriate tool; if the answer is sufficiently supported, return the final answer. Task- and tool-specific usage patterns (good / bad trajectory examples) are provided inside each tool's description — those examples adapt to whether the current task is single-image or multi-frame, so follow them for the concrete call shape.
{Perceive Description} / {Retrieve Description}
\end{TemplateVerb}
\end{tcolorbox}

\end{templatebox}

\subsection{\wiki}
\begin{figure*}[t]
    \centering
    \includegraphics[width=\textwidth]{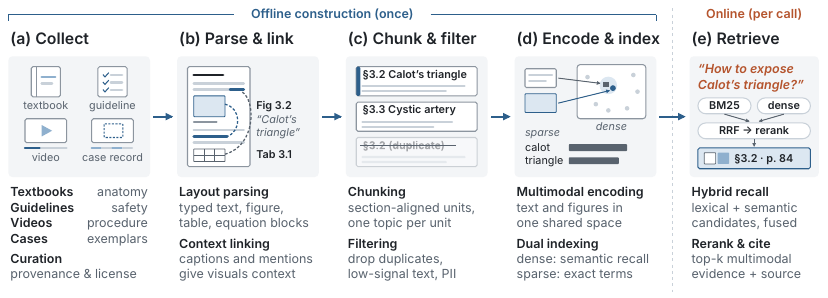}
\caption{\wiki knowledge base construction pipeline.
\textbf{(A)}~Source Collection assembles provenance-tracked corpora from
reference texts, clinical guidelines, tutorial videos, and case records.
\textbf{(B)}~Parsing \& Linking uses MinerU~\citep{mineru2024} to
extract structured text, figures, and tables, linking visual elements
to captions and in-text mentions.
\textbf{(C)}~Chunking \& Filtering applies quality filtering and heading-aware segmentation. \textbf{(D)}~Encoding \& Indexing embeds knowledge units with \texttt{jina-embeddings-v4}~\citep{jina_embeddings_v4} and stores them in dense Qdrant\textsuperscript{\ref{fn:qdrant}} and sparse BM25~\citep{robertson2009probabilistic} indices.
\textbf{(E)}~Retrieval \& Search combines BM25 and dense recall with
RRF fusion~\citep{cormack2009rrf}, followed by text-only
reranking~\citep{jina_reranker_v2} over textual candidates and
multimodal text surrogates, to return top-$k$ evidence with source
attribution.}

    \label{fig:kb_pipeline}
\end{figure*}

\wiki is built through a four-stage pipeline, illustrated in Fig.~\ref{fig:kb_pipeline}; the figure separates curation into parsing/linking and chunking/filtering. Sources are first assembled from open-access repositories and licensed reference materials across four provenance-tracked families (Stage~1: \emph{Source Collection}). Collected material is then processed through a unified multimodal pipeline that parses source documents, removes low-signal content, and decomposes them into retrievable knowledge units (Stage~2: \emph{Curation and Ingestion}). The resulting units are serialised under a versioned schema and indexed for hybrid lexical and semantic search (Stage~3: \emph{Storage and Indexing}). At query time, agents access the corpus through a deterministic retrieval backend that combines sparse, dense, and cross-modal signals (Stage~4: \emph{Retrieval and Search}). Each stage is described in turn below; the agent-facing tool contract that consumes the retrieval backend is specified in Section~\ref{sec:tool_definition}.

\subsubsection{\wiki Source}
\label{app:kb_sources}

We assembled \wiki from four provenance-tracked source families: surgical textbooks, clinical guidelines, narrated tutorial videos, and annotated case records. Sources were selected to provide complementary declarative, procedural, and visually grounded surgical knowledge~\cite{krathwohl2002revision,drake2013general} while prioritising reproducible provenance, stable open accessibility, and coverage of the procedures and task families evaluated in \bench.

The text corpus incorporates major surgical textbooks and operative references, including Bailey \& Love~\cite{williams2018bailey}, Gray's Anatomy~\cite{standring2015gray}, and the Vanderbilt Open Manual of Surgery~\cite{open_manual_surgery}; clinical guidelines from SAGES~\cite{sages_guidelines}, WSES~\cite{de20212020}, and TG18~\cite{tg18_guidelines}; and StatPearls articles~\cite{patel2025statpearls}. The multimodal corpus adds SurgLaVi narrated tutorial-video clips~\cite{surglavi2024} and case records sampled from training annotations in public surgical-video datasets, including CholecT50~\cite{nwoye2022rendezvous}, Cholec80~\cite{twinanda2017endonet}, Cholec80-CVS~\cite{rios2023cholec80cvs}, HeiChole~\cite{wagner2023heichole}, PitVis~\cite{das2025pitvis}, Cataract-101~\cite{schoeffmann2018cataract101}, CATARACTS~\cite{alhajj2019cataracts}, GraSP~\cite{ayobi2025tapis}, AutoLaparo~\cite{wang2022autolaparo}, HeiCo~\cite{maierhein2021heico}, and MultiBypass140~\cite{lavanchy2024multibypass140}.

\subsubsection{\wiki curation and ingestion}
\label{app:kb_ingestion}

All collected source material is processed through a unified multimodal ingestion pipeline that converts heterogeneous source formats into structured, retrievable knowledge units. The pipeline comprises five sequential stages: parsing and layout extraction, multimodal element handling, content filtering, heading-aware chunking, and dense embedding.

\paragraph{Surgical textbooks and clinical guidelines.}
Document-based sources---surgical textbooks, the Open Manual of Surgery, StatPearls articles, and clinical guidelines---are parsed using MinerU~\cite{mineru2024}, a layout-aware PDF extraction system. The heterogeneous typographic structure of surgical documents, which routinely combines dense prose, hierarchical section trees, numbered clinical recommendations, data tables, and annotated figures, motivates this choice over plain-text extraction. MinerU produces a structured element sequence comprising paragraphs, headings, lists, tables, and figures; this structure is preserved throughout the pipeline to retain the semantic context in which surgical concepts, procedural descriptions, and clinical criteria are originally presented.

Tables and figures are treated as first-class multimodal elements. For each table, we retain both a Markdown serialisation for text-based retrieval and the corresponding image crop for visual retrieval, since layout-dependent information---column alignment, merged cells, highlighting---may be lost during serialisation. Figures are extracted as individual image items and linked to their captions. Beyond caption linkage, we identify all in-text cross-references to each table or figure and store the surrounding passage as a supplementary textual anchor. This two-tier association connects every non-textual element not only with its caption but also with the broader discussion in which it is introduced, referenced, or interpreted, reducing retrieval fragmentation at query time.

\paragraph{Tutorial videos.}
SurgLaVi captions define the tutorial segments, so no separate ASR or scene-boundary detector is used. We filter very short or implausibly long spans and extract representative frames at 1 FPS. Each chunk stores the transcript, procedure, time span, and \texttt{frame\_ids}. Retrieval is performed over the transcript; frames are attached to the returned chunk for inspection.

\paragraph{Case records.}
Case-record ingestion reads only training-split annotations. Task-specific renderers convert labels and rubric scores into short natural-language statements. For frame-level tasks, consecutive frames with the same normalized answer are collapsed into one interval; interval-level assessment rows are kept as-is. Each unit stores the text statement, image reference, timestamps, frame IDs, dataset, task, procedure or phase hints, and labels. Retrieval is performed over the text statement; frames are attached to the returned record for inspection. Missing media, empty text, and implausibly long intervals are discarded, and evaluation labels are never read.

\paragraph{Filtering.}
Prior to chunking, low-signal passages are identified and removed. Filtered material includes reference lists, author affiliations, journal and publisher metadata, copyright notices, running headers and footers, and watermark artefacts introduced during PDF rendering. This step prevents retrieval models from surfacing uninformative bibliographic fragments in response to clinical queries.

\paragraph{Chunking.}
The filtered element sequence is partitioned into retrievable text units using a hierarchical, heading-aware strategy. This design follows prior work identifying document segmentation as a key preprocessing decision in retrieval-augmented generation~\cite{wang-etal-2025-document}. Each chunk targets approximately 512 tokens and inherits the nearest ancestor heading as its section label, so that every indexed unit retains local topical context. Chunk boundaries are constrained by the detected document hierarchy, preventing chunks from spanning major section transitions. This yields retrieval units that remain compact for semantic search while preserving section-level coherence.

\paragraph{Embedding.}
Textual chunks are encoded with \texttt{jina-embeddings-v4}~\cite{jina_embeddings_v4} into a 2{,}048-dimensional dense representation. Non-textual elements, including tables, figures, and tutorial-video frames, are encoded with the corresponding vision encoder from the same model family. This provides a shared representation space for textual and visual knowledge units and allows multimodal elements to be retrieved without introducing an additional cross-modal alignment model. Each embedding is stored together with provenance metadata. For document sources, this includes the publication identity, section path, and page reference; for tutorial clips, the video identifier and temporal span; and for case records, the dataset identity, task type, split designation, and annotation labels. When a source contributes both \bench evaluation episodes and \wiki case records, only training-side annotations are included in \wiki. Gold labels from evaluation episodes are excluded from the retrieval corpus to prevent label leakage.

\paragraph{Embedder selection.}
We selected \texttt{jina-embeddings-v4} through a controlled ablation against eight alternative dense encoders. The comparison included general-purpose text retrievers, namely BGE-base-en-v1.5 and BGE-large-en-v1.5~\cite{xiao2024cpack}, E5-large-v2~\cite{wang2022e5}, GTE-large~\cite{li2023gte}, Nomic-embed-text-v1.5~\cite{nussbaum2024nomic}, and \texttt{all-mpnet-base-v2} from the Sentence-Transformers library~\cite{reimers2019sbert,song2020mpnet}; one biomedical-domain retriever, MedCPT~\cite{jin2023medcpt}; and the earlier Jina text-embedding model, \texttt{jina-embeddings-v3}~\cite{jina_embeddings_v3}. 

To isolate the effect of the embedding model, all rows in the ablation use the same document index, chunking strategy, retrieval backend, and evaluation queries. Probe queries and gold chunk identifiers were generated once with \texttt{gpt-4o}~\cite{openai2024gpt4ocard} over the indexed corpus and then held fixed across all encoders. For each model, the same encoder is used at indexing time and query time. We evaluate retrieval quality using Recall@$k$, mean reciprocal rank, and NDCG@10 over 1{,}515 probe queries against the merged document index. The full results are reported in Tab.~\ref{tab:embedder_ablation}.

\begin{table}[htbp]
\centering
\small
\setlength{\tabcolsep}{4pt}
\caption{Embedding-model ablation on the \wiki document index. All models are evaluated with the same chunked corpus, probe queries, gold chunk identifiers, and dense-retrieval pipeline; only the encoder used for indexing and querying is varied. Probe queries ($n=1{,}515$) and gold chunk identifiers were generated once with \texttt{gpt-4o}~\cite{openai2024gpt4ocard}. \emph{Time} denotes total wall-clock time for query embedding and ranking; \emph{Dim} denotes embedding dimensionality. \textbf{Bold} and \underline{underlining} indicate the best and second-best values per column.}
\label{tab:embedder_ablation}
\begin{tabular}{@{}lrrrrrrr@{}}
\toprule
Model & R@1$\uparrow$ & R@5$\uparrow$ & R@10$\uparrow$ & MRR$\uparrow$ & NDCG@10$\uparrow$ & Time (s)$\downarrow$ & Dim \\
\midrule
\multicolumn{8}{@{}l}{\textit{General-purpose dense encoders}}\\
\cmidrule(lr){1-8}
BGE-base-en-v1.5~\cite{xiao2024cpack}        & 0.448 & 0.696 & \underline{0.771} & 0.553 & 0.605 & \underline{363.6} & 768 \\
BGE-large-en-v1.5~\cite{xiao2024cpack}       & \textbf{0.475} & \underline{0.706} & 0.770 & \textbf{0.574} & \textbf{0.622} & 404.8 & 1024 \\
E5-large-v2~\cite{wang2022e5}                & 0.448 & 0.689 & 0.757 & 0.551 & 0.601 & 947.4 & 1024 \\
GTE-large~\cite{li2023gte}                   & 0.460 & 0.695 & 0.761 & 0.558 & 0.607 & 1{,}292.9 & 1024 \\
Nomic-embed-text-v1.5~\cite{nussbaum2024nomic} & 0.432 & 0.668 & 0.741 & 0.530 & 0.581 & 370.8 & 768 \\
all-mpnet-base-v2~\cite{reimers2019sbert,song2020mpnet} & 0.371 & 0.584 & 0.662 & 0.459 & 0.507 & \textbf{354.0} & 768 \\
\midrule
\multicolumn{8}{@{}l}{\textit{Biomedical-domain encoder}}\\
\cmidrule(lr){1-8}
MedCPT~\cite{jin2023medcpt}                  & 0.215 & 0.449 & 0.563 & 0.313 & 0.372 & 368.0 & 768 \\
\midrule
\multicolumn{8}{@{}l}{\textit{Jina encoders (v3: text-only; v4: multimodal)}}\\
\cmidrule(lr){1-8}
jina-embeddings-v3~\cite{jina_embeddings_v3} & 0.361 & 0.621 & 0.720 & 0.472 & 0.531 & 1{,}372.1 & 1024 \\
\textbf{jina-embeddings-v4}~\cite{jina_embeddings_v4} & \underline{0.461} & \textbf{0.715} & \textbf{0.785} & \underline{0.569} & \underline{0.621} & 1{,}334.0 & 2048 \\
\bottomrule
\end{tabular}
\end{table}

Although the strongest text-only baseline performs competitively on strict top-rank metrics, \texttt{jina-embeddings-v4} provides the best overall fit for \wiki because retrieval must operate over both textual and visual knowledge units. In contrast to the text-only baselines, \texttt{jina-embeddings-v4} supports both text and image encoding within the same model family, which is required for retrieving tables, figures, and tutorial-video frames through a unified interface. Using a single multimodal encoder also avoids modality-specific embedding spaces and removes the need for an additional alignment stage before rank fusion. We therefore adopt \texttt{jina-embeddings-v4} as the production embedder. Document embeddings are precomputed during index construction; query encoding is performed at retrieval time. The timing in Tab.~\ref{tab:embedder_ablation} includes query embedding and ranking.

\subsubsection{\wiki storage and indexing}

\paragraph{Knowledge unit schema.}
Each unit produced by the ingestion pipeline is represented as a structured knowledge unit with standardized content, modality, and provenance fields. Textual units contain the chunk text, a unique unit identifier, source identifier, source type, document type, ancestor heading path, and page reference where available. Visual units additionally store the relative image path and the visual element type, such as figure, table, or tutorial frame. Case-record units store the dataset identifier, task type, split designation, and annotation label set. All units are serialized as JSONL, with one record per line, enabling offline inspection, versioned corpus snapshots, and ablation studies over source families, modalities, and metadata fields.

\paragraph{Index structure.}
We build complementary dense and sparse indices over the stored knowledge units. The dense index stores embeddings produced during the embedding stage and is implemented using Qdrant\footnote{\label{fn:qdrant}\url{https://qdrant.tech}}. Text chunks, figure crops, and table crops are stored in separate modality-specific collections, allowing retrieval to be filtered by modality when required. Each collection uses 2{,}048-dimensional embeddings produced by \texttt{jina-embeddings-v4}~\cite{jina_embeddings_v4}. In parallel, we construct a sparse lexical index over textual units using BM25~\cite{robertson2009probabilistic}.

\subsubsection{\wiki retrieval and search}
\label{app:kb_retrieval}
The retrieval backend is the agent-facing entry point into \wiki. It is deterministic---identical \texttt{(query, source)} arguments return identical ranked lists across runs---and isolates retrieval mechanics from agent behaviour. The agent-side tool contract that consumes this backend is specified in Section~\ref{sec:tool_definition}; the present section describes the index-side mechanics that produce the ranked list.

For textual knowledge items, the knowledge search backend is based on a fusion of lexical retrieval (BM25)~\cite{robertson2009probabilistic} and dense semantic retrieval with the text-embedding model \texttt{jina-embeddings-v4}~\cite{jina_embeddings_v4}.
Indexing and retrieval are implemented with Qdrant\textsuperscript{\ref{fn:qdrant}}.
For multimodal knowledge (\eg, tables, figures, videos) where visual representations are available, text embeddings serve as the anchor representation, since they still generalize to domain knowledge.
Visual embeddings contribute only an additional dense retrieval signal due to a lack of domain fine-tuning. The multimodal ranking is fused with Reciprocal Rank Fusion~\citep{cormack2009rrf}.
A cross-encoder reranker then rescores the top 30 fused candidates, and the highest-ranked chunk is returned with source attribution. For tutorial-video and case-record results, representative frames are returned alongside the retrieved text, allowing the agent to inspect visual evidence in the same tool observation.

\paragraph{Hybrid retrieval.}
For a textual query, the backend executes lexical and dense passes in parallel. The lexical pass scores knowledge units with BM25~\cite{robertson2009probabilistic} over the sparse index, surfacing literal matches on instrument names, anatomical landmarks, rubric terms, and standardised guideline phrasing. The dense pass encodes the query with \texttt{jina-embeddings-v4}~\cite{jina_embeddings_v4} into the same 2{,}048-dimensional space as the indexed text chunks and ranks units by cosine similarity over the Qdrant\textsuperscript{\ref{fn:qdrant}} dense collection, recovering semantically related passages whose surface form differs from the query. The two passes are complementary: BM25 anchors retrieval to surface vocabulary, while dense retrieval generalises across paraphrases and synonymous surgical terminology. Their ranked lists are combined into a single hybrid ranking, and cross-modal results from the visual collections are integrated through Reciprocal Rank Fusion~\cite{cormack2009rrf} as described below.

\paragraph{Multimodal retrieval.}
For sources whose units carry both textual and visual representations---document tables, figures, and tutorial-video frames---textual embeddings serve as the anchor signal. Visual-only retrieval over surgical imagery is unreliable with general-purpose multimodal encoders, since \texttt{jina-embeddings-v4} has not been domain-adapted to laparoscopic, microscopic, or open-surgery scenes. Visual embeddings therefore contribute an \emph{additional} dense signal that is fused with the lexical and textual-dense passes via Reciprocal Rank Fusion~\cite{cormack2009rrf}, ensuring that visually salient items remain reachable when their textual description is sparse without letting an unreliable visual ranking dominate the result. When a multimodal unit is returned, representative frames or image crops are attached to the same observation alongside the retrieved text, so the agent can inspect visual evidence in the same tool response.

\paragraph{Cross-encoder reranking.}
The fused ranked list is rescored with a text-only cross-encoder reranker~\cite{nogueira2019passage}, implemented with \texttt{jina-reranker-v2-base-multilingual}~\cite{jina_reranker_v2}. Reranking is applied to textual candidates and textual surrogates of multimodal candidates: text chunks are represented by their body text, tables by their Markdown serialization, and figures by their captions, in-text mentions, and surrounding context. Image embeddings contribute to the initial dense retrieval stage, but the reranker does not directly consume image pixels or visual embeddings. This separates multimodal candidate generation from text-based relevance refinement and avoids introducing an additional image-aware reranking model. Reranking is applied only within a bounded prefetch window over the fused candidates, and the rescored top-$k$ items are returned with full source attribution.

\paragraph{Source-level filtering.}
Each query is dispatched only to the units associated with the agent-selected source family, enforced as a hard filter before the lexical and dense passes are executed. This keeps the retrieval pool aligned with the agent's epistemic decision (\emph{where to search}) and prevents implicit cross-source mixing in the ranked list. Source filtering is also the unit at which the per-episode call budget is enforced (Section~\ref{sec:tool_definition}).

\subsubsection{Retrieval quality estimate}
\label{app:cr_retrieval_quality}
Reporting retrieval precision is nontrivial in our setting since no gold relevance labels exist for agent-generated queries. Relevance and precision require human judgment. We therefore evaluate the \texttt{retrieve()} tool on real trajectories with a TREC-style graded relevance protocol:
\begin{itemize}
\item A designed rubric covering (i) topical match, (ii) constraint satisfaction, and (iii) information coverage, assessing the query-item relevance;
\item Two surgeon experts annotate 50 \texttt{retrieve()} calls sampled from pooled real trajectories as a calibration set; GPT-5 as an LLM judge scores the full pooled sample with the same rubric;
\item We report expert--expert and expert--LLM agreement. On graded relevance, expert--expert linearly-weighted $\kappa$ as 0.460; expert--LLM $\kappa=0.574/0.539$. The LLM judge agrees with each expert at least as well as the experts agree with each other.
\end{itemize}
\texttt{retrieve()} returns a self-calculated relevance score with every item, passed to the agent as part of its observation. This score separates relevant from non-relevant evidence annotated at AUROC = 0.894 against the LLM judge and 0.817 against human expert, and holds within every knowledge source (0.68--0.87).

Regarding the precision, we use relevance as a proxy. Since retrieval quality and precision are also determined by agent queries, precision@1 in our real trajectories can go up to 57\%--82\%.

\subsection{Tool implementation}
\label{app:perceive_details}
Because laparoscopic images differ from natural-image scenes in illumination, texture, and object scale, 
we relax two SAM~2 confidence thresholds: \texttt{pred\_iou\_thresh} is lowered from \(0.80\) to \(0.70\), and \texttt{stability\_score\_thresh} is lowered from \(0.95\) to \(0.90\). 

\paragraph{\texttt{perceive()} proposal filtering and ranking.}
Each proposal returned by \texttt{perceive()} carries a numeric identifier, a confidence score, a binary mask, a bounding box, a pixel area, and an image-normalized area ratio.
\texttt{perceive()} supports two granularities. In \emph{coarse} mode, a proposal is retained only if its area ratio lies in \([0.5\%, 75\%]\), its bounding box covers less than \(60\%\) of the image, and it does not touch two or more image edges with at least \(95\%\) extent. These geometric filters remove whole-frame envelopes and small noise fragments. Surviving proposals are ranked by
$\mathrm{score} \times \mathrm{area\_ratio}$
with raw score, area ratio, and pixel area used as successive tie-breakers; the top \(10\) proposals are returned. 
In \emph{fine} mode, only the lower area-ratio bound is enforced, with threshold \(0.05\%\). No edge-touching or bounding-box coverage filter is applied. Fine-mode proposals are ranked by \((\mathrm{score}, \mathrm{area\_ratio})\), and the top \(30\) proposals are returned. 
If filtering removes all proposals in either mode, the tool falls back to ranking the unfiltered SAM~2 output.

\FloatBarrier

\section{Extended results}
\subsection{Stage I: zero-shot diagnostic evaluation}
\label{app_stage1}
This section expands the Stage~I diagnostic results summarized in Section~\ref{sec:diag}. Fig.~\ref{fig:wosota} compares selected dataset--task scores with random and majority baselines.
\newsavebox{\stageOnePlotBox}
\sbox{\stageOnePlotBox}{\includegraphics{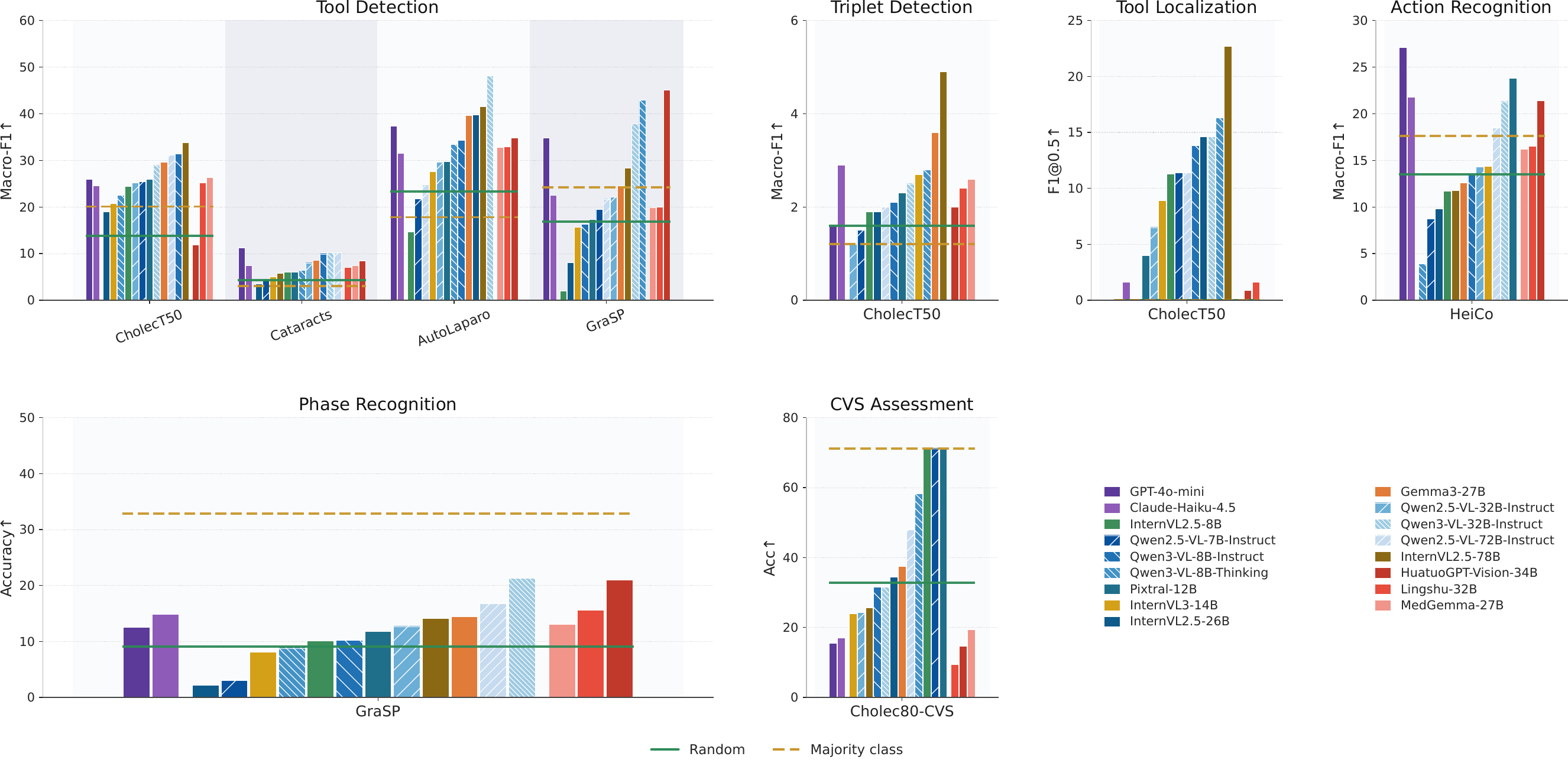}}
\newlength{\stageOnePlotWidth}
\newlength{\stageOnePlotHeight}
\setlength{\stageOnePlotWidth}{\wd\stageOnePlotBox}
\setlength{\stageOnePlotHeight}{\ht\stageOnePlotBox}
\begin{figure}[p]
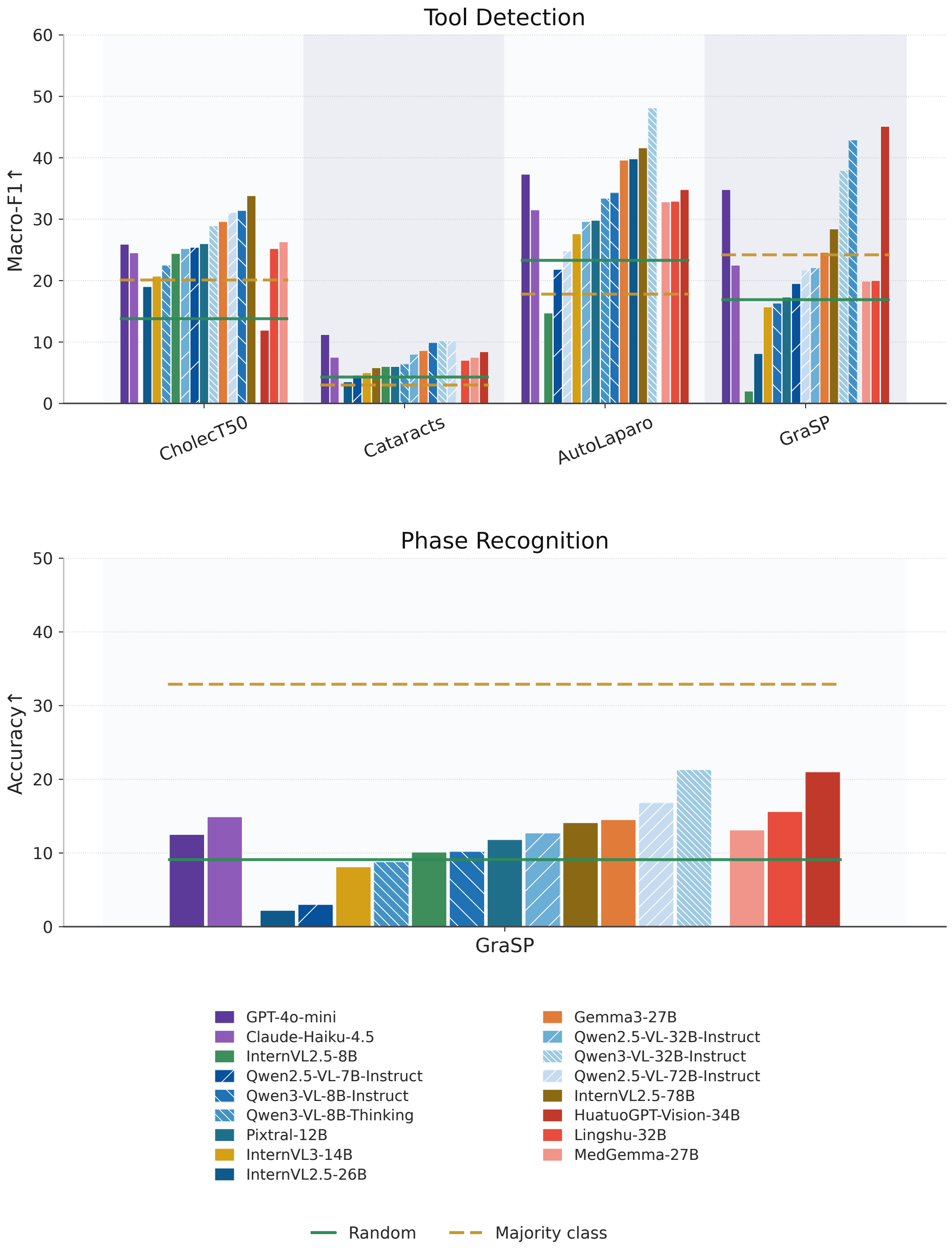

    \centering
    \includegraphics[trim={0pt} {0.035\stageOnePlotHeight} {0.54\stageOnePlotWidth} {0pt},clip,width=\linewidth]{figs/c0_results_without_sota_vertical_seaborn.pdf}
    \par\medskip
    \includegraphics[trim={0.69\stageOnePlotWidth} {0.15\stageOnePlotHeight} {0pt} {0.62\stageOnePlotHeight},clip,width=0.62\linewidth]{figs/c0_results_without_sota_vertical_seaborn.pdf}
    \par\smallskip
    \includegraphics[trim={0.42\stageOnePlotWidth} {0pt} {0.40\stageOnePlotWidth} {0.965\stageOnePlotHeight},clip,width=0.36\linewidth]{figs/c0_results_without_sota_vertical_seaborn.pdf}
    \caption{Stage I performance on dataset--task cells with random and majority baselines: tool detection and phase recognition.}
    \label{fig:wosota}
\end{figure}
\begin{figure}[p]
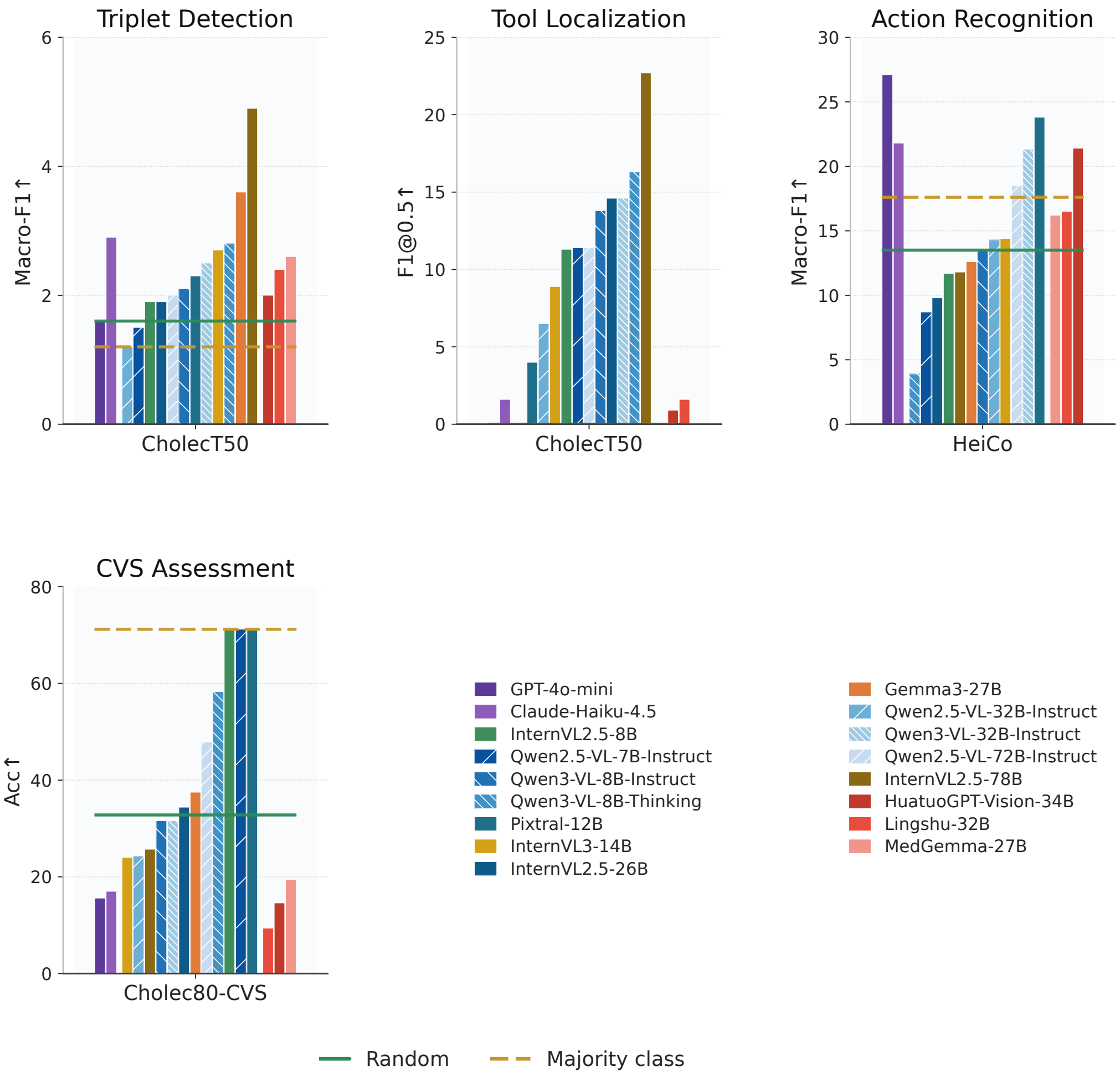

    \ContinuedFloat
    \centering
    \includegraphics[trim={0.46\stageOnePlotWidth} {0.035\stageOnePlotHeight} {0pt} {0pt},clip,width=\linewidth]{figs/c0_results_without_sota_vertical_seaborn.pdf}
    \par\smallskip
    \includegraphics[trim={0.42\stageOnePlotWidth} {0pt} {0.40\stageOnePlotWidth} {0.965\stageOnePlotHeight},clip,width=0.36\linewidth]{figs/c0_results_without_sota_vertical_seaborn.pdf}
    \caption[]{Stage I performance on dataset--task cells with random and majority baselines (continued): triplet detection, tool localization, action recognition, and CVS assessment.}
\end{figure}

\noindent\textbf{Tool detection.}
To assess whether instrument recognition is consistent across surgical datasets, Tab.~\ref{tab:c0_tool_detection_breakdown} separates the Stage~I scores by dataset. The random and majority baselines help distinguish useful recognition from predictions that exploit the label distribution.

\begin{table}[htbp]
\centering
\scriptsize
\setlength{\tabcolsep}{3pt}
\caption{\bench Stage~I tool-detection per-dataset breakdown. We report macro-F1 scores (\%) across the five tool-detection datasets; the main-table value is the unweighted (macro-) average across these columns. \textbf{Bold} / \underline{underline} mark the best / second-best per column \emph{among models}; the \emph{Random} and \emph{Majority} rows are uninformed reference baselines (see main-table caption).}
\label{tab:c0_tool_detection_breakdown}
\begin{tabular}{@{}lccccc@{}}
\toprule
Model & CholecT50 & PitVis & Cataracts & AutoLaparo & GraSP \\
 & {\scriptsize macro-F1$\uparrow$} & {\scriptsize macro-F1$\uparrow$} & {\scriptsize macro-F1$\uparrow$} & {\scriptsize macro-F1$\uparrow$} & {\scriptsize macro-F1$\uparrow$} \\
\midrule
\multicolumn{6}{@{}l}{\textit{Proprietary}}\\
\cmidrule{1-6}
GPT-4o-mini & 25.9 & 4.7 & \textbf{11.2} & 37.3 & 34.8 \\
Claude-Haiku-4.5 & 24.5 & 5.2 & 7.5 & 31.5 & 22.5 \\
\midrule
\multicolumn{6}{@{}l}{\textit{Open-source ($<$10B)}}\\
\cmidrule{1-6}
InternVL2.5-8B & 24.4 & 5.0 & 6.0 & 14.7 & 2.0 \\
Qwen2.5-VL-7B-Instruct & 25.4 & 4.0 & 4.5 & 21.8 & 19.5 \\
Qwen3-VL-8B-Instruct & \underline{31.4} & 4.5 & 9.9 & 34.3 & 16.3 \\
Qwen3-VL-8B-Thinking & 22.5 & \underline{5.4} & 6.4 & 33.4 & \underline{42.9} \\
\midrule
\multicolumn{6}{@{}l}{\textit{Open-source (10--30B)}}\\
\cmidrule{1-6}
Pixtral-12B-2409 & 26.0 & 4.6 & 6.0 & 29.8 & 17.3 \\
InternVL3-14B & 20.7 & 4.5 & 5.0 & 27.6 & 15.7 \\
InternVL2.5-26B & 19.0 & 2.6 & 3.5 & 39.8 & 8.1 \\
Gemma-3-27B-it & 29.6 & 5.1 & 8.6 & 39.6 & 24.6 \\
\midrule
\multicolumn{6}{@{}l}{\textit{Open-source ($>$30B)}}\\
\cmidrule{1-6}
Qwen2.5-VL-32B-Instruct & 25.2 & 5.2 & 8.0 & 29.6 & 22.1 \\
Qwen3-VL-32B-Instruct & 28.9 & 5.4 & \underline{10.2} & \textbf{48.1} & 37.9 \\
Qwen2.5-VL-72B-Instruct & 31.1 & 3.5 & 10.2 & 24.8 & 21.7 \\
InternVL2.5-78B & \textbf{33.8} & 3.4 & 5.8 & \underline{41.6} & 28.4 \\
\midrule
\multicolumn{6}{@{}l}{\textit{Medical-domain}}\\
\cmidrule{1-6}
HuatuoGPT-Vision-34B-hf & 11.9 & 5.2 & 8.4 & 34.8 & \textbf{45.1} \\
Lingshu-32B & 25.2 & 4.4 & 7.0 & 32.9 & 20.0 \\
MedGemma-27B-it & 26.3 & \textbf{5.5} & 7.5 & 32.8 & 19.9 \\
\midrule
\multicolumn{6}{@{}l}{\textit{Statistical Baseline}}\\
\cmidrule{1-6}
Random & 13.8 & 4.1 & 4.3 & 23.3 & 16.9 \\
Majority & 20.1 & 2.5 & 3.0 & 17.8 & 24.2 \\
\bottomrule
\end{tabular}
\end{table}

\noindent\textbf{Triplet detection.}
Recognizing an instrument or an action individually does not establish that a model can identify their interaction. Tab.~\ref{tab:c0_triplet_detection_breakdown} compares full-triplet F1 with the instrument, verb, and target components to make this distinction explicit.

\begin{table}[htbp]
\centering
\scriptsize
\setlength{\tabcolsep}{4pt}
\caption{\bench Stage~I triplet-detection breakdown on CholecT50. We report macro-F1 (\%) for the full $\langle$instrument, verb, target$\rangle$ triplet (the main-table value) and for each component evaluated as a standalone label set. \textbf{Bold} / \underline{underline} mark the best / second-best per column \emph{among models}; the \emph{Random} and \emph{Majority} rows are uninformed reference baselines (see main-table caption).}
\label{tab:c0_triplet_detection_breakdown}
\begin{tabular}{@{}lcccc@{}}
\toprule
Model & Triplet & Instrument & Verb & Target \\
 & {\scriptsize macro-F1$\uparrow$} & {\scriptsize macro-F1$\uparrow$} & {\scriptsize macro-F1$\uparrow$} & {\scriptsize macro-F1$\uparrow$} \\
\midrule
\multicolumn{5}{@{}l}{\textit{Proprietary}}\\
\cmidrule{1-5}
GPT-4o-mini & 1.6 & 13.5 & 9.4 & 8.7 \\
Claude-Haiku-4.5 & 2.9 & 24.7 & 18.1 & 8.4 \\
\midrule
\multicolumn{5}{@{}l}{\textit{Open-source ($<$10B)}}\\
\cmidrule{1-5}
InternVL2.5-8B & 1.9 & 17.9 & 10.4 & 7.3 \\
Qwen2.5-VL-7B-Instruct & 1.5 & 18.6 & 14.1 & 6.1 \\
Qwen3-VL-8B-Instruct & 2.1 & 21.2 & 13.0 & 7.8 \\
Qwen3-VL-8B-Thinking & 2.8 & 20.2 & 15.2 & 11.1 \\
\midrule
\multicolumn{5}{@{}l}{\textit{Open-source (10--30B)}}\\
\cmidrule{1-5}
Pixtral-12B-2409 & 2.3 & 11.4 & 11.2 & 5.7 \\
InternVL3-14B & 2.7 & 21.4 & 14.5 & 9.8 \\
InternVL2.5-26B & 1.9 & 16.9 & 9.6 & 6.1 \\
Gemma-3-27B-it & \underline{3.6} & 22.5 & 16.2 & \underline{11.4} \\
\midrule
\multicolumn{5}{@{}l}{\textit{Open-source ($>$30B)}}\\
\cmidrule{1-5}
Qwen2.5-VL-32B-Instruct & 1.2 & 22.0 & 15.4 & 6.2 \\
Qwen3-VL-32B-Instruct & 2.5 & 18.8 & 12.6 & 9.0 \\
Qwen2.5-VL-72B-Instruct & 2.0 & \underline{28.1} & 15.7 & 10.0 \\
InternVL2.5-78B & \textbf{4.9} & \textbf{34.5} & \textbf{23.5} & \textbf{15.1} \\
\midrule
\multicolumn{5}{@{}l}{\textit{Medical-domain}}\\
\cmidrule{1-5}
HuatuoGPT-Vision-34B-hf & 2.0 & 13.1 & 10.7 & 9.6 \\
Lingshu-32B & 2.4 & 27.4 & \underline{20.8} & 10.9 \\
MedGemma-27B-it & 2.6 & 18.3 & 12.0 & 11.4 \\
\midrule
\multicolumn{5}{@{}l}{\textit{Statistical Baseline}}\\
\cmidrule{1-5}
Random & 1.6 & 15.7 & 10.3 & 6.8 \\
Majority & 1.2 & 10.8 & 7.1 & 6.9 \\
\bottomrule
\end{tabular}
\end{table}

\noindent\textbf{Phase recognition.}
Tab.~\ref{tab:c0_phase_recognition_breakdown} examines phase recognition separately across surgical datasets, showing the variation hidden by the aggregate task score. Each dataset's majority baseline provides a reference for performance obtained by predicting a frequent phase without distinguishing the current workflow state.

\begin{table}[htbp]
\centering
\scriptsize
\setlength{\tabcolsep}{2.5pt}
\caption{\bench Stage~I phase-recognition per-dataset breakdown. We report accuracy (\%) across the seven phase-recognition datasets; the main-table value is the unweighted (macro-) average across these columns. \textbf{Bold} / \underline{underline} mark the best / second-best per column \emph{among models}; the \emph{Random} and \emph{Majority} rows are uninformed reference baselines (see main-table caption).}
\label{tab:c0_phase_recognition_breakdown}
\begin{tabular}{@{}lccccccc@{}}
\toprule
Model & Cataract-101 & Cholec80 & AutoLaparo & GraSP & HeiCo & PitVis & MultiBypass140 \\
 & {\scriptsize Acc.$\uparrow$} & {\scriptsize Acc.$\uparrow$} & {\scriptsize Acc.$\uparrow$} & {\scriptsize Acc.$\uparrow$} & {\scriptsize Acc.$\uparrow$} & {\scriptsize Acc.$\uparrow$} & {\scriptsize Acc.$\uparrow$} \\
\midrule
\multicolumn{8}{@{}l}{\textit{Proprietary}}\\
\cmidrule{1-8}
GPT-4o-mini & 30.0 & \textbf{40.1} & \underline{24.1} & 12.5 & 27.0 & 7.3 & \underline{31.9} \\
Claude-Haiku-4.5 & 19.2 & 20.6 & 17.8 & 14.9 & 25.7 & 7.9 & 22.7 \\
\midrule
\multicolumn{8}{@{}l}{\textit{Open-source ($<$10B)}}\\
\cmidrule{1-8}
InternVL2.5-8B & 30.5 & 38.9 & 20.3 & 10.1 & 17.2 & \textbf{34.5} & 14.8 \\
Qwen2.5-VL-7B-Instruct & 28.7 & 13.6 & 19.4 & 3.0 & 23.3 & 4.9 & 5.7 \\
Qwen3-VL-8B-Instruct & 30.9 & 21.7 & 15.8 & 10.2 & 17.0 & 13.0 & 23.6 \\
Qwen3-VL-8B-Thinking & 30.6 & 36.9 & 19.2 & 8.8 & \underline{27.4} & 11.7 & 27.7 \\
\midrule
\multicolumn{8}{@{}l}{\textit{Open-source (10--30B)}}\\
\cmidrule{1-8}
Pixtral-12B-2409 & 30.5 & 8.9 & 21.2 & 11.8 & 26.6 & 10.7 & 14.1 \\
InternVL3-14B & 30.0 & 10.9 & 21.6 & 8.1 & 27.1 & 14.6 & 24.6 \\
InternVL2.5-26B & \textbf{31.6} & 30.7 & 23.7 & 2.2 & 24.6 & 15.1 & 21.1 \\
Gemma-3-27B-it & 23.8 & 25.9 & 18.9 & 14.5 & 23.8 & 18.1 & 25.0 \\
\midrule
\multicolumn{8}{@{}l}{\textit{Open-source ($>$30B)}}\\
\cmidrule{1-8}
Qwen2.5-VL-32B-Instruct & 31.1 & 21.6 & 20.6 & 12.7 & 26.9 & 16.5 & 24.8 \\
Qwen3-VL-32B-Instruct & 21.2 & \underline{39.6} & \textbf{26.2} & \textbf{21.3} & 27.1 & 3.7 & 26.7 \\
Qwen2.5-VL-72B-Instruct & 30.2 & 39.5 & 20.2 & 16.8 & 27.3 & 9.2 & 21.9 \\
InternVL2.5-78B & \underline{31.1} & 38.7 & 20.9 & 14.1 & 25.5 & 14.8 & 23.2 \\
\midrule
\multicolumn{8}{@{}l}{\textit{Medical-domain}}\\
\cmidrule{1-8}
HuatuoGPT-Vision-34B-hf & 30.0 & 19.2 & 22.6 & \underline{21.0} & \textbf{29.5} & 17.7 & \textbf{33.3} \\
Lingshu-32B & 30.1 & 29.2 & 20.4 & 15.6 & 26.2 & \underline{22.4} & 28.1 \\
MedGemma-27B-it & 30.2 & 23.5 & 19.6 & 13.1 & 15.2 & 13.5 & 23.0 \\
\midrule
\multicolumn{8}{@{}l}{\textit{Statistical Baseline}}\\
\cmidrule{1-8}
Random & 10.0 & 14.2 & 14.2 & 9.1 & 7.3 & 8.4 & 8.3 \\
Majority & 30.5 & 37.4 & 25.1 & 32.9 & 24.8 & 35.2 & 23.1 \\
\bottomrule
\end{tabular}
\end{table}

\noindent\textbf{CVS assessment.}
For ordinal safety criteria, accuracy alone can obscure predictions concentrated on a frequent rating. Tab.~\ref{tab:c0_cvs_breakdown} therefore pairs criterion-level accuracy with chance-corrected agreement, while MAE and exact dictionary accuracy describe rating errors and joint correctness across criteria.

\begin{table}[htbp]
\centering
\scriptsize
\setlength{\tabcolsep}{5pt}
\caption{\bench Stage~I CVS-assessment breakdown on Cholec80-CVS (3-key dict, ordinal labels in $\{0,1,2\}$). \emph{Per-key Acc.}\ and $\kappa$ (quadratic-weighted, multiplied by 100) are the main-table metrics; \emph{MAE} (lower is better; range $[0,2]$) and \emph{Dict-Acc.}\ (strict whole-dictionary accuracy) are reported as diagnostics. \textbf{Bold} / \underline{underline} mark the best / second-best per column \emph{among models}; the \emph{Random} and \emph{Majority} rows are uninformed reference baselines (see main-table caption).}
\label{tab:c0_cvs_breakdown}
\begin{tabular}{@{}lcccc@{}}
\toprule
Model & Per-key Acc. & $\kappa$ & MAE & Dict-Acc. \\
 & {\scriptsize Acc.$\uparrow$} & {\scriptsize $\kappa\uparrow$} & {\scriptsize MAE$\downarrow$} & {\scriptsize Acc.$\uparrow$} \\
\midrule
\multicolumn{5}{@{}l}{\textit{Proprietary}}\\
\cmidrule{1-5}
GPT-4o-mini & 15.6 & 0.9 & 1.00 & 0.0 \\
Claude-Haiku-4.5 & 17.0 & -4.8 & 1.15 & 1.0 \\
\midrule
\multicolumn{5}{@{}l}{\textit{Open-source ($<$10B)}}\\
\cmidrule{1-5}
InternVL2.5-8B & \textbf{71.2} & 0.0 & \textbf{0.32} & \textbf{45.8} \\
Qwen2.5-VL-7B-Instruct & \underline{71.2} & 0.0 & \underline{0.32} & \underline{45.8} \\
Qwen3-VL-8B-Instruct & 31.6 & \underline{5.4} & 0.89 & 12.5 \\
Qwen3-VL-8B-Thinking & 58.3 & 2.7 & 0.42 & 13.5 \\
\midrule
\multicolumn{5}{@{}l}{\textit{Open-source (10--30B)}}\\
\cmidrule{1-5}
Pixtral-12B-2409 & 71.2 & 0.0 & 0.32 & 45.8 \\
InternVL3-14B & 24.0 & -0.4 & 0.78 & 1.0 \\
InternVL2.5-26B & 34.4 & -0.7 & 0.78 & 4.2 \\
Gemma-3-27B-it & 37.5 & -3.9 & 0.63 & 2.1 \\
\midrule
\multicolumn{5}{@{}l}{\textit{Open-source ($>$30B)}}\\
\cmidrule{1-5}
Qwen2.5-VL-32B-Instruct & 24.3 & -0.4 & 0.79 & 2.1 \\
Qwen3-VL-32B-Instruct & 31.6 & \textbf{8.2} & 0.72 & 4.2 \\
Qwen2.5-VL-72B-Instruct & 47.9 & -2.7 & 0.53 & 12.5 \\
InternVL2.5-78B & 25.7 & -0.4 & 0.76 & 2.1 \\
\midrule
\multicolumn{5}{@{}l}{\textit{Medical-domain}}\\
\cmidrule{1-5}
HuatuoGPT-Vision-34B-hf & 14.6 & 0.1 & 1.02 & 0.0 \\
Lingshu-32B & 9.4 & -0.2 & 1.49 & 2.1 \\
MedGemma-27B-it & 19.4 & 4.0 & 1.17 & 9.4 \\
\midrule
\multicolumn{5}{@{}l}{\textit{Statistical Baseline}}\\
\cmidrule{1-5}
Random & 32.8 & 1.7 & 0.92 & 4.2 \\
Majority & 71.2 & 0.0 & 0.32 & 45.8 \\
\bottomrule
\end{tabular}
\end{table}

\noindent\textbf{Skill assessment.}
Tab.~\ref{tab:c0_skill_breakdown} evaluates whether predicted skill ratings track the reference assessments. Reading MAE together with quadratic-weighted $\kappa$ separates closeness to common ratings from agreement beyond chance; the majority baseline makes this distinction visible.

\begin{table}[htbp]
\centering
\scriptsize
\setlength{\tabcolsep}{5pt}
\caption{\bench Stage~I surgical-skill assessment on HeiChole (GOALS-style 5-key rubric, 1--5 Likert). We report MAE (lower is better; range $[0,4]$) and quadratic-weighted Cohen's $\kappa$ (multiplied by 100), each macro-averaged across the three sub-axes (holistic, Calot-phase, dissection-phase). \textbf{Bold} / \underline{underline} mark the best / second-best per column \emph{among models}; the \emph{Random} and \emph{Majority} rows are uninformed reference baselines (see main-table caption).}
\label{tab:c0_skill_breakdown}
\begin{tabular}{@{}lcc@{}}
\toprule
Model & MAE & $\kappa$ \\
 & {\scriptsize MAE$\downarrow$} & {\scriptsize $\kappa\uparrow$} \\
\midrule
\multicolumn{3}{@{}l}{\textit{Proprietary}}\\
\cmidrule{1-3}
GPT-4o-mini & 0.84 & 0.2 \\
Claude-Haiku-4.5 & 0.99 & \underline{10.9} \\
\midrule
\multicolumn{3}{@{}l}{\textit{Open-source ($<$10B)}}\\
\cmidrule{1-3}
InternVL2.5-8B & 0.86 & -7.2 \\
Qwen2.5-VL-7B-Instruct & 1.14 & 8.0 \\
Qwen3-VL-8B-Instruct & 1.05 & 3.9 \\
Qwen3-VL-8B-Thinking & 0.96 & 4.7 \\
\midrule
\multicolumn{3}{@{}l}{\textit{Open-source (10--30B)}}\\
\cmidrule{1-3}
Pixtral-12B-2409 & 1.21 & -0.5 \\
InternVL3-14B & 0.99 & 4.0 \\
InternVL2.5-26B & \textbf{0.64} & 0.0 \\
Gemma-3-27B-it & 1.39 & 3.3 \\
\midrule
\multicolumn{3}{@{}l}{\textit{Open-source ($>$30B)}}\\
\cmidrule{1-3}
Qwen2.5-VL-32B-Instruct & 0.91 & 3.9 \\
Qwen3-VL-32B-Instruct & 0.81 & \textbf{14.3} \\
Qwen2.5-VL-72B-Instruct & 1.08 & 7.2 \\
InternVL2.5-78B & 0.92 & 0.1 \\
\midrule
\multicolumn{3}{@{}l}{\textit{Medical-domain}}\\
\cmidrule{1-3}
HuatuoGPT-Vision-34B-hf & \underline{0.77} & 1.7 \\
Lingshu-32B & 0.80 & 0.4 \\
MedGemma-27B-it & 0.92 & 1.6 \\
\midrule
\multicolumn{3}{@{}l}{\textit{Statistical Baseline}}\\
\cmidrule{1-3}
Random & 0.91 & 1.6 \\
Majority & 0.63 & 0.0 \\
\bottomrule
\end{tabular}
\end{table}

\noindent\textbf{Tool localization.}
Identifying that an instrument is present is different from locating it accurately. Tab.~\ref{tab:c0_tool_localization_breakdown} reports localization F1 alongside overlap, precision, and recall, allowing missed instruments and inaccurate or spurious boxes to be examined separately.

\begin{table}[htbp]
\centering
\scriptsize
\setlength{\tabcolsep}{4pt}
\caption{\bench Stage~I tool-localization breakdown on CholecT50. F1@0.5 is the main-table metric (Hungarian matching with same-class constraint, IoU${\geq}0.5$); mIoU, precision@0.5, and recall@0.5 are reported as auxiliary diagnostics. All values are percentages. \textbf{Bold} / \underline{underline} mark the best / second-best per column \emph{among models}; the \emph{Random} and \emph{Majority} rows are uninformed reference baselines (see main-table caption).}
\label{tab:c0_tool_localization_breakdown}
\begin{tabular}{@{}lcccc@{}}
\toprule
Model & F1@0.5 & mIoU & Prec.@0.5 & Rec.@0.5 \\
 & {\scriptsize F1@0.5$\uparrow$} & {\scriptsize mIoU$\uparrow$} & {\scriptsize Prec.@0.5$\uparrow$} & {\scriptsize Rec.@0.5$\uparrow$} \\
\midrule
\multicolumn{5}{@{}l}{\textit{Proprietary}}\\
\cmidrule{1-5}
GPT-4o-mini & 0.0 & 1.3 & 0.0 & 0.0 \\
Claude-Haiku-4.5 & 1.6 & 5.8 & 1.7 & 1.6 \\
\midrule
\multicolumn{5}{@{}l}{\textit{Open-source ($<$10B)}}\\
\cmidrule{1-5}
InternVL2.5-8B & 11.3 & 13.1 & 11.7 & 11.0 \\
Qwen2.5-VL-7B-Instruct & 11.4 & 12.0 & 11.8 & 11.0 \\
Qwen3-VL-8B-Instruct & 13.8 & 14.8 & 14.2 & 13.4 \\
Qwen3-VL-8B-Thinking & \underline{16.3} & \underline{16.6} & \underline{17.0} & \underline{15.8} \\
\midrule
\multicolumn{5}{@{}l}{\textit{Open-source (10--30B)}}\\
\cmidrule{1-5}
Pixtral-12B-2409 & 4.0 & 10.0 & 4.2 & 3.9 \\
InternVL3-14B & 8.9 & 12.7 & 9.2 & 8.7 \\
InternVL2.5-26B & 14.6 & 15.6 & 15.1 & 14.2 \\
Gemma-3-27B-it & 0.0 & 5.1 & 0.0 & 0.0 \\
\midrule
\multicolumn{5}{@{}l}{\textit{Open-source ($>$30B)}}\\
\cmidrule{1-5}
Qwen2.5-VL-32B-Instruct & 6.5 & 9.1 & 6.7 & 6.3 \\
Qwen3-VL-32B-Instruct & 14.6 & 13.9 & 15.0 & 14.2 \\
Qwen2.5-VL-72B-Instruct & 11.4 & 12.7 & 11.9 & 11.0 \\
InternVL2.5-78B & \textbf{22.7} & \textbf{23.3} & \textbf{23.3} & \textbf{22.1} \\
\midrule
\multicolumn{5}{@{}l}{\textit{Medical-domain}}\\
\cmidrule{1-5}
HuatuoGPT-Vision-34B-hf & 0.9 & 3.3 & 1.2 & 0.8 \\
Lingshu-32B & 1.6 & 6.4 & 1.7 & 1.6 \\
MedGemma-27B-it & 0.0 & 2.2 & 0.0 & 0.0 \\
\midrule
\multicolumn{5}{@{}l}{\textit{Statistical Baseline}}\\
\cmidrule{1-5}
Random & 0.0 & 0.6 & 0.0 & 0.0 \\
Majority & 0.0 & 3.4 & 0.0 & 0.0 \\
\bottomrule
\end{tabular}
\end{table}

\noindent\textbf{Action recognition.}
Tab.~\ref{tab:c0_action_recognition_breakdown} complements class-balanced macro-F1 with pooled precision, recall, and micro-F1. Their comparison shows how performance depends on common versus less frequent actions, while set-exact accuracy measures whether the complete action label set is recovered.

\begin{table}[htbp]
\centering
\scriptsize
\setlength{\tabcolsep}{4pt}
\caption{\bench Stage~I action-recognition breakdown on HeiChole (multi-label, 4 actions). The main-table value is macro-F1; we additionally report micro-F1 / precision / recall (set-level, sample-aggregated) and set-exact accuracy as diagnostics. All values are percentages. \textbf{Bold} / \underline{underline} mark the best / second-best per column \emph{among models}; the \emph{Random} and \emph{Majority} rows are uninformed reference baselines (see main-table caption).}
\label{tab:c0_action_recognition_breakdown}
\begin{tabular}{@{}lccccc@{}}
\toprule
Model & macro-F1 & micro-F1 & Precision & Recall & Set-Acc. \\
 & {\scriptsize macro-F1$\uparrow$} & {\scriptsize F1$\uparrow$} & {\scriptsize Precision$\uparrow$} & {\scriptsize Recall$\uparrow$} & {\scriptsize Acc.$\uparrow$} \\
\midrule
\multicolumn{6}{@{}l}{\textit{Proprietary}}\\
\cmidrule{1-6}
GPT-4o-mini & \textbf{27.1} & \textbf{51.1} & \textbf{41.9} & \textbf{65.4} & \textbf{24.9} \\
Claude-Haiku-4.5 & 21.8 & \underline{44.6} & \underline{35.1} & \underline{61.2} & 7.1 \\
\midrule
\multicolumn{6}{@{}l}{\textit{Open-source ($<$10B)}}\\
\cmidrule{1-6}
InternVL2.5-8B & 11.7 & 10.9 & 10.1 & 11.7 & 7.7 \\
Qwen2.5-VL-7B-Instruct & 8.7 & 7.1 & 6.7 & 7.5 & 6.2 \\
Qwen3-VL-8B-Instruct & 13.5 & 13.7 & 12.4 & 15.2 & 10.7 \\
Qwen3-VL-8B-Thinking & 3.9 & 3.4 & 2.7 & 4.4 & 1.1 \\
\midrule
\multicolumn{6}{@{}l}{\textit{Open-source (10--30B)}}\\
\cmidrule{1-6}
Pixtral-12B-2409 & \underline{23.8} & 38.2 & 29.9 & 52.8 & 9.6 \\
InternVL3-14B & 14.4 & 23.8 & 21.1 & 27.4 & 6.1 \\
InternVL2.5-26B & 9.8 & 10.9 & 10.0 & 11.8 & 4.5 \\
Gemma-3-27B-it & 12.6 & 13.5 & 10.3 & 19.8 & 3.8 \\
\midrule
\multicolumn{6}{@{}l}{\textit{Open-source ($>$30B)}}\\
\cmidrule{1-6}
Qwen2.5-VL-32B-Instruct & 14.3 & 17.6 & 16.8 & 18.5 & 10.2 \\
Qwen3-VL-32B-Instruct & 21.3 & 29.3 & 23.5 & 39.0 & 9.8 \\
Qwen2.5-VL-72B-Instruct & 18.5 & 30.4 & 28.7 & 32.2 & \underline{20.2} \\
InternVL2.5-78B & 11.8 & 9.9 & 8.8 & 11.4 & 8.8 \\
\midrule
\multicolumn{6}{@{}l}{\textit{Medical-domain}}\\
\cmidrule{1-6}
HuatuoGPT-Vision-34B-hf & 21.4 & 31.1 & 26.2 & 38.2 & 9.9 \\
Lingshu-32B & 16.5 & 25.2 & 24.3 & 26.1 & 20.2 \\
MedGemma-27B-it & 16.2 & 23.0 & 18.5 & 30.5 & 4.5 \\
\midrule
\multicolumn{6}{@{}l}{\textit{Statistical Baseline}}\\
\cmidrule{1-6}
Random & 13.5 & 23.1 & 20.4 & 26.5 & 14.9 \\
Majority & 17.6 & 78.3 & 79.0 & 77.7 & 77.3 \\
\bottomrule
\end{tabular}
\end{table}

\subsection{Stage II: agentic evaluation}
\label{app_stage2}
This section provides per-dataset and auxiliary-metric breakdowns for the Stage~II Full-setting
evaluation on \bench-Hard, complementing the aggregate results in
Section~\ref{sec:agent_eval}.
\noindent\textbf{Tool detection with agent interaction.}
Tab.~\ref{tab:c2_hard_tool_detection_breakdown} breaks down instrument-recognition performance on the hard subset by dataset. These absolute scores show where the full agent setting succeeds or remains weak, beyond the aggregate across-stage changes in the main results.

\begin{table}[htbp]
\centering
\scriptsize
\setlength{\tabcolsep}{4pt}
\caption{\bench-Hard Stage~II tool-detection per-dataset breakdown. macro-F1 (\%) across the five datasets. \textbf{Bold} / \underline{underline} mark the best / second-best per column \emph{among models}; the \emph{Random} and \emph{Majority} rows are uninformed reference baselines computed on the hard GT.}
\label{tab:c2_hard_tool_detection_breakdown}
\begin{tabular}{@{}lccccc@{}}
\toprule
Model & CholecT50 & PitVis & Cataracts & AutoLaparo & GraSP \\
 & {\scriptsize macro-F1$\uparrow$} & {\scriptsize macro-F1$\uparrow$} & {\scriptsize macro-F1$\uparrow$} & {\scriptsize macro-F1$\uparrow$} & {\scriptsize macro-F1$\uparrow$} \\
\midrule
\multicolumn{6}{@{}l}{\textit{Proprietary}}\\
\cmidrule{1-6}
GPT-4o-mini & 7.0 & 0.0 & \textbf{23.7} & 18.6 & 12.6 \\
Claude-Haiku-4.5 & 5.3 & 0.9 & 4.4 & 4.5 & 19.8 \\
\midrule
\multicolumn{6}{@{}l}{\textit{Open-source ($<$10B)}}\\
\cmidrule{1-6}
InternVL2.5-8B & 8.5 & 4.6 & \underline{11.7} & 9.8 & 0.0 \\
Qwen2.5-VL-7B-Instruct & 3.5 & 0.0 & 3.1 & 13.0 & 0.0 \\
Qwen3-VL-8B-Instruct & 4.2 & 0.0 & 4.1 & 17.4 & 11.5 \\
Qwen3-VL-8B-Thinking & 1.3 & 0.9 & 1.1 & 18.7 & 24.7 \\
\midrule
\multicolumn{6}{@{}l}{\textit{Open-source (10--30B)}}\\
\cmidrule{1-6}
Pixtral-12B-2409 & \underline{14.0} & \textbf{8.3} & 6.5 & \textbf{23.5} & 29.9 \\
InternVL3-14B & 2.5 & 0.0 & 6.2 & 17.0 & 15.9 \\
InternVL2.5-26B & 0.0 & 3.3 & 2.1 & 1.6 & 17.8 \\
Gemma-3-27B-it & 7.9 & 0.0 & 0.0 & 21.7 & 13.2 \\
\midrule
\multicolumn{6}{@{}l}{\textit{Open-source ($>$30B)}}\\
\cmidrule{1-6}
Qwen2.5-VL-32B-Instruct & 5.7 & 2.2 & 3.1 & 17.4 & 26.2 \\
Qwen3-VL-32B-Instruct & 1.3 & 2.7 & 11.2 & 5.0 & 14.8 \\
Qwen2.5-VL-72B-Instruct & 1.2 & 0.7 & 8.8 & 17.6 & 2.7 \\
InternVL2.5-78B & 11.4 & 0.9 & 0.0 & 9.2 & 11.1 \\
\midrule
\multicolumn{6}{@{}l}{\textit{Medical-domain}}\\
\cmidrule{1-6}
Lingshu-32B & 10.0 & \underline{8.3} & 2.9 & \underline{23.1} & \textbf{36.9} \\
MedGemma-27B-it & \textbf{15.4} & 2.9 & 2.1 & 22.6 & \underline{33.3} \\
\midrule
\multicolumn{6}{@{}l}{\textit{Statistical Baseline}}\\
\cmidrule{1-6}
Random & 20.3 & 10.1 & 15.7 & 25.1 & 44.1 \\
Majority & 14.7 & 5.9 & 23.7 & 17.1 & 41.9 \\
\bottomrule
\end{tabular}
\end{table}

\noindent\textbf{Triplet detection with agent interaction.}
Tab.~\ref{tab:c2_hard_triplet_detection_breakdown} tests whether agents recover complete surgical interactions on hard episodes or only their individual components. Comparing the triplet score with instrument, verb, and target F1 helps locate the remaining compositional difficulty.

\begin{table}[htbp]
\centering
\scriptsize
\setlength{\tabcolsep}{4pt}
\caption{\bench-Hard Stage~II triplet-detection breakdown on CholecT50. Macro-F1 (\%) for the full triplet (the headline metric) and each component as a standalone label set. \textbf{Bold} / \underline{underline} mark best / second-best per column \emph{among models}; the \emph{Random} and \emph{Majority} rows are uninformed reference baselines.}
\label{tab:c2_hard_triplet_detection_breakdown}
\begin{tabular}{@{}lcccc@{}}
\toprule
Model & Triplet & Instrument & Verb & Target \\
 & {\scriptsize macro-F1$\uparrow$} & {\scriptsize macro-F1$\uparrow$} & {\scriptsize macro-F1$\uparrow$} & {\scriptsize macro-F1$\uparrow$} \\
\midrule
\multicolumn{5}{@{}l}{\textit{Proprietary}}\\
\cmidrule{1-5}
GPT-4o-mini & 0.0 & 16.5 & 11.4 & 0.0 \\
Claude-Haiku-4.5 & 0.0 & 21.2 & 14.0 & 5.7 \\
\midrule
\multicolumn{5}{@{}l}{\textit{Open-source ($<$10B)}}\\
\cmidrule{1-5}
InternVL2.5-8B & 0.3 & 16.7 & 11.1 & 6.6 \\
Qwen2.5-VL-7B-Instruct & 0.0 & 8.3 & 8.2 & 7.8 \\
Qwen3-VL-8B-Instruct & 0.0 & 14.6 & 10.3 & 7.9 \\
Qwen3-VL-8B-Thinking & 0.0 & 9.4 & 9.6 & 5.0 \\
\midrule
\multicolumn{5}{@{}l}{\textit{Open-source (10--30B)}}\\
\cmidrule{1-5}
Pixtral-12B-2409 & 0.0 & 19.4 & 14.0 & \textbf{12.7} \\
InternVL3-14B & 2.1 & 8.2 & 5.1 & 7.9 \\
InternVL2.5-26B & 1.5 & 12.2 & 4.3 & 6.3 \\
Gemma-3-27B-it & 2.1 & \textbf{23.7} & 14.4 & 5.9 \\
\midrule
\multicolumn{5}{@{}l}{\textit{Open-source ($>$30B)}}\\
\cmidrule{1-5}
Qwen2.5-VL-32B-Instruct & \textbf{7.7} & \underline{23.0} & \textbf{17.5} & 10.6 \\
Qwen3-VL-32B-Instruct & 0.0 & 12.4 & 10.0 & 9.0 \\
Qwen2.5-VL-72B-Instruct & 0.0 & 13.0 & 7.4 & 7.8 \\
InternVL2.5-78B & \underline{3.8} & 20.3 & \underline{16.2} & \underline{10.8} \\
\midrule
\multicolumn{5}{@{}l}{\textit{Medical-domain}}\\
\cmidrule{1-5}
Lingshu-32B & 0.0 & 19.9 & 7.8 & 7.6 \\
MedGemma-27B-it & 1.0 & 10.7 & 5.9 & 7.8 \\
\midrule
\multicolumn{5}{@{}l}{\textit{Statistical Baseline}}\\
\cmidrule{1-5}
Random & 4.7 & 13.0 & 9.0 & 6.3 \\
Majority & 3.7 & 11.8 & 7.9 & 6.6 \\
\bottomrule
\end{tabular}
\end{table}

\noindent\textbf{Phase recognition with agent interaction.}
Tab.~\ref{tab:c2_hard_phase_recognition_breakdown} reports the full agent's phase-recognition scores for each dataset in the hard subset. The dataset-specific baselines provide context for interpreting absolute accuracy across procedures with different phase distributions.

\begin{table}[htbp]
\centering
\scriptsize
\setlength{\tabcolsep}{4pt}
\caption{\bench-Hard Stage~II phase-recognition per-dataset breakdown. Accuracy (\%) across the seven datasets. \textbf{Bold} / \underline{underline} mark best / second-best per column \emph{among models}; the \emph{Random} and \emph{Majority} rows are uninformed reference baselines.}
\label{tab:c2_hard_phase_recognition_breakdown}
\begin{tabular}{@{}lccccccc@{}}
\toprule
Model & Cataract-101 & Cholec80 & AutoLaparo & GraSP & HeiCo & PitVis & MultiBypass140 \\
 & {\scriptsize Acc.$\uparrow$} & {\scriptsize Acc.$\uparrow$} & {\scriptsize Acc.$\uparrow$} & {\scriptsize Acc.$\uparrow$} & {\scriptsize Acc.$\uparrow$} & {\scriptsize Acc.$\uparrow$} & {\scriptsize Acc.$\uparrow$} \\
\midrule
\multicolumn{8}{@{}l}{\textit{Proprietary}}\\
\cmidrule{1-8}
GPT-4o-mini & 0.0 & 0.0 & 0.0 & 0.0 & 0.0 & \underline{4.0} & 0.0 \\
Claude-Haiku-4.5 & \underline{6.0} & 8.0 & 2.0 & 0.0 & 0.0 & \textbf{20.0} & 8.0 \\
\midrule
\multicolumn{8}{@{}l}{\textit{Open-source ($<$10B)}}\\
\cmidrule{1-8}
InternVL2.5-8B & 0.0 & 0.0 & 6.0 & 0.0 & 0.0 & 0.0 & 0.0 \\
Qwen2.5-VL-7B-Instruct & 4.0 & 4.0 & 4.0 & 0.0 & \underline{4.0} & 0.0 & 6.0 \\
Qwen3-VL-8B-Instruct & \textbf{8.0} & 2.0 & \textbf{12.0} & 0.0 & 0.0 & \underline{4.0} & 0.0 \\
Qwen3-VL-8B-Thinking & 0.0 & 2.0 & 4.0 & 0.0 & 2.0 & \underline{4.0} & 0.0 \\
\midrule
\multicolumn{8}{@{}l}{\textit{Open-source (10--30B)}}\\
\cmidrule{1-8}
Pixtral-12B-2409 & 2.0 & \underline{14.0} & 2.0 & 0.0 & 0.0 & 0.0 & 0.0 \\
InternVL3-14B & 0.0 & 4.0 & 8.0 & 0.0 & 0.0 & \underline{4.0} & 0.0 \\
InternVL2.5-26B & 0.0 & 0.0 & \underline{10.0} & 0.0 & 0.0 & 0.0 & 0.0 \\
Gemma-3-27B-it & 0.0 & 2.0 & 6.0 & 0.0 & \textbf{6.0} & 0.0 & 8.0 \\
\midrule
\multicolumn{8}{@{}l}{\textit{Open-source ($>$30B)}}\\
\cmidrule{1-8}
Qwen2.5-VL-32B-Instruct & 4.0 & \textbf{22.0} & \textbf{12.0} & 0.0 & \underline{4.0} & 2.0 & 8.0 \\
Qwen3-VL-32B-Instruct & 4.0 & 6.0 & \underline{10.0} & \textbf{6.0} & 0.0 & 0.0 & 10.0 \\
Qwen2.5-VL-72B-Instruct & \textbf{8.0} & 4.0 & 8.0 & 0.0 & 2.0 & 0.0 & \underline{12.0} \\
InternVL2.5-78B & \underline{6.0} & 0.0 & 2.0 & 0.0 & 2.0 & 0.0 & 0.0 \\
\midrule
\multicolumn{8}{@{}l}{\textit{Medical-domain}}\\
\cmidrule{1-8}
Lingshu-32B & 2.0 & 6.0 & 8.0 & \underline{2.0} & 2.0 & 0.0 & \textbf{14.0} \\
MedGemma-27B-it & 0.0 & 8.0 & 8.0 & 0.0 & 0.0 & 0.0 & 6.0 \\
\midrule
\multicolumn{8}{@{}l}{\textit{Statistical Baseline}}\\
\cmidrule{1-8}
Random & 19.2 & 25.6 & 17.2 & 35.6 & 12.0 & 15.6 & 15.6 \\
Majority & 40.0 & 54.0 & 38.0 & 68.0 & 40.0 & 66.0 & 38.0 \\
\bottomrule
\end{tabular}
\end{table}

\noindent\textbf{CVS assessment with agent interaction.}
Tab.~\ref{tab:c2_hard_cvs_breakdown} examines whether tool-mediated judgments agree with the reference CVS ratings on hard episodes. Criterion-level accuracy and $\kappa$ are shown together so that frequent-rating predictions are not mistaken for reliable assessment; the auxiliary metrics describe the size and joint pattern of errors.

\begin{table}[htbp]
\centering
\scriptsize
\setlength{\tabcolsep}{5pt}
\caption{\bench-Hard Stage~II CVS-assessment breakdown on Cholec80-CVS (3-key dict, ordinal $\{0,1,2\}$). \emph{Per-key Acc.}\ and $\kappa$ (quadratic-weighted, multiplied by 100) are the headline metrics; accuracy values are percentages; \emph{MAE} (lower is better; range $[0,2]$) and \emph{Dict-Acc.} are diagnostics. \textbf{Bold} / \underline{underline} mark best / second-best per column \emph{among models} (highest for $\uparrow$, lowest for MAE); the \emph{Random} and \emph{Majority} rows are uninformed reference baselines.}
\label{tab:c2_hard_cvs_breakdown}
\begin{tabular}{@{}lcccc@{}}
\toprule
Model & Per-key Acc. & $\kappa$ & MAE & Dict-Acc. \\
 & {\scriptsize Acc.$\uparrow$} & {\scriptsize $\kappa\uparrow$} & {\scriptsize MAE$\downarrow$} & {\scriptsize Acc.$\uparrow$} \\
\midrule
\multicolumn{5}{@{}l}{\textit{Proprietary}}\\
\cmidrule{1-5}
GPT-4o-mini & 18.8 & -1.0 & 0.96 & 2.1 \\
Claude-Haiku-4.5 & 23.3 & 2.2 & 0.87 & 2.1 \\
\midrule
\multicolumn{5}{@{}l}{\textit{Open-source ($<$10B)}}\\
\cmidrule{1-5}
InternVL2.5-8B & 36.5 & -4.8 & 0.86 & 3.1 \\
Qwen2.5-VL-7B-Instruct & 40.6 & -4.3 & 0.80 & 11.5 \\
Qwen3-VL-8B-Instruct & \underline{45.8} & 3.6 & 0.62 & \underline{21.9} \\
Qwen3-VL-8B-Thinking & 29.9 & -0.8 & 1.03 & 9.4 \\
\midrule
\multicolumn{5}{@{}l}{\textit{Open-source (10--30B)}}\\
\cmidrule{1-5}
Pixtral-12B-2409 & 22.9 & -1.6 & 0.89 & 4.2 \\
InternVL3-14B & 25.7 & 0.6 & 0.75 & 2.1 \\
InternVL2.5-26B & 22.9 & 0.7 & 0.97 & 2.1 \\
Gemma-3-27B-it & 33.3 & -5.4 & 0.77 & 6.2 \\
\midrule
\multicolumn{5}{@{}l}{\textit{Open-source ($>$30B)}}\\
\cmidrule{1-5}
Qwen2.5-VL-32B-Instruct & 28.8 & 3.3 & 0.76 & 3.1 \\
Qwen3-VL-32B-Instruct & 40.6 & \textbf{7.7} & \underline{0.61} & 13.5 \\
Qwen2.5-VL-72B-Instruct & \textbf{49.7} & \underline{4.8} & \textbf{0.51} & \textbf{22.9} \\
InternVL2.5-78B & 36.8 & -2.7 & 0.64 & 8.3 \\
\midrule
\multicolumn{5}{@{}l}{\textit{Medical-domain}}\\
\cmidrule{1-5}
Lingshu-32B & 8.0 & 1.1 & 1.41 & 1.0 \\
MedGemma-27B-it & 22.6 & -0.2 & 1.16 & 7.3 \\
\midrule
\multicolumn{5}{@{}l}{\textit{Statistical Baseline}}\\
\cmidrule{1-5}
Random & 34.4 & -1.5 & 0.89 & 4.2 \\
Majority & 71.2 & 0.0 & 0.32 & 45.8 \\
\bottomrule
\end{tabular}
\end{table}

\noindent\textbf{Skill assessment with agent interaction.}
Tab.~\ref{tab:c2_hard_skill_breakdown} reports absolute skill-rating performance after agent interaction. MAE measures rating distance, whereas quadratic-weighted $\kappa$ tests agreement beyond chance, providing complementary views of whether agents distinguish different levels of surgical skill.

\begin{table}[htbp]
\centering
\scriptsize
\setlength{\tabcolsep}{5pt}
\caption{\bench-Hard Stage~II surgical-skill assessment on HeiChole (GOALS-style 5-key rubric, 1--5 Likert). MAE (lower is better; range $[0,4]$) and quadratic-weighted Cohen's $\kappa$ (multiplied by 100), each macro-averaged across the three sub-axes (holistic, Calot-phase, dissection-phase). \textbf{Bold} / \underline{underline} mark best / second-best per column \emph{among models} (lowest for MAE, highest for $\kappa$); the \emph{Random} and \emph{Majority} rows are uninformed reference baselines.}
\label{tab:c2_hard_skill_breakdown}
\begin{tabular}{@{}lcc@{}}
\toprule
Model & MAE & $\kappa$ \\
 & {\scriptsize MAE$\downarrow$} & {\scriptsize $\kappa\uparrow$} \\
\midrule
\multicolumn{3}{@{}l}{\textit{Proprietary}}\\
\cmidrule{1-3}
GPT-4o-mini & \textbf{0.68} & 1.4 \\
Claude-Haiku-4.5 & 0.93 & \underline{6.2} \\
\midrule
\multicolumn{3}{@{}l}{\textit{Open-source ($<$10B)}}\\
\cmidrule{1-3}
InternVL2.5-8B & 0.96 & 5.8 \\
Qwen2.5-VL-7B-Instruct & 0.88 & -4.7 \\
Qwen3-VL-8B-Instruct & 1.07 & 1.8 \\
Qwen3-VL-8B-Thinking & 0.73 & \textbf{12.1} \\
\midrule
\multicolumn{3}{@{}l}{\textit{Open-source (10--30B)}}\\
\cmidrule{1-3}
Pixtral-12B-2409 & 0.79 & 6.0 \\
InternVL3-14B & 0.84 & 4.8 \\
InternVL2.5-26B & \underline{0.69} & -0.9 \\
Gemma-3-27B-it & 0.88 & -1.1 \\
\midrule
\multicolumn{3}{@{}l}{\textit{Open-source ($>$30B)}}\\
\cmidrule{1-3}
Qwen2.5-VL-32B-Instruct & 0.82 & -0.7 \\
Qwen3-VL-32B-Instruct & 0.69 & -0.8 \\
Qwen2.5-VL-72B-Instruct & 0.89 & 4.4 \\
InternVL2.5-78B & 0.92 & -6.3 \\
\midrule
\multicolumn{3}{@{}l}{\textit{Medical-domain}}\\
\cmidrule{1-3}
Lingshu-32B & 0.75 & -2.1 \\
MedGemma-27B-it & 0.94 & -8.9 \\
\midrule
\multicolumn{3}{@{}l}{\textit{Statistical Baseline}}\\
\cmidrule{1-3}
Random & 0.94 & -3.9 \\
Majority & 0.63 & 0.0 \\
\bottomrule
\end{tabular}
\end{table}

\noindent\textbf{Tool localization with agent interaction.}
Tab.~\ref{tab:c2_hard_tool_localization_breakdown} evaluates the spatial accuracy of the full agent on hard localization episodes. Precision and recall separate false or inaccurate proposals from missed instruments, while mIoU complements the thresholded localization F1.

\begin{table}[htbp]
\centering
\scriptsize
\setlength{\tabcolsep}{4pt}
\caption{\bench-Hard Stage~II tool-localization breakdown on CholecT50. $F_1@0.5$ is the headline metric; mIoU, precision@0.5, recall@0.5 are diagnostics. All values are percentages. \textbf{Bold} / \underline{underline} mark best / second-best per column \emph{among models}; the \emph{Random} and \emph{Majority} rows are uninformed reference baselines.}
\label{tab:c2_hard_tool_localization_breakdown}
\begin{tabular}{@{}lcccc@{}}
\toprule
Model & F1@0.5 & mIoU & Prec.@0.5 & Rec.@0.5 \\
 & {\scriptsize $F_1@0.5\uparrow$} & {\scriptsize mIoU$\uparrow$} & {\scriptsize Prec.@0.5$\uparrow$} & {\scriptsize Rec.@0.5$\uparrow$} \\
\midrule
\multicolumn{5}{@{}l}{\textit{Proprietary}}\\
\cmidrule{1-5}
GPT-4o-mini & 5.7 & 5.4 & 5.8 & 5.5 \\
Claude-Haiku-4.5 & 1.6 & 6.1 & 1.7 & 1.6 \\
\midrule
\multicolumn{5}{@{}l}{\textit{Open-source ($<$10B)}}\\
\cmidrule{1-5}
InternVL2.5-8B & 0.7 & 4.3 & 0.7 & 0.8 \\
Qwen2.5-VL-7B-Instruct & 10.5 & 11.4 & 10.8 & 10.2 \\
Qwen3-VL-8B-Instruct & \underline{15.4} & \textbf{16.2} & \underline{15.8} & \underline{15.0} \\
Qwen3-VL-8B-Thinking & 10.5 & 12.4 & 10.8 & 10.2 \\
\midrule
\multicolumn{5}{@{}l}{\textit{Open-source (10--30B)}}\\
\cmidrule{1-5}
Pixtral-12B-2409 & 8.9 & 10.4 & 9.2 & 8.7 \\
InternVL3-14B & 11.3 & 11.8 & 11.7 & 11.0 \\
InternVL2.5-26B & 4.9 & 5.6 & 5.0 & 4.7 \\
Gemma-3-27B-it & 5.7 & 7.2 & 5.8 & 5.5 \\
\midrule
\multicolumn{5}{@{}l}{\textit{Open-source ($>$30B)}}\\
\cmidrule{1-5}
Qwen2.5-VL-32B-Instruct & 6.5 & 9.9 & 6.7 & 6.3 \\
Qwen3-VL-32B-Instruct & \textbf{17.8} & \underline{15.7} & \textbf{18.3} & \textbf{17.3} \\
Qwen2.5-VL-72B-Instruct & 11.4 & 11.3 & 11.8 & 11.0 \\
InternVL2.5-78B & 3.2 & 3.1 & 3.3 & 3.1 \\
\midrule
\multicolumn{5}{@{}l}{\textit{Medical-domain}}\\
\cmidrule{1-5}
Lingshu-32B & 6.5 & 6.0 & 6.7 & 6.3 \\
MedGemma-27B-it & 4.0 & 6.9 & 4.2 & 3.9 \\
\midrule
\multicolumn{5}{@{}l}{\textit{Statistical Baseline}}\\
\cmidrule{1-5}
Random & 0.2 & 1.0 & 0.2 & 0.2 \\
Majority & 0.0 & 3.4 & 0.0 & 0.0 \\
\bottomrule
\end{tabular}
\end{table}

\begin{samepage}
\noindent\textbf{Action recognition with agent interaction.}
Tab.~\ref{tab:c2_hard_action_recognition_breakdown} reports absolute action-recognition performance on the hard subset. Macro-F1 and the majority baseline contextualize the gains reported in the main text, while pooled metrics and exact-set accuracy reveal how those gains translate into complete multi-label predictions.
\par
\end{samepage}

\begin{table}[htbp]
\centering
\scriptsize
\setlength{\tabcolsep}{4pt}
\caption{\bench-Hard Stage~II action-recognition breakdown on HeiChole (multi-label, 4 actions). macro-F1 is the headline metric; micro-F1 / precision / recall (set-level, sample-aggregated) and set-exact accuracy are diagnostics. All values are percentages. \textbf{Bold} / \underline{underline} mark best / second-best per column \emph{among models}; the \emph{Random} and \emph{Majority} rows are uninformed reference baselines.}
\label{tab:c2_hard_action_recognition_breakdown}
\begin{tabular}{@{}lccccc@{}}
\toprule
Model & macro-F1 & micro-F1 & Precision & Recall & Set-Acc. \\
 & {\scriptsize macro-F1$\uparrow$} & {\scriptsize F1$\uparrow$} & {\scriptsize Precision$\uparrow$} & {\scriptsize Recall$\uparrow$} & {\scriptsize Acc.$\uparrow$} \\
\midrule
\multicolumn{6}{@{}l}{\textit{Proprietary}}\\
\cmidrule{1-6}
GPT-4o-mini & 13.2 & 20.0 & 20.0 & 20.0 & 20.0 \\
Claude-Haiku-4.5 & 22.8 & 26.7 & 21.2 & 36.0 & 0.0 \\
\midrule
\multicolumn{6}{@{}l}{\textit{Open-source ($<$10B)}}\\
\cmidrule{1-6}
InternVL2.5-8B & 18.0 & 30.9 & 28.3 & 34.0 & \underline{32.0} \\
Qwen2.5-VL-7B-Instruct & 3.0 & 3.9 & 3.9 & 4.0 & 4.0 \\
Qwen3-VL-8B-Instruct & \underline{27.3} & \textbf{50.0} & \textbf{48.1} & 52.0 & \textbf{48.0} \\
Qwen3-VL-8B-Thinking & 13.5 & 10.8 & 9.8 & 12.0 & 8.0 \\
\midrule
\multicolumn{6}{@{}l}{\textit{Open-source (10--30B)}}\\
\cmidrule{1-6}
Pixtral-12B-2409 & \textbf{27.6} & 46.6 & 35.4 & \textbf{68.0} & 4.0 \\
InternVL3-14B & 4.2 & 5.4 & 4.9 & 6.0 & 0.0 \\
InternVL2.5-26B & 4.4 & 5.7 & 5.5 & 6.0 & 2.0 \\
Gemma-3-27B-it & 12.8 & 15.3 & 12.3 & 20.0 & 2.0 \\
\midrule
\multicolumn{6}{@{}l}{\textit{Open-source ($>$30B)}}\\
\cmidrule{1-6}
Qwen2.5-VL-32B-Instruct & 25.1 & 29.3 & 24.7 & 36.0 & 8.0 \\
Qwen3-VL-32B-Instruct & 26.5 & 22.1 & 18.2 & 28.0 & 2.0 \\
Qwen2.5-VL-72B-Instruct & 25.6 & 45.5 & 36.6 & 60.0 & 10.0 \\
InternVL2.5-78B & 9.2 & 12.4 & 11.1 & 14.0 & 6.0 \\
\midrule
\multicolumn{6}{@{}l}{\textit{Medical-domain}}\\
\cmidrule{1-6}
Lingshu-32B & 26.7 & \underline{47.1} & \underline{37.2} & \underline{64.0} & 12.0 \\
MedGemma-27B-it & 24.0 & 38.6 & 30.0 & 54.0 & 18.0 \\
\midrule
\multicolumn{6}{@{}l}{\textit{Statistical Baseline}}\\
\cmidrule{1-6}
Random & 25.4 & 36.4 & 33.1 & 40.4 & 21.6 \\
Majority & 30.0 & 82.0 & 82.0 & 82.0 & 82.0 \\
\bottomrule
\end{tabular}
\end{table}

\subsubsection{Matched tool comparison}
\label{app:matched_tool_comparison}
To separate the contribution of tool access from the change in the overall agent setting, we compare tools off and on under the same task instruction and two-sentence reasoning preamble. The tool-enabled condition adds a neutral tool-availability instruction. Both conditions use \bench-Hard with the full knowledge base and reasoning enabled. Tab.~\ref{tab:stage_delta_summary}, Panel~B, reports the matched differences for five backbones; Tab.~\ref{tab:matched_tool_scores} provides the corresponding absolute scores for InternVL2.5-8B, Lingshu-32B, and Gemma-3-27B.

\begin{table}[htbp]\centering\tiny\setlength{\tabcolsep}{5pt}
\caption{Absolute scores under a matched prompt on \bench-Hard (952 episodes, full knowledge base, reasoning enabled). The no-tool and tool conditions share the task instruction and reasoning preamble; only tool availability changes. $\Delta$ is the tool-enabled score minus the no-tool score. All scores and differences are multiplied by 100.}
\label{tab:matched_tool_scores}
\begin{tabular}{@{}ll ccc cc cc@{}}
\toprule
Model & Regime & Tool Det. & Triplet Det. & Tool Loc. & Phase Rec. & Action Rec. & CVS Ass. & Skill Ass. \\
 & & {\scriptsize F1$\uparrow$} & {\scriptsize F1$\uparrow$} & {\scriptsize $F_1@0.5\uparrow$} & {\scriptsize Acc.$\uparrow$} & {\scriptsize F1$\uparrow$} & {\scriptsize $\kappa\uparrow$} & {\scriptsize $\kappa\uparrow$} \\
\midrule
\multirow{3}{*}{InternVL2.5-8B} & Tools off & $0.0$ & $0.0$ & $0.0$ & $1.4$ & $0.0$ & $-1.3$ & $0.0$ \\
 & Tools on & $1.9$ & $0.6$ & $8.1$ & $1.7$ & $14.0$ & $5.0$ & $-1.7$ \\
 & $\Delta$ & \cellcolor[HTML]{E9F7E9} $+1.9$ & \cellcolor[HTML]{E9F7E9} $+0.6$ & \cellcolor[HTML]{C9EAC9} $+8.1$ & \cellcolor[HTML]{E9F7E9} $+0.3$ & \cellcolor[HTML]{BEE6BE} $+14.0$ & \cellcolor[HTML]{C9EAC9} $+6.3$ & \cellcolor[HTML]{FCF0F0} $-1.7$ \\
\midrule
\multirow{3}{*}{Lingshu-32B} & Tools off & $0.4$ & $2.2$ & $2.4$ & $0.0$ & $21.7$ & $-1.2$ & $3.8$ \\
 & Tools on & $10.1$ & $0.0$ & $0.0$ & $2.3$ & $38.2$ & $-0.3$ & $-1.9$ \\
 & $\Delta$ & \cellcolor[HTML]{C9EAC9} $+9.7$ & \cellcolor[HTML]{FBE7E7} $-2.2$ & \cellcolor[HTML]{FBE7E7} $-2.4$ & \cellcolor[HTML]{D6EFD6} $+2.3$ & \cellcolor[HTML]{BEE6BE} $+16.5$ & \cellcolor[HTML]{E9F7E9} $+0.9$ & \cellcolor[HTML]{F7D4D4} $-5.7$ \\
\midrule
\multirow{3}{*}{Gemma-3-27B} & Tools off & $5.2$ & $0.0$ & $0.0$ & $0.9$ & $22.9$ & $-1.6$ & $2.1$ \\
 & Tools on & $9.3$ & $3.0$ & $4.9$ & $2.6$ & $29.5$ & $0.9$ & $6.6$ \\
 & $\Delta$ & \cellcolor[HTML]{D6EFD6} $+4.1$ & \cellcolor[HTML]{D6EFD6} $+3.0$ & \cellcolor[HTML]{D6EFD6} $+4.9$ & \cellcolor[HTML]{E9F7E9} $+1.7$ & \cellcolor[HTML]{C9EAC9} $+6.6$ & \cellcolor[HTML]{D6EFD6} $+2.5$ & \cellcolor[HTML]{D6EFD6} $+4.5$ \\
\bottomrule
\end{tabular}
\end{table}

\subsection{Tool performance matrix}

We compare four representative open-source backbones---InternVL2.5-8B, Qwen3-VL-32B-Instruct, Qwen2.5-VL-72B-Instruct, and Lingshu-32B---spanning differences in parameter scale, architecture family, model generation, and medical specialization. In particular, Lingshu-32B serves as a medically aligned backbone, while the remaining models are general-purpose MLLMs. \emph{Full} enables both \texttt{perceive()} and \texttt{retrieve()}, whereas w/o \texttt{perceive()} and w/o \texttt{retrieve()} disable the corresponding tool.

\noindent\textbf{Findings.}
Each $(\text{model},\text{config},\text{dataset},\text{task})$ combination corresponds to a Stage~II evaluation with greedy decoding on \bench-Hard using the same decoding, frame, and interaction budgets as the main experiments. We report the headline metrics from Tab.~\ref{tab:c0_full_suite_summary}, macro-averaged across constituent datasets. Three key observations emerge from Tab.~\ref{tab:ablation}.

\begin{table}[htbp]
  \centering
  \scriptsize
  \setlength{\tabcolsep}{2.6pt}
  \renewcommand{\arraystretch}{0.96}
  \caption{Toolbox ablation on \textbf{\bench-Hard} under Stage~II interaction. \emph{Full} enables both \texttt{perceive()} and \texttt{retrieve()}, while w/o \texttt{perceive()} and w/o \texttt{retrieve()} disable the corresponding tool. F1 and accuracy are percentages, $\kappa$ is multiplied by 100, and MAE is on the original ordinal scale.}
  \label{tab:ablation}
  \begin{tabular}{@{}llccccccc@{}}
    \toprule
     &  & \multicolumn{3}{c}{\textit{Spatial Recognition}}
        & \multicolumn{2}{c}{\textit{Temporal Reasoning}}
        & \multicolumn{2}{c}{\textit{Workflow Assessment}} \\
    \cmidrule(lr){3-5} \cmidrule(lr){6-7} \cmidrule(lr){8-9}
    Model & Config & Tool Det. & Triplet Det. & Tool Loc.
                   & Phase Rec. & Action Rec.
                   & CVS Ass. & Skill Ass. \\
     &  & {\scriptsize macro-F1$\uparrow$} & {\scriptsize macro-F1$\uparrow$}
        & {\scriptsize $F_1@0.5\uparrow$}
        & {\scriptsize Acc.$\uparrow$} & {\scriptsize macro-F1$\uparrow$}
        & {\scriptsize Acc./$\kappa\uparrow$}
        & {\scriptsize MAE$\downarrow$/$\kappa\uparrow$} \\
    \midrule

    \multirow{3}{*}{Qwen3-VL-32B-Instruct~\citep{qwen2025qwen3vl}}
      & \cellcolor{gray!8}Full        & \cellcolor{gray!8}7.0 & \cellcolor{gray!8}0.0 & \cellcolor{gray!8}17.8 & \cellcolor{gray!8}5.1 & \cellcolor{gray!8}26.5 & \cellcolor{gray!8}40.6/7.7 & \cellcolor{gray!8}0.69/-0.8 \\
      & w/o \texttt{perceive()}                   & 8.1 & 5.1 & 12.2 & 4.0 & 24.8 & 47.2/4.5 & 1.06/6.6 \\
      & w/o \texttt{retrieve()}                   & 6.2 & 0.0 & 16.2 & 4.0 & 17.0 & 27.4/4.2 & 0.66/9.7 \\

    \addlinespace[2pt]

    \multirow{3}{*}{Qwen2.5-VL-72B-Instruct~\citep{bai2025qwen25vl}}
      & \cellcolor{gray!8}Full        & \cellcolor{gray!8}6.2 & \cellcolor{gray!8}0.0 & \cellcolor{gray!8}11.4 & \cellcolor{gray!8}4.9 & \cellcolor{gray!8}25.6 & \cellcolor{gray!8}49.7/4.8 & \cellcolor{gray!8}0.89/4.4 \\
      & w/o \texttt{perceive()}                   & 14.5 & 0.0 & 13.1 & 3.7 & 16.4 & 52.8/10.1 & 1.01/-16.8 \\
      & w/o \texttt{retrieve()}                   & 7.3 & 0.0 & 14.6 & 9.7 & 20.2 & 36.8/-1.6 & 0.83/-0.1 \\

    \addlinespace[2pt]

    \multirow{3}{*}{Lingshu-32B~\citep{lingshu2025}}
      & \cellcolor{gray!8}Full        & \cellcolor{gray!8}16.3 & \cellcolor{gray!8}0.0 & \cellcolor{gray!8}6.5 & \cellcolor{gray!8}4.9 & \cellcolor{gray!8}26.7 & \cellcolor{gray!8}8.0/1.1 & \cellcolor{gray!8}0.75/-2.1 \\
      & w/o \texttt{perceive()}                   & 10.5 & 2.0 & 0.0 & 4.3 & 35.6 & 10.1/1.7 & 0.76/9.6 \\
      & w/o \texttt{retrieve()}                   & 8.8 & 2.3 & 12.2 & 4.3 & 45.8 & 14.2/1.9 & 0.80/-1.3 \\

    \addlinespace[2pt]

    \multirow{3}{*}{InternVL2.5-8B~\citep{chen2024internvl25}}
      & \cellcolor{gray!8}Full        & \cellcolor{gray!8}6.9 & \cellcolor{gray!8}0.3 & \cellcolor{gray!8}0.7 & \cellcolor{gray!8}0.9 & \cellcolor{gray!8}18.0 & \cellcolor{gray!8}36.5/-4.8 & \cellcolor{gray!8}0.96/5.8 \\
      & w/o \texttt{perceive()}                   & 4.3 & 2.0 & 0.0 & 1.7 & 11.5 & 31.2/0.4 & 1.10/2.8 \\
      & w/o \texttt{retrieve()}                   & 3.6 & 0.8 & 0.0 & 2.3 & 12.6 & 54.9/0.0 & 2.37/0.7 \\

    \bottomrule
  \end{tabular}
\end{table}

\textbf{(i) The optimal toolbox is backbone-dependent.}
No single tool configuration consistently dominates across models. While the \emph{Full} setup achieves strictly the best performance on 4/7 evaluation axes for Qwen3-VL-32B-Instruct and InternVL2.5-8B, it wins only 2/7 axes for Qwen2.5-VL-72B-Instruct and Lingshu-32B, where ablated configurations outperform the full toolbox on most tasks. Counts use $\kappa$ for workflow assessment and exclude ties. This suggests that the utility of external perception and retrieval depends strongly on the backbone's internal representations and priors.

\textbf{(ii) Retrieval can hurt medically specialized models.}
For Lingshu-32B, disabling \textsc{retrieve} improves performance on five of seven axes, most notably on Action Recognition ($26.7\!\to\!45.8$). Tool Localization ($6.5\!\to\!12.2$) and Triplet Detection ($0.0\!\to\!2.3$) exhibit similar trends. Disabling retrieval also improves three of seven axes for Qwen2.5-VL-72B-Instruct (tool detection, tool localization, and phase recognition), three for InternVL2.5-8B (triplet detection, phase recognition, and CVS assessment), and one for Qwen3-VL-32B-Instruct (skill assessment). These counts use $\kappa$ for workflow-assessment axes.

\textbf{(iii) Skill assessment is highly configuration-sensitive.}
Unlike perception-heavy tasks, skill assessment does not benefit consistently from any single toolbox configuration. Quadratic-weighted $\kappa$ varies substantially across settings for every backbone, with maximum swings of $21.2$, $11.7$, $10.5$, and $5.1$ points for Qwen2.5-VL-72B-Instruct, Lingshu-32B, Qwen3-VL-32B-Instruct, and InternVL2.5-8B, respectively. The optimal configuration differs across models, indicating that higher-level workflow evaluation depends critically on how external evidence is integrated, rather than on the mere availability of tools.

\subsubsection{Perception backbone, retrieval budget, and knowledge sources}
\label{app:tool_configuration}
Tab.~\ref{tab:tool_configuration} examines alternative perception backbones, retrieval budgets, and knowledge-base configurations for Qwen3-VL-8B and Qwen2.5-VL-7B. Entries are changes relative to Stage~I direct answers. They describe the performance of each configuration; the matched tool-access comparison is reported separately in Appendix~\ref{app:matched_tool_comparison}.

\begin{table}[htbp]\centering\tiny\setlength{\tabcolsep}{4pt}
\caption{Tool-configuration comparisons on \bench-Hard. Entries are differences from Stage~I direct answers, multiplied by 100. The action-recognition and skill-assessment ablations use reasoning enabled and a four-frame cap; the other ablation columns use reasoning disabled. These across-setting changes are distinct from the matched tool comparison in Tab.~\ref{tab:stage_delta_summary}, Panel~B.}
\label{tab:tool_configuration}
\begin{tabular}{@{}ll ccc cc cc@{}}
\toprule
Setting & Model & Tool Det. & Triplet Det. & Tool Loc. & Phase Rec. & Action Rec. & CVS Ass. & Skill Ass. \\
 & & {\scriptsize $\Delta$F1$\uparrow$} & {\scriptsize $\Delta$F1$\uparrow$} & {\scriptsize $\Delta F_1@0.5\uparrow$} & {\scriptsize $\Delta$Acc.$\uparrow$} & {\scriptsize $\Delta$F1$\uparrow$} & {\scriptsize $\Delta\kappa\uparrow$} & {\scriptsize $\Delta\kappa\uparrow$} \\
\midrule
\multirow{2}{*}{Main (budget $=2$, full KB, SAM2)} & Qwen3-VL-8B & \cellcolor[HTML]{C9EAC9} $+6.2$ & \textemdash & \cellcolor[HTML]{E9F7E9} $+1.6$ & \cellcolor[HTML]{D6EFD6} $+3.7$ & \cellcolor[HTML]{BEE6BE} $+27.3$ & \cellcolor[HTML]{FCF0F0} $-1.8$ & \cellcolor[HTML]{FBE7E7} $-2.1$ \\
 & Qwen2.5-VL-7B & \cellcolor[HTML]{D6EFD6} $+3.9$ & \textemdash & \cellcolor[HTML]{FCF0F0} $-0.8$ & \cellcolor[HTML]{D6EFD6} $+3.1$ & \cellcolor[HTML]{D6EFD6} $+3.0$ & \cellcolor[HTML]{FBE7E7} $-4.3$ & \cellcolor[HTML]{F5C8C8} $-12.7$ \\
\midrule
\multirow{2}{*}{Main w/ \wiki-Open} & Qwen3-VL-8B & \cellcolor[HTML]{C9EAC9} $+6.4$ & \textemdash & \cellcolor[HTML]{F7D4D4} $-8.0$ & \cellcolor[HTML]{C9EAC9} $+5.1$ & \cellcolor[HTML]{BEE6BE} $+22.2$ & \cellcolor[HTML]{F7D4D4} $-8.6$ & \cellcolor[HTML]{F5C8C8} $-12.2$ \\
 & Qwen2.5-VL-7B & \cellcolor[HTML]{D6EFD6} $+3.4$ & \textemdash & $0.0$ & \cellcolor[HTML]{D6EFD6} $+3.1$ & \cellcolor[HTML]{D6EFD6} $+4.5$ & \cellcolor[HTML]{FCF0F0} $-1.4$ & \cellcolor[HTML]{F5C8C8} $-10.8$ \\
\midrule
\multirow{2}{*}{Retrieve budget $=1$} & Qwen3-VL-8B & \cellcolor[HTML]{BEE6BE} $+14.6$ & \textemdash & $0.0$ & \cellcolor[HTML]{D6EFD6} $+2.9$ & \cellcolor[HTML]{BEE6BE} $+20.3$ & \cellcolor[HTML]{FBE7E7} $-3.0$ & \cellcolor[HTML]{F7D4D4} $-6.5$ \\
 & Qwen2.5-VL-7B & \cellcolor[HTML]{BEE6BE} $+13.6$ & \textemdash & $0.0$ & \cellcolor[HTML]{D6EFD6} $+2.3$ & \cellcolor[HTML]{C9EAC9} $+6.9$ & \cellcolor[HTML]{E9F7E9} $+1.3$ & \cellcolor[HTML]{E9F7E9} $+0.4$ \\
\midrule
\multirow{2}{*}{Retrieve budget $=3$} & Qwen3-VL-8B & \cellcolor[HTML]{BEE6BE} $+15.6$ & \textemdash & \cellcolor[HTML]{FCF0F0} $-0.8$ & \cellcolor[HTML]{D6EFD6} $+3.1$ & \cellcolor[HTML]{BEE6BE} $+20.9$ & \cellcolor[HTML]{FCF0F0} $-1.6$ & \cellcolor[HTML]{F5C8C8} $-13.4$ \\
 & Qwen2.5-VL-7B & \cellcolor[HTML]{BEE6BE} $+14.0$ & \textemdash & \cellcolor[HTML]{FCF0F0} $-0.8$ & \cellcolor[HTML]{D6EFD6} $+2.6$ & \cellcolor[HTML]{C9EAC9} $+5.7$ & \cellcolor[HTML]{D6EFD6} $+2.5$ & \cellcolor[HTML]{F5C8C8} $-12.0$ \\
\midrule
\multirow{2}{*}{w/o case\_record source} & Qwen3-VL-8B & \cellcolor[HTML]{BEE6BE} $+14.7$ & \cellcolor[HTML]{E9F7E9} $+0.7$ & \cellcolor[HTML]{FCF0F0} $-0.8$ & \cellcolor[HTML]{D6EFD6} $+3.4$ & \cellcolor[HTML]{BEE6BE} $+25.1$ & \cellcolor[HTML]{FBE7E7} $-4.0$ & \cellcolor[HTML]{BEE6BE} $+16.2$ \\
 & Qwen2.5-VL-7B & \cellcolor[HTML]{BEE6BE} $+15.0$ & \textemdash & \cellcolor[HTML]{E9F7E9} $+0.8$ & \cellcolor[HTML]{E9F7E9} $+0.6$ & \cellcolor[HTML]{BEE6BE} $+10.5$ & \cellcolor[HTML]{D6EFD6} $+2.5$ & \cellcolor[HTML]{C9EAC9} $+6.1$ \\
\midrule
\multirow{2}{*}{SurgSAM-2 \texttt{perceive()} backbone} & Qwen3-VL-8B & \cellcolor[HTML]{BEE6BE} $+14.8$ & \textemdash & \cellcolor[HTML]{F5C8C8} $-10.5$ & \cellcolor[HTML]{D6EFD6} $+2.3$ & \cellcolor[HTML]{BEE6BE} $+17.7$ & \cellcolor[HTML]{FBE7E7} $-2.1$ & \cellcolor[HTML]{C9EAC9} $+6.1$ \\
 & Qwen2.5-VL-7B & \cellcolor[HTML]{BEE6BE} $+14.8$ & \textemdash & \cellcolor[HTML]{F7D4D4} $-6.7$ & \cellcolor[HTML]{C9EAC9} $+5.1$ & \cellcolor[HTML]{BEE6BE} $+10.5$ & \cellcolor[HTML]{C9EAC9} $+7.2$ & \cellcolor[HTML]{F5C8C8} $-12.3$ \\
\bottomrule
\end{tabular}
\end{table}

\paragraph{Perception backbone.}
Replacing SAM~2 with the surgically adapted SurgSAM-2 retains task-dependent gains and regressions. Both backbones achieve a $+14.8$-point tool-detection gain relative to Stage~I, but tool localization declines by $10.5$ and $6.7$ points for Qwen3-VL-8B and Qwen2.5-VL-7B, respectively. Thus, a medically adapted perception backend alone does not remove the localization bottleneck.

\paragraph{Retrieval budget.}
We vary the per-source \texttt{retrieve()} call budget $k\in\{1,2,3\}$, with $k=2$ in the default setting. Increasing the budget from one to three calls does not produce uniform gains: tool detection and phase recognition improve slightly for both backbones, while skill-assessment changes fall from $-6.5$ to $-13.4$ points for Qwen3-VL-8B and from $+0.4$ to $-12.0$ for Qwen2.5-VL-7B. Access to more retrieved evidence therefore does not ensure better rubric-based assessment.

\paragraph{Case-record retrieval.}
Removing the case-record source retains positive tool-detection and action-recognition changes relative to Stage~I for both backbones. Workflow assessment remains model-dependent: without case records, the CVS-assessment change is $-4.0$ points for Qwen3-VL-8B and $+2.5$ for Qwen2.5-VL-7B. These results show that the utility of a knowledge source depends on how the agent uses its evidence.

\paragraph{\wiki-Open.}
Tab.~\ref{tab:cr_rb_open} compares Stage~II minus Stage~I changes for the full knowledge base and \wiki-Open, which excludes commercial sources, across temporal reasoning and workflow assessment.

\begin{table}[htbp]
\centering
\scriptsize
\setlength{\tabcolsep}{4pt}
\renewcommand{\arraystretch}{1.15}
\caption{Benchmark with \wiki-Open. Entries report Stage II minus Stage I changes on \bench-Hard, multiplied by 100. The table covers temporal reasoning and workflow assessment; higher values indicate larger gains.}
\label{tab:cr_rb_open}
\begin{tabular}{@{}llrrrr@{}}
\toprule
Model & Knowledge base & Phase $\Delta$Acc. & Action $\Delta$F1 & CVS $\Delta\kappa$ & Skill $\Delta\kappa$ \\
\midrule
Qwen3-VL-8B & Full KB & $+3.7$ & $+27.3$ & $-1.8$ & $-2.1$ \\
Qwen3-VL-8B & \wiki-Open & $+5.1$ & $+22.2$ & $-8.6$ & $-12.2$ \\
Qwen2.5-VL-7B & Full KB & $+3.1$ & $+3.0$ & $-4.3$ & $-12.7$ \\
Qwen2.5-VL-7B & \wiki-Open & $+3.1$ & $+4.5$ & $-1.4$ & $-10.8$ \\
\bottomrule
\end{tabular}
\end{table}

\subsection{State-of-the-art baselines}
To contextualize zero-shot generalist performance, Tab.~\ref{tab:supervised_sota_reference} collects published results from task-specific supervised methods. Their training data and evaluation protocols differ, so these values serve as reference points rather than a matched ranking against our agents.

\begin{table}[htbp]
\centering
\scriptsize
\caption{
Published supervised reference results for each task--dataset combination
in \bench, organized using the same capability taxonomy as
Tab.~\ref{tab:c0_full_suite_summary}. These methods typically rely on
task-specific architectures, dense supervision, or benchmark-specific
optimization, in contrast to the zero-shot generalist MLLM setting
evaluated in \bench. Results are intended as contextual reference
points only and are not directly comparable to \bench Stage~I
evaluation due to differences in supervision, protocol, and metric
definitions.
}
\label{tab:supervised_sota_reference}

\resizebox{\columnwidth}{!}{
\begin{tabular}{@{}lllll@{}}
\toprule
Task & Dataset & Reference Method & Reported Metric & Score \\
\midrule

\multicolumn{5}{@{}l}{\textit{Spatial Recognition}}\\
\cmidrule{1-5}

\multirow{4}{*}{Tool Detection}
  & CholecT50  & UniSurg \citep{wu2026unisurg}                & AP-I               & 91.55 \\
  & PitVis     & DKFZ Ensemble \citep{das2025pitvis}          & Macro-F1           & 41.70 \\
  & CATARACTS  & OVFM \citep{tu2026ovfm}                      & Micro-AUC          & 98.50 \\
  & GraSP      & TAPIS \citep{ayobi2025tapis}                 & mAP                & 94.33 \\

\addlinespace[1pt]

Triplet Detection
  & CholecT50  & UniSurg \citep{wu2026unisurg}                & AP-IVT             & 39.54 \\

\addlinespace[1pt]

Tool Localization
  & CholecT50  & MCIT-IG \citep{sharma2023mcitig}$^{\S}$     & Instrument mAP@0.5 & 60.10 \\

\midrule

\multicolumn{5}{@{}l}{\textit{Temporal Reasoning}}\\
\cmidrule{1-5}

\multirow{7}{*}{Phase Recognition}
  & Cataract-101    & GL-MSTCN \citep{fang2022glmstcn}           & Accuracy                & 96.50 \\
  & Cholec80        & Surgformer + NFSM \citep{ding2025nfsm}     & Accuracy / Jaccard      & 92.30 / 78.00 \\
  & AutoLaparo      & UniSurg \citep{wu2026unisurg}              & Accuracy / Jaccard      & 86.40 / 69.90 \\
  & GraSP           & TAPIS \citep{ayobi2025tapis}               & mAP / F1                & 76.72 / 63.42 \\
  & HeiCo           & V-JEPA \citep{pezold2025vjepa}$^{\P}$      & Accuracy                & 64.00 \\
  & PitVis          & UniSurg \citep{wu2026unisurg}              & Accuracy / Jaccard      & 86.50 / 65.50 \\
  & MultiBypass140  & ZEN \citep{park2026zen}                    & F1 / Accuracy / Jaccard & 89.42 / 72.73 / 62.36 \\

\addlinespace[1pt]

Action Recognition
  & HeiChole
  & Wintegral (ResNet50 ensemble) \citep{wagner2023heichole}
  & Average F1
  & 23.28 \\

\midrule

\multicolumn{5}{@{}l}{\textit{Workflow Assessment}}\\
\cmidrule{1-5}

CVS Assessment
  & Cholec80-CVS
  & ---$^{\dagger}$
  & ---
  & --- \\

\addlinespace[1pt]

Skill Assessment
  & HeiChole
  & CareSyntax (3D-ResNet18) \citep{wagner2023heichole}$^{\ddagger}$
  & MAE
  & 0.78 \\

\bottomrule
\end{tabular}
}

\vspace{2pt}

{\footnotesize

\noindent$^{\dagger}$ No published supervised reference currently exists
for the original 0/1/2 ordinal Strasberg--Sanford CVS protocol on
Cholec80-CVS. The closest related setting is ZEN
\citep{park2026zen}, which binarizes the labels and reports
38.56\% mAP / 33.22\% macro-F1 under a different task formulation.\par

\noindent$^{\ddagger}$ Sole submission to the EndoVis~2019 skill
sub-challenge. The dataset authors note that the model produced nearly
constant criterion scores across test videos and is therefore not
representative of robust surgical skill assessment.\par

\noindent$^{\P}$ Reported on the held-out HeiCo split following the
V-JEPA evaluation protocol.\par

\noindent$^{\S}$ MCIT-IG evaluates on the CholecTriplet2022 private
test set using external bounding-box supervision and is therefore not
directly comparable to \bench. The fully weakly-supervised
challenge winner ResNet-CAM-YOLOv5
\citep{nwoye2023cholectriplet} reports 41.9\% AP-I at the same
threshold.\par

}

\end{table}

\subsection{Analysis of agent tool-use behavior}
\label{app_toolbehavior}

\begin{figure}[htbp]  
    \centering
    \begin{subfigure}[t]{0.48\linewidth}
        \centering
        \includegraphics[width=\linewidth]{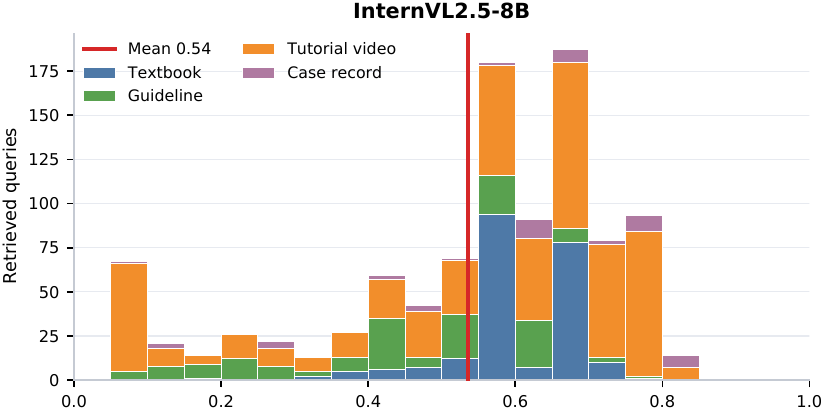}
        \caption{InternVL2.5-8B}
        \label{fig:rerank_intern}
    \end{subfigure}
    \hfill
    \begin{subfigure}[t]{0.48\linewidth}
        \centering
        \includegraphics[width=\linewidth]{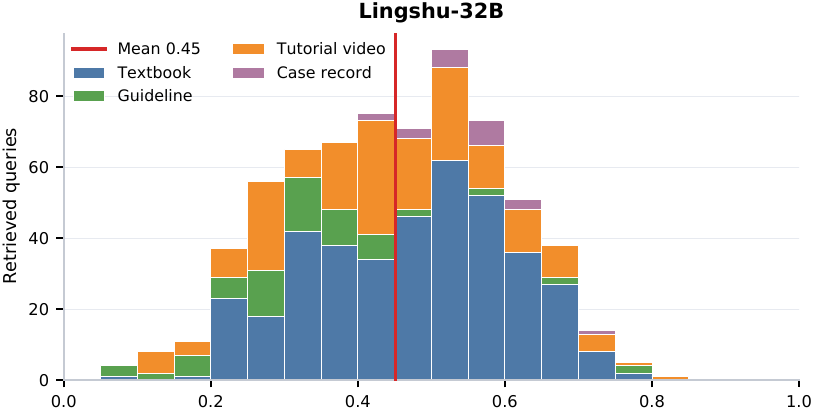}
        \caption{Lingshu-32B}
        \label{fig:rerank_lingshu}
    \end{subfigure}

    \vspace{0.5em}

    \begin{subfigure}[t]{0.48\linewidth}
        \centering
        \includegraphics[width=\linewidth]{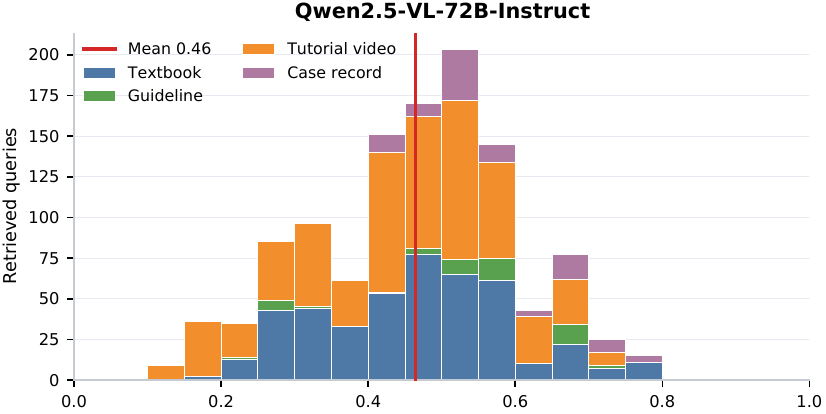}
        \caption{Qwen2.5-VL-72B-Instruct}
        \label{fig:rerank_qwen25}
    \end{subfigure}
    \hfill
    \begin{subfigure}[t]{0.48\linewidth}
        \centering
        \includegraphics[width=\linewidth]{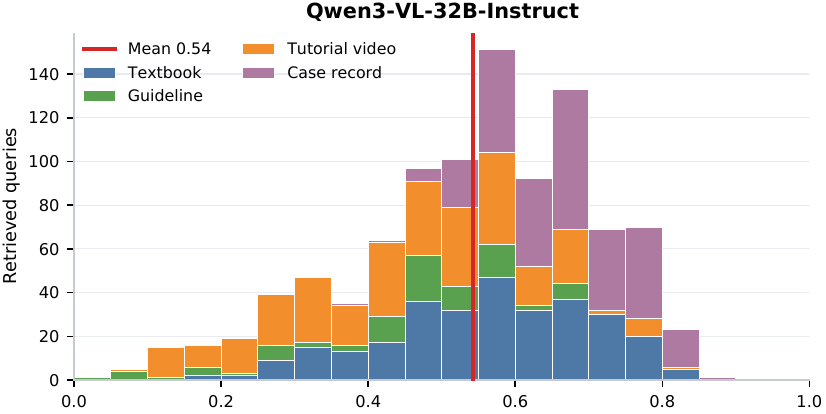}
        \caption{Qwen3-VL-32B-Instruct}
        \label{fig:rerank_qwen3}
    \end{subfigure}
    \caption{Reranker score distributions across four backbones on
    \textsc{\bench}-Hard in the Full setting. Bars are stacked
    by retrieval source type, and the red vertical line marks the mean
    Jina reranker score for each model.}
    \label{fig:rerank_all}
\end{figure}

To further inspect retrieval quality, we analyze the Jina reranker scores
returned by \texttt{retrieve()} in the Full setting. We focus on the four
trajectory-analysis models used in Section~\ref{sec:trajectory_analysis}. Each histogram in Fig.~\ref{fig:rerank_all}
includes only successful retrieval calls with a returned reranker score; rejected or warning-triggering calls are excluded from the score
distribution.

The score distributions show that retrieval behavior differs substantially across backbones. Qwen3-VL-32B-Instruct and InternVL2.5-8B obtain higher mean reranker scores, with many calls
concentrated around moderate-to-high relevance. Qwen2.5-VL-72B-Instruct and Lingshu-32B have broader distributions centered at lower scores. These differences suggest that retrieval quality is
not determined only by access to the same knowledge base and backend; it also depends on how each agent formulates queries and chooses source types. Thus, the retrieval tool exposes a second
bottleneck beyond final-answer integration: agents must first ask searchable, source-appropriate questions.

\subsection{Retrieval behavior}
\begin{figure}[htbp]
      \centering
      \includegraphics[width=1\linewidth]{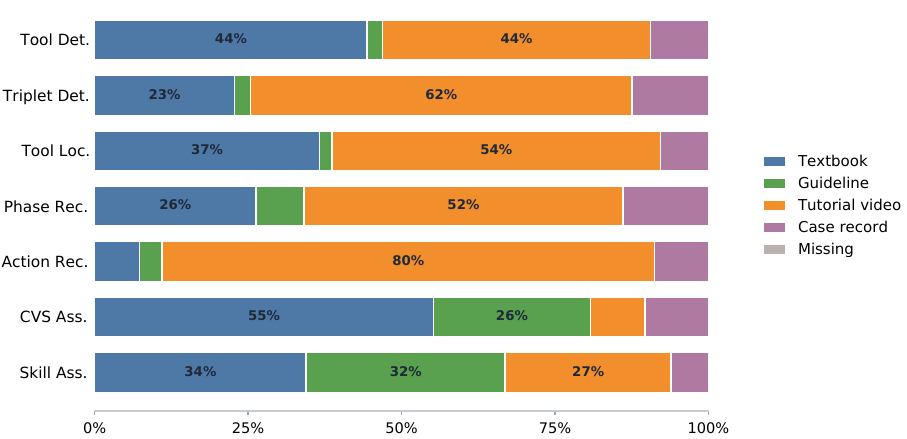}
      \caption{Retrieval source distribution aggregated over the four trajectory-analysis models. Percentages are computed over retrieval calls for each task.}
      \label{fig:traj_retrieve_source}
  \end{figure}

Fig.~\ref{fig:traj_retrieve_source} separates retrieval calls by evidence source for each task, complementing the tool-invocation patterns in Fig.~\ref{fig:traj_tool_use_pattern}.

\noindent\textbf{Tool use and episode outcomes.}
Tab.~\ref{tab:traj_success_conditioned} compares tool-use patterns between successful and failed episodes for the four trajectory-analysis models. The success-conditioned rates and call counts describe associations between agent behavior and outcomes; they do not isolate the causal effect of a tool call.

\begin{table}[htbp]
  \centering
  \tiny
  \setlength{\tabcolsep}{5pt}
  \caption{Success-conditioned trajectory statistics for the four-model Full-setting analysis on \bench-Hard. Success is computed by strict agreement between the parsed structured
  prediction and ground truth, independent of reward. Tool localization is excluded from this table because exact equality between continuous predicted and ground-truth bounding boxes is not a
  meaningful success criterion. The table covers 3,328 model--episode pairs (four models $\times$ 832 episodes). \texttt{perceive()}, \texttt{retrieve()}, Both, and Warning are percentages within each outcome group; Calls/ep. is the mean number of tool calls per episode.}
  \label{tab:traj_success_conditioned}
  \begin{tabular}{@{}llrrrrrr@{}}
  \toprule
  Task family & Outcome & Episodes & \texttt{perceive()} & \texttt{retrieve()} & Both & Warning & Calls/ep. \\
  \midrule
  Spatial Recognition & Success & 71 & 66.2 & 69.0 & 60.6 & 25.4 & 2.62 \\
  Spatial Recognition & Failure & 1129 & 81.9 & 60.2 & 59.5 & 22.8 & 2.28 \\
  \addlinespace[1pt]
  Temporal Reasoning & Success & 83 & 47.0 & 78.3 & 37.3 & 9.6 & 1.94 \\
  Temporal Reasoning & Failure & 1517 & 63.4 & 69.1 & 47.9 & 20.8 & 2.12 \\
  \addlinespace[1pt]
  Workflow Assessment & Success & 41 & 58.5 & 63.4 & 48.8 & 9.8 & 2.59 \\
  Workflow Assessment & Failure & 487 & 61.4 & 63.2 & 54.2 & 22.2 & 3.28 \\
  \addlinespace[1pt]
  All & Success & 195 & 56.4 & 71.8 & 48.2 & 15.4 & 2.32 \\
  All & Failure & 3133 & 69.8 & 65.0 & 53.0 & 21.7 & 2.36 \\
  \bottomrule
  \end{tabular}
  \end{table}

\section{Computational resources}
\label{app:compute}

All open-source experiments were conducted on the
\textbf{NHR@FAU Alex} cluster. Closed-source model evaluations
were run through their respective public APIs and did not consume
local GPU resources.

\paragraph{Hardware.}
We used two GPU partitions of Alex:
(i) \textbf{\texttt{alex-a100}}: nodes with NVIDIA A100~80\,GB SXM4
GPUs interconnected via NVLink, used for all VLM inference;
(ii) \textbf{\texttt{alex-a40}}: nodes with NVIDIA A40~48\,GB GPUs,
used for lighter-weight evaluation and data preprocessing.

\paragraph{Software stack.}
PyTorch, HuggingFace Transformers, and vLLM for high-throughput
inference, running under CUDA on Rocky Linux~8.

\paragraph{Total compute.}
Including preliminary, failed, and ablation runs, the total compute
budget of this paper is approximately
\textbf{4{,}766 GPU\,h on A100~80\,GB} and
\textbf{703 GPU\,h on A40~48\,GB}, for a combined
\textbf{$\sim$5{,}470 GPU\,h}.

\section{License}
\label{app:license}

We provide the evaluation code and a data curation pipeline; we do not distribute data. Users obtain the source datasets directly from their original providers under the applicable access conditions and licenses. \bench harmonizes these datasets into a unified evaluation format (Appendix~\ref{app:dataset_harmonization}). Our harmonization code is released under \textbf{CC~BY-NC~4.0}; the source datasets remain subject to their original licenses, listed below with access links.

\begingroup
\renewcommand{\UrlFont}{\ttfamily\footnotesize}

\textbf{CholecT50}~\citep{nwoye2022rendezvous}: CC~BY-NC-SA~4.0. \url{https://github.com/CAMMA-public/cholect50/blob/master/LICENSE}

\textbf{Cholec80}~\citep{twinanda2017endonet}: CC~BY-NC-SA~4.0. \url{https://camma.unistra.fr/datasets/}

\textbf{Cholec80-CVS}~\citep{rios2023cholec80cvs}: the CVS annotations are published under CC0; the underlying videos remain subject to the Cholec80 license. \url{https://doi.org/10.6084/m9.figshare.22183885}

\textbf{HeiChole}~\citep{wagner2023heichole}: CC~BY-NC-SA~4.0. \url{https://www.synapse.org/Synapse:syn18824884}

\textbf{PitVis}~\citep{das2025pitvis}: the provider's current release (version~3) is licensed under CC~BY-NC-SA~4.0. \url{https://rdr.ucl.ac.uk/articles/dataset/PitVis_Challenge_Endoscopic_Pituitary_Surgery_videos/26531686}

\textbf{Cataract-101}~\citep{schoeffmann2018cataract101}: CC~BY-NC~4.0. \url{https://ftp.itec.aau.at/datasets/ovid/cat-101/}

\textbf{CATARACTS}~\citep{alhajj2019cataracts}: research-use only. \url{https://cataracts.grand-challenge.org/}

\textbf{GraSP}~\citep{ayobi2025tapis}: MIT license (public repository); dataset access through the project repository. \url{https://github.com/BCV-Uniandes/GraSP/blob/main/LICENSE}

\textbf{AutoLaparo}~\citep{wang2022autolaparo}: CC~BY-NC-SA~4.0; academic research only. \url{https://autolaparo.github.io/}

\textbf{HeiCo}~\citep{maierhein2021heico}: CC~BY-NC-SA~4.0. \url{https://www.synapse.org/Synapse:syn21903917}

\textbf{MultiBypass140}~\citep{lavanchy2024multibypass140}: CC~BY-NC-SA~4.0. \url{https://github.com/CAMMA-public/MultiBypass140/blob/main/LICENSE}

\endgroup

\FloatBarrier
\RemoveFromHook{cmd/section/before}[appendix-float-order]
\RemoveFromHook{cmd/subsection/before}[appendix-float-order]
\RemoveFromHook{cmd/subsubsection/before}[appendix-float-order]
\RemoveFromHook{env/figure/before}[appendix-float-order]
\RemoveFromHook{env/figure*/before}[appendix-float-order]
\RemoveFromHook{env/figure/after}[appendix-float-order]
\RemoveFromHook{env/figure*/after}[appendix-float-order]

\end{document}